%% file: main.tex
\documentclass{article}

\usepackage{microtype}
\usepackage{graphicx}
\usepackage{subfigure}
\usepackage{booktabs} 

\usepackage{hyperref}

\usepackage[accepted]{icml2024}

\usepackage{amsmath}
\usepackage{amssymb}
\usepackage{mathtools}
\usepackage{amsthm}

\usepackage[noabbrev,capitalize,nameinlink]{cleveref}
\hypersetup{colorlinks={true},linkcolor={darkred},citecolor=darkblue}
\theoremstyle{plain}
\newtheorem{theorem}{Theorem}[section]
\newtheorem{proposition}[theorem]{Proposition}

\theoremstyle{definition}

\theoremstyle{remark}

\usepackage{algorithmic}
\usepackage[algo2e]{algorithm2e}
\usepackage[utf8]{inputenc}
\usepackage[cjk]{kotex}
\usepackage{subfigure}
\usepackage{multirow}
\usepackage{pbox}
\usepackage{booktabs}
\usepackage{graphicx}
\usepackage{bm}
\usepackage{siunitx}
\usepackage{lipsum}
\usepackage{wrapfig}
\usepackage{mathtools}
\input{math_commands.tex}

\usepackage[textsize=footnotesize]{todonotes}
\usepackage{marginnote}
 
\newcommand{\std}[1]{\scriptsize{\ensuremath{\;\pm\;}}\text{\scriptsize #1}}

\newcommand{\BEST}[1]{\textcolor{red}{#1}}
\newcommand{\SECOND}[1]{\textcolor{blue}{#1}}
\newcommand{\THIRD}[1]{\textcolor{violet}{#1}}

\definecolor{bluegray}{rgb}{0.4, 0.6, 0.8}
\definecolor{electriclime}{rgb}{0.8, 1.0, 0.0}
\definecolor{malachite}{rgb}{0.04, 0.85, 0.32}
\definecolor{darkred}{rgb}{0.55, 0.0, 0.0}
\definecolor{darkblue}{rgb}{0.0, 0.0, 0.55}
\definecolor{darkgreen}{rgb}{0.0, 0.2, 0.13}
\definecolor{darkorchid}{rgb}{0.6, 0.2, 0.8}

\Crefname{figure}{Fig.}{Figs.}
\usepackage{pifont}
\usepackage{listings}
\definecolor{codegreen}{rgb}{0,0.6,0}
\definecolor{codegray}{rgb}{0.5,0.5,0.5}
\definecolor{codepurple}{rgb}{0.58,0,0.82}
\definecolor{backcolour}{rgb}{0.95,0.95,0.92}
\lstdefinestyle{mystyle}{
    backgroundcolor=\color{backcolour},   
    commentstyle=\color{codegreen},
    keywordstyle=\color{magenta},
    numberstyle=\tiny\color{codegray},
    stringstyle=\color{codepurple},
    basicstyle=\ttfamily\footnotesize,
    breakatwhitespace=false,         
    breaklines=true,                 
    captionpos=b,                    
    keepspaces=true,                 
    numbers=left,                    
    numbersep=1pt,
    showspaces=false,                
    showstringspaces=false,
    showtabs=false,                  
    tabsize=2
}
\definecolor{diffstart}{named}{gray}
\definecolor{diffincl}{named}{green}
\definecolor{diffrem}{named}{orange}
\lstdefinelanguage{diff}{
basicstyle=\ttfamily\small,
morecomment=[f][\color{diffstart}]{@@},
morecomment=[f][\color{diffincl}]{+\ },
morecomment=[f][\color{diffrem}]{-\ },
}
\definecolor{commentcolor}{RGB}{110,154,155}   
\definecolor{lightRed}{HTML}{F8CECC}
\definecolor{redBorder}{HTML}{CC0000}
\definecolor{greenBorder}{HTML}{82B366}
\definecolor{lightGreen}{HTML}{D5E8D4}
\definecolor{blueBorder}{HTML}{6C8EBF}
\definecolor{lightBlue}{HTML}{DAE8FC}

\DeclareRobustCommand\myrednode{\raisebox{-2pt} {\tikz[]{\node[shape=circle,draw=redBorder,fill=lightRed,text=lightRed, inner sep=1pt]{\scriptsize{A}};}}}
\DeclareRobustCommand\mybluenode{\raisebox{-2pt} {\tikz[]{\node[shape=circle,draw=blueBorder,fill=lightBlue,text=lightBlue,inner sep=1pt]{\scriptsize{A}};}}}

\icmltitlerunning{PANDA: Expanded Width-Aware Message Passing Beyond Rewiring}

\begin{document}

\twocolumn[

\icmltitle{PANDA: Expanded Width-Aware Message Passing Beyond Rewiring}



\icmlsetsymbol{equal}{*}

\begin{icmlauthorlist}
\icmlauthor{Jeongwhan Choi}{yonsei}
\icmlauthor{Sumin Park}{dni}
\icmlauthor{Hyowon Wi}{yonsei}
\icmlauthor{Sung-Bae Cho}{yonsei}
\icmlauthor{Noseong Park}{kaist}
\end{icmlauthorlist}

\icmlaffiliation{yonsei}{Yonsei University, Seoul, South Korea.}
\icmlaffiliation{dni}{DNI Consulting, Seoul, South Korea.}
\icmlaffiliation{kaist}{KAIST, Daejeon, South Korea}

\icmlcorrespondingauthor{Noseong Park}{noseong@kaist.ac.kr}

\icmlkeywords{Message Passing, Over-squashing, Graph Neural Networks}

\vskip 0.3in
]



\printAffiliationsAndNotice{}  

\begin{abstract}
Recent research in the field of graph neural network (GNN) has identified a critical issue known as ``over-squashing,'' resulting from the bottleneck phenomenon in graph structures, which impedes the propagation of long-range information. 
Prior works have proposed a variety of graph rewiring concepts that aim at optimizing the spatial or spectral properties of graphs to promote the signal propagation. 
However, such approaches inevitably deteriorate the original graph topology, which may lead to a distortion of information flow. 
To address this, we introduce an ex\textbf{pand}ed width-\textbf{a}ware (\textbf{PANDA}) message passing, a new message passing paradigm where nodes with high centrality, a potential source of over-squashing, are selectively expanded in width to encapsulate the growing influx of signals from distant nodes. 
Experimental results show that our method outperforms existing rewiring methods, suggesting that selectively expanding the hidden state of nodes can be a compelling alternative to graph rewiring for addressing the over-squashing.
\end{abstract}

\section{Introduction}
Graph Neural Networks (GNNs) have emerged as a powerful tool for graph data processing and are being studied extensively in various domains, as evidenced by significant research contributions~\citep{defferrard2016chebnet,velickovic2018GAT,chen2020gcnii,chien2021GPRGNN,choi2021ltocf,chamberlain2021grand,hwang2021climate,choi2023STGNRDE,choi2023gread,kang2023node,gruber2023reversible}. A key subclass within GNNs is Message Passing Neural Networks (MPNNs), which excel in propagating information through neighbouring nodes. As MPNNs propagate only for a 1-hop distance within a single layer, long-range interactions can only be captured when the depth of the network is comparable to the distance. However, increasing the number of layers for capturing long-range dependencies results in exponential growths of the receptive field and excessive aggregations of repeated messages that are subsequently packed into a feature vector of fixed length, which causes the problem called ``over-squashing''~\citep{alon2021oversquashing,shi2023exposition}. This phenomenon severely decreases the expressivity of the networks and thus impairs the model performance. Previous studies have attempted to identify topological bottlenecks in a graph and directly modify graph connectivity with various optimization targets, including curvature, effective resistance, and spectral gap~\citep{alon2021oversquashing,topping2022riccurvature,banerjee2022rlef,deac2022expander,arnaiz2022diffwire,karhadkar2023fosr,black2023gtr,fesser2023afrc,sonthalia2023relwire,giraldo2023sjlr,shi2023curvature,barbero2023laser}.

\begin{figure}[t]
    \centering
    \subfigure[Molecular graph]{\includegraphics[width=0.44\columnwidth]{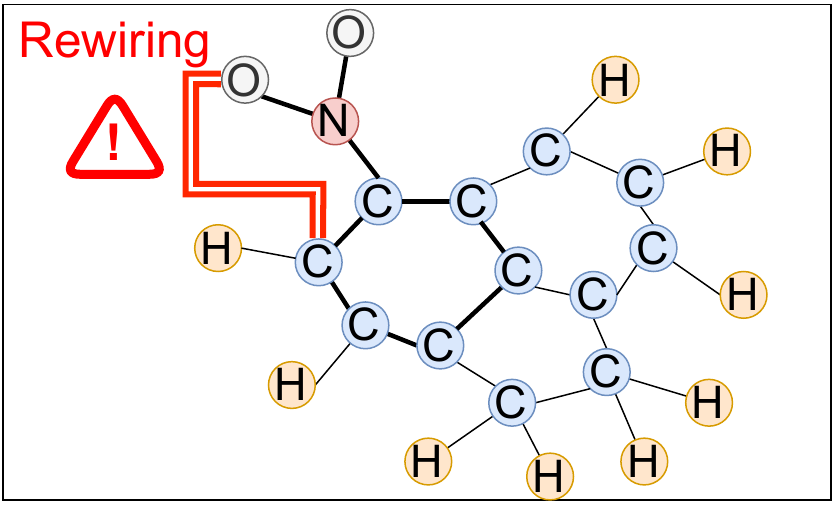}\label{fig:intro-rewiring-a}}
    \subfigure[Social graph]{\includegraphics[width=0.44\columnwidth]{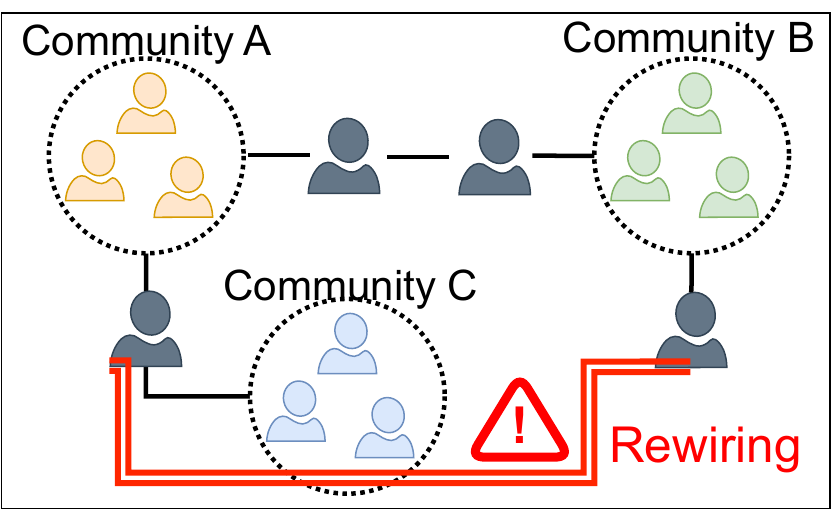}\label{fig:intro-rewiring-b}}
    \caption{Potential pitfalls of rewiring in domain-specific graphs: (a) In a molecular graph, rewiring the edge in red to a benzene ring violates the domain knowledge. (b) In a social graph, connecting a user to his/her enemy may lead to totally different meaning.}
    \label{fig:intro-rewiring}
\end{figure}
\begin{figure}[t!]
    \centering
    \subfigure[\textsc{Mutag}]{\includegraphics[width=0.49\columnwidth]{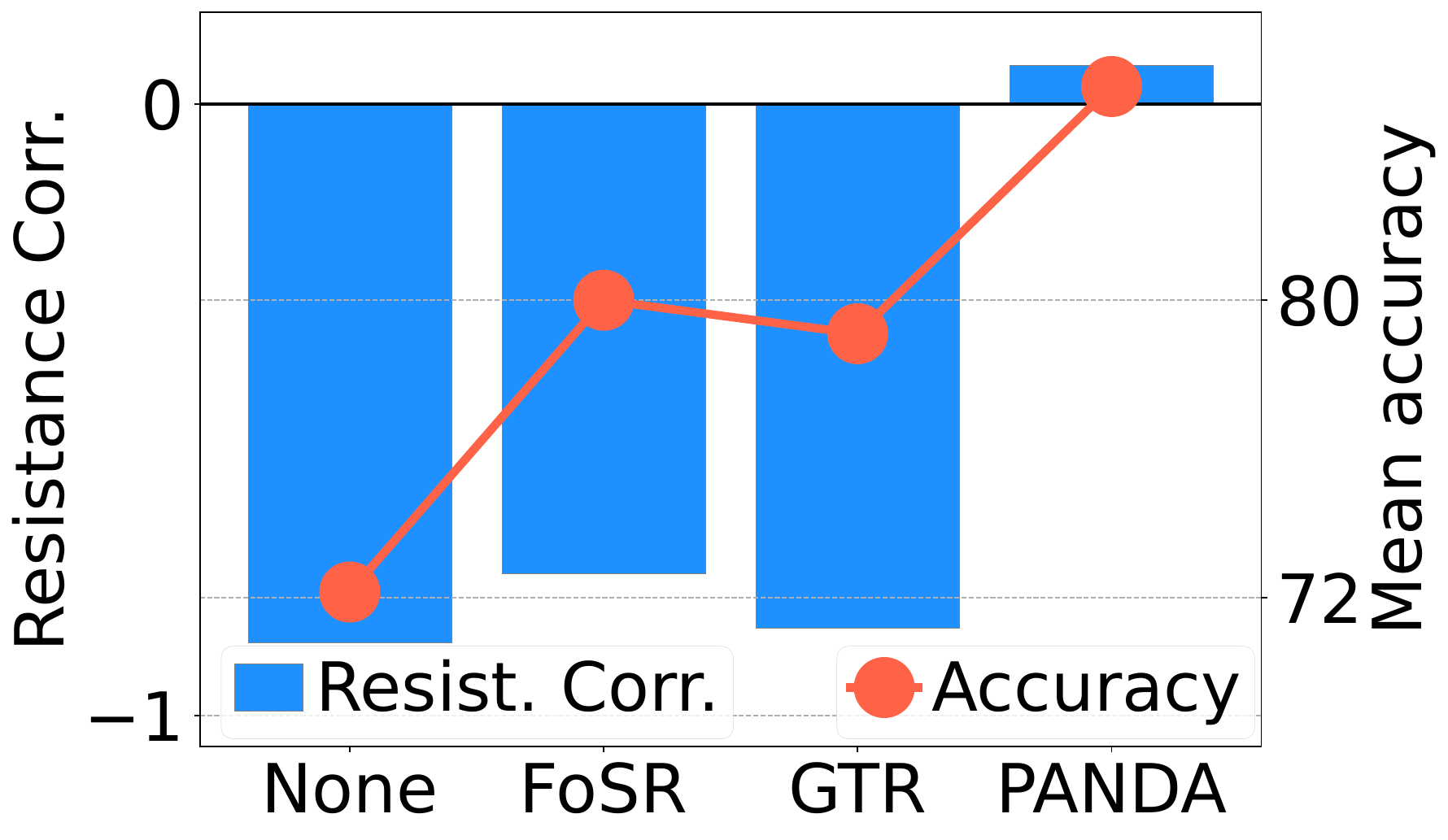}}
    \subfigure[\textsc{Reddit-Binary}]{\includegraphics[width=0.49\columnwidth]{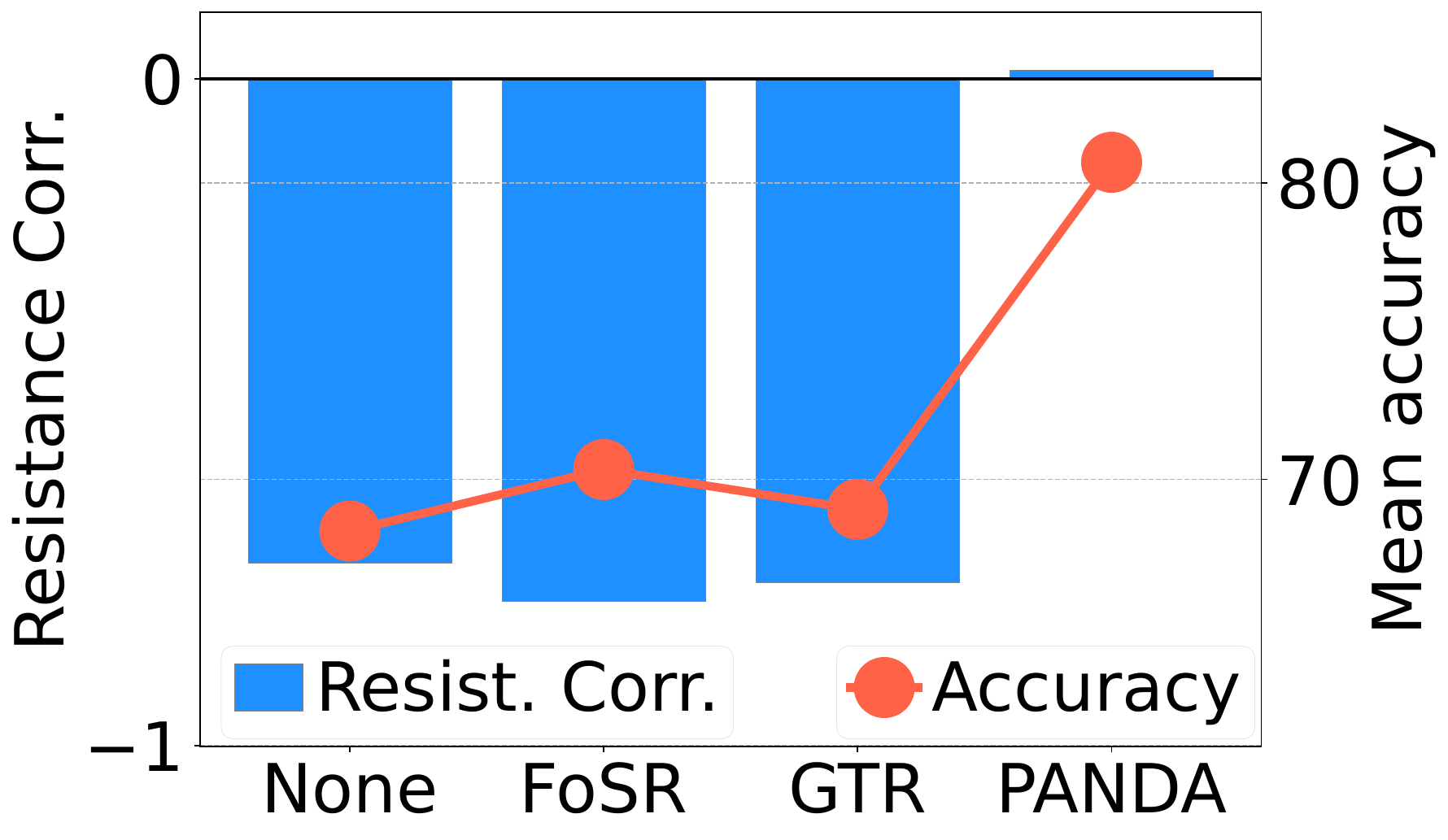}}
    \caption{Comparison of resistance correlation and mean accuracy across different methods for GCN. A large negative correlation reflects that a higher total effective resistance is associated with reduced signal propagation (see \Cref{sec:why}).} 
    \label{fig:resistance}
\end{figure}

\paragraph{Motivation.}
Existing rewiring methods that alter the graph topology to resolve over-squashing, while potentially beneficial, can inadvertently introduce inaccuracies within domain-specific contexts.
\Cref{fig:intro-rewiring-a} demonstrates how modifying the molecular graph of a benzene ring could contradict chemical principles, while \Cref{fig:intro-rewiring-b} shows that rewiring in social networks could result in the distortion of underlying community structures. These examples highlight the necessity to preserve the original graph structure to maintain the validity of domain-specific semantic information. 

Prior works have tried to identify various factors associated with the over-squashing through the lens of sensitivity analysis of Jacobian of node features~\citep{topping2022riccurvature,giovanni2023oversquashing}. As one of those trials, \citet{giovanni2023oversquashing}, an insightful analytical paper, provides a theoretical justification that increasing the width of the model (i.e., hidden dimension) can also contribute to improving the sensitivity of the model. However, it points out a major concern regarding expanding the width of the model to address over-squashing; increasing the hidden dimension globally affects the whole networks at the expense of the generalization capacity of the model. 

Inspired by the limitation of existing rewiring methods and the promising role of width expansion for addressing the over-squashing, we aim to design a novel message passing paradigm that mitigates the over-squashing by targeting the nodes that are mainly involved in creating bottlenecks in the signal flow and selectively expanding their width without the risk to change the underlying graph topology. 

\begin{figure*}[t]
    \centering
    \subfigure[Expanded nodes]{\includegraphics[width=0.18\textwidth]{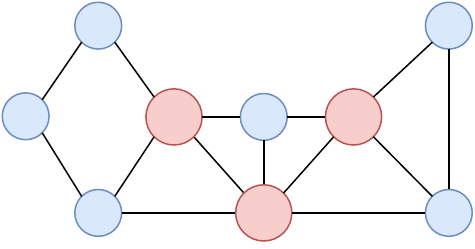}\label{fig:example-a}}\hfill
    \subfigure[Low-to-low]{\includegraphics[width=0.18\textwidth]{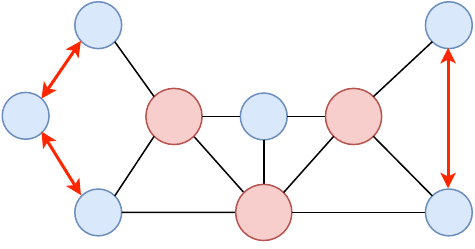}\label{fig:example-b}}\hfill
    \subfigure[High-to-high]{\includegraphics[width=0.18\textwidth]{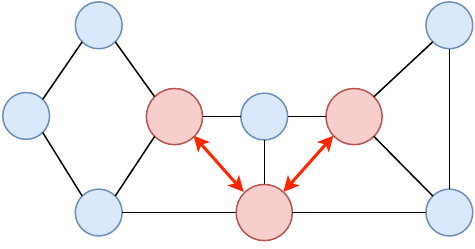}\label{fig:example-c}}\hfill
    \subfigure[Low-to-high]{\includegraphics[width=0.18\textwidth]{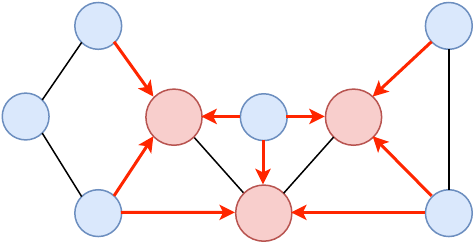}\label{fig:example-d}}\hfill
    \subfigure[High-to-low]{\includegraphics[width=0.18\textwidth]{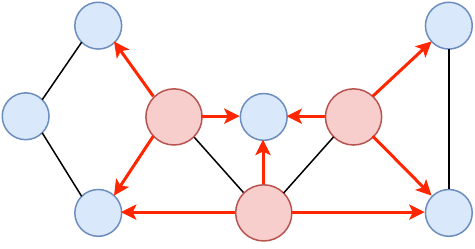}\label{fig:example-e}}
    \caption{Examples of PANDA's message passing mechanism. The size of the node indicates the size of hidden dimension.}
    \label{fig:example}
\end{figure*}

\paragraph{Main idea.}
It is possible to define graph bottlenecks as nodes with high centrality, such as betweenness centrality~\citep{yu2007importance,topping2022riccurvature}. Nodes with high centrality can show hub-like behavior in that they connect to a relatively larger set of nodes lying on the different parts of the graph, and thus receive excessive information from them. Increasing the width of the feature vector of these nodes provides more space for them to process information flowing from other nodes. Then, we can take advantage of their enhanced capacities to alleviate over-squashing. In order to do this, it requires exchanging messages between nodes with different hidden dimensions. This raises a natural question:

\textit{``Is it possible to design a message passing that enables information exchange between nodes with different width only in a way that mitigates over-squashing without rewiring?''}

We propose a new message passing framework that addresses the above question. First, the entire node set is divided into two subsets, expanded and non-expanded nodes, which are colored in red (\myrednode) and blue (\mybluenode), respectively in \Cref{fig:example}. Then, we classify four different edge types depending on the expansion state of the nodes at source (edge tail) and target (edge head) position: low-to-low, high-to-high, low-to-high, and high-to-low. Each edge type requires a distinct message passing scheme.

The standard message passing between nodes with equal widths corresponds to the conventional message passing (see \Cref{fig:example-b,fig:example-c}).
For high-to-low message passing (see \Cref{fig:example-d}), a low-dimensional node selects and receives a subset of the information sent by a high-dimensional node.
The low-to-high message passing involves augmenting the hidden vectors of low-dimensional nodes to match the higher dimension size for propagation (see \Cref{fig:example-e}).

Our framework can potentially enhance the capacity of nodes in the perspective of signal propagation. As shown in \Cref{fig:resistance}, our framework maintains constant signal propagation w.r.t. effective resistance and \emph{simultaneously} improves the accuracy compared to the existing rewiring methods. We use the correlation coefficient to quantify how signal propagation decreases as effective resistance increases.

\paragraph{Contributions and outline.} 
We introduce an ex\textbf{pand}ed width-\textbf{a}ware message passing (\textbf{PANDA})\footnote{Our source code is available here: \url{https://github.com/jeongwhanchoi/panda}.}, a novel paradigm for message passing with expanded widths for nodes that are potentially bottlenecks. Our contributions can be summarized as follows:
\begin{itemize}
    \item In \Cref{sec:method}, we propose \textbf{PANDA} that enables the signal propagation across the nodes of different widths.
    \item In \Cref{sec:why}, we discuss how our \textbf{PANDA} can alleviate over-squashing in terms of various perspectives. Verifying a higher feature sensitivity compared to other models, we show \textbf{PANDA} can overcome topological bottleneck, maintaining consistent signal propagation even under large effective resistance. Finally, we show that our method solves the limitation of existing rewiring methods, e.g., over-smoothing.
    \item In \Cref{sec:exp}, we empirically demonstrate that our \textbf{PANDA} outperforms existing rewiring methods.
\end{itemize}

\section{Preliminaries}
We first examine the notation used in this paper.
Given a graph $\mathcal{G}=(\mathcal{V},\mathcal{E})$, we use $\mathcal{V}$ and $\mathcal{E}$ to denote its nodes and edges, respectively.
The nodes are indexed by $v$ and $u$ such that $v,u\in\mathcal{V}$, and an edge connecting nodes $v$ and $u$ is denoted by $(v,u)\in\mathcal{E}$.
The connectivity is encoded in the adjacency matrix $\mathbf{A} \in \mathbb{R}^{n\times n}$ where $n$ is the number of nodes. 
$p$ denotes the width (hidden dimension size), while $\ell$ is the number of layers. 
The feature of node $v$ at layer $\ell$ is written as $\mathbf{h}_v^{(\ell+1)}$.

\subsection{Message Passing Neural Networks}
We consider the case where each node $v$ has a feature $\mathbf{h}^{(0)}_v \in \mathbb{R}^p$. 
MPNNs iteratively update node representations using the following equations:
\begin{align}
    \mathbf{m}_v^{(\ell)} &= \psi^{(\ell)}( \{ \mathbf{h}_{u}^{(\ell)} :u \in \mathcal{N}(v) \}),\\
    \mathbf{h}_v^{(\ell+1)} &= \phi^{(\ell)} (\mathbf{h}_v^{(\ell)},  \mathbf{m}_v^{(\ell)} ).
\end{align}
where $\mathbf{m}_v^{(\ell)}$ is the aggregated node feature of $v$'s neighbourhood. The aggregation function $\psi^{(\ell)}$ and the update function $\phi^{(\ell)}$ are learnable, and their different definitions result in different architectures~\citep{kipf2017GCN,xu2018GIN,velickovic2018GAT}.
In default, all nodes have the same hidden dimension size. 
We will introduce a method that enables message passing among nodes with different hidden dimensions in \Cref{sec:method}.

\subsection{Over-squashing Problem}
Initially identified by \citet{alon2021oversquashing}, over-squashing has become a significant challenge in GNNs when dealing with long-range dependencies. It mainly occurs when the information aggregated from too many neighbours is squashed into a fixed-sized node feature vector, resulting in a substantial loss of information~\citep{shi2023exposition}. 
We will review the literature, including rewiring methods, to address this problem, in \Cref{sec:related_work}. We now describe the Jacobian sensitivity and effective resistance widely used to estimate the over-squashing. In \Cref{sec:why}, we justify our method in terms of the two metrics.

\paragraph{Sensitivity bound.} Recent theoretical insights on GNNs have provided a formal understanding of over-squashing via the lens of the sensitivity analysis~\citep{topping2022riccurvature}. The sensitivity bound theorem, as proposed by~\citep[Theorm 3.2]{giovanni2023oversquashing}, posits that the impact of over-squashing can be quantified by examining the following 3 factors: i) the Lipschitz continuity of the non-linearity in the model, ii) the width of the model, and iii) the topology of a graph. The sensitivity of $\mathbf{h}_v^{(\ell)}$ to $\mathbf{h}_u^{(0)}$, assuming that there exists a $\ell$-path between nodes $v$ and $u$, is bounded by
\begin{align}    
\left \| \frac{\partial \mathbf{h}_v^{(\ell)}}{\partial \mathbf{h}_u^{(0)}} \right \|_1 \leq  (\underbrace{zwp}_{\textrm{model}})^{\ell} \underbrace{(\mathbf{S}^{\ell})_{vu}}_{\textrm{topology}}.\label{eq:sens}
\end{align} 
The impact of the model to over-squashing is bounded by the Lipschitz constant $z$, the width $p$, and the maximum weight $w$, while the topology of an input graph affects the sensitivity by $\ell$-th power of the graph shift matrix $\mathbf{S}$. Hereinafter, $\mathbf{S}$ can be any (normalized) adjacency matrix (with self-loops).

A small Jacobian norm indicates that the final representations of a node learned by a model are not much affected by the variations in its neighbouring inputs, implying that the network is suffered from over-squashing. This condition typically arises when the information has not been sufficiently propagated over the graph due to the excessive aggregation of information, i.e., compression into a small hidden vector, through several layers of message passing.

\paragraph{Effective resistance and signal propagation.} Derived from the field of electrical engineering, the effective resistance between two nodes $u$ and $v$ in an electrical network is defined as the potential difference induced across the edges when a unit current is injected at one of each end~\citep{ghosh2008minimizing}. An algebraic expression for the effective resistance is given in \Cref{app:effresdef}. 
Rayleigh’s monotonicity principle, which says that adding paths or shortening existing paths can only decrease the effective resistance between two nodes~\citep{thomassen1990resistances}, leads to the following interpretation: more and shorter disjoint paths connecting the nodes $u$ and $v$ lead to a lower resistance between them~\citep{black2023gtr, devri2022effcurvature}. Therefore, edges with high effective resistance struggle to propagate information, making bottlenecks. 
The total effective resistance $R_{tot}$, the sum of the effective resistance among all pairs of nodes (see Eq.~\eqref{eq:r_total}), is a key measure for measuring the overall degree of over-squashing in a graph. It is known that the signal propagation of GNNs is inversely proportional to $R_{tot}$~\citep{giovanni2023oversquashing}, which will be covered in detail in \Cref{sec:why} with a comparison to our method.

\paragraph{Over-smoothing problem.}\label{pre:tradeoff}
Over-smoothing is another well-known problem that reduces the expressiveness of GNNs in deeper layers, resulting in a loss of discriminative power in the representation of node features~\citep{li2018deeper,nt2019revisiting,oono2020oversmoothing}. Rewiring methods to address the over-squashing problem focus on improving graph connectivity to ease the signal flow. However, one limitation of the rewiring method is that adding too many edges potentially leads to the over-smoothing issue~\cite{karhadkar2023fosr}. To explore this trade-off between over-squashing and over-smoothing, FoSR~\citep{karhadkar2023fosr} analyzes the change in Dirichlet energy upon adding edges. Their results demonstrate that rewiring the graph to enhance information propagation leads to the over-smoothing problem. 

\section{Proposed Method}\label{sec:method}
We first introduce our motivation and design rationale. We then propose our PANDA message-passing framework.

\subsection{Motivation and Design Rationale}
\paragraph{Sensitivity bound.}
Our design rationale is anchored in the insight derived from the sensitivity bound theorem \citep[Theorem 3.2]{giovanni2023oversquashing}.
Given the sensitivity bound in Eq.~\eqref{eq:sens}, we consider two scenarios: one with a standard width $p$ and the other with an increased width $p_{\mathrm{high}}$, where $p_{\mathrm{high}} > p$. 
For nodes with the increased width, the sensitivity bound is improved, demonstrating a potentially higher sensitivity to input features.
Regarding this, we provide a \Cref{proposition:dimension} in \Cref{app:proposition} 
and show empirical results well aligned with our argument in \Cref{sec:why}. We do not increase the width of all nodes, but \emph{selectively}.

\paragraph{Design rationale for selecting node centrality.}
We consider node centrality metrics to identify a node as a potential source of the bottleneck in a graph. 
Betweenness centrality has been considered as a significant indicator of the bottleneck, which measures the number of shortest paths going through a node~\citep{freeman1977betweenness}. The more frequently a node appears on the shortest paths of pairs of different nodes, the more likely it is to cause the bottleneck~\citep[Definition 9.]{topping2022riccurvature}. However, a hub-like node with inter-community edges also has eccentric properties that are different from those of the high betweenness centrality node, such as a high degree. 
We apply various centrality metrics in our framework to understand how each is related to the over-squashing problem. \Cref{app:centrality} describes the centralities we consider.

\subsection{Expanded Width-Aware Message Passing}
\begin{figure}[t]
    \centering
    \includegraphics[width=\columnwidth]{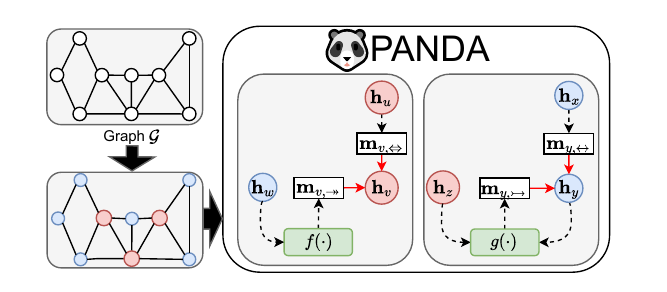}
    \vspace{-1em}
    \caption{Our proposed \textbf{PANDA} message passing framework. First, we selectively expand widths (i.e., hidden dimension sizes) according to a centrality in $\mathcal{G}$, and our \textbf{PANDA} message passing enables signal propagation among nodes with different widths (\mybluenode\; low-dimensional nodes and \myrednode\; high-dimensional nodes).}
    \label{fig:panda}
\end{figure}

\paragraph{Node expansion criteria.}
In our proposed method, we stratify nodes into two categories: low-dimensional nodes (\mybluenode) and high-dimensional nodes (\myrednode) (see \Cref{fig:panda}). This stratification is based on a specific centrality that quantifies the importance of each node.
The centrality, denoted as $C(\mathcal{G})$, can be one of the following: degree~\citep{borgatti2011analyzing}, betweenness~\citep{freeman1977betweenness}, closeness~\citep{freeman1978centralitycls}, PageRank~\citep{page1999pagerank}, or load centrality~\citep{goh2001universal}.
We represent the centrality values for all nodes as a vector $\mathbf{c}\in\mathbb{R}^{n}$ and then sort the elements of $\mathbf{c}$ in descending order.
Based on the sorted $\mathbf{c}$, we select the top-$k$ nodes, resulting in a binary mask vector, $\mathbf{b}\in\{0,1\}^{n}$: 
\begin{align}
    \mathbf{b}_{v} = \begin{cases}
    1 & \text{if } v \text{ is among the top } k \text{ nodes in } \mathbf{c}, \\
    0 & \text{otherwise}.
    \end{cases}
\end{align}
\paragraph{Define subsets of neighbourhoods.}
Based on a centrality measure, we define four subsets of neighbouring nodes.
For a node $v$ classified as low-dimensional, denoted by $\mathbf{b}_v = 0$:
\begin{align}
    \mathcal{N}_{\textrm{low} \leftrightarrow \textrm{low}}(v) &= \{ u \in \mathcal{N}(v) \mid \mathbf{b}_v = 0 \wedge \mathbf{b}_u = 0 \}, \\
    \mathcal{N}_{\textrm{high} \rightarrowtail \textrm{low}}(v) &= \{ u \in \mathcal{N}(v) \mid \mathbf{b}_v = 0 \wedge \mathbf{b}_u = 1 \}.
\end{align}

In addition, for a node $v$ classified as high-dimensional:
\begin{align}
    \mathcal{N}_{\textrm{high} \Leftrightarrow \textrm{high}}(v) &= \{ u \in \mathcal{N}(v) \mid \mathbf{b}_v=1 \wedge \mathbf{b}_u = 1 \}, \\
    \mathcal{N}_{\textrm{low} \twoheadrightarrow \textrm{high}}(v) &= \{ u \in \mathcal{N}(v) \mid \mathbf{b}_v = 1 \wedge \mathbf{b}_u = 0\}.
\end{align}

The comprehensive union of these subsets across all nodes in $\mathcal{V}$ encompasses the entire edge set $\mathcal{E}$, thereby preserving the original graph structure.

\paragraph{A new framework: PANDA message passing.}
The main contribution of our framework is that, given any MPNNs, we can make it work with different widths for nodes.
To this end, we replace the standard neighbour aggregator $ \psi^{(\ell)}$ with 4 different aggregators, each applied to the corresponding type of neighbours: $\psi^{(\ell)}_{\leftrightarrow}$, $\psi^{(\ell)}_{\Leftrightarrow}$, $\psi^{(\ell)}_{\twoheadrightarrow}$, and $\psi_{\rightarrowtail}^{(\ell)}$. Then, we use a common update function for all types. The final message passed from layer $\ell$ to layer $\ell+1$ is defined as follows:
\begin{align}
    \mathbf{m}^{(\ell)}_{v, \leftrightarrow} &=
    \psi^{(\ell)}_{\leftrightarrow}( \{ \mathbf{h}_u^{(\ell)} :u \in \mathcal{N}_{\textrm{low} \leftrightarrow \textrm{low}}(v) \}), \\
    \mathbf{m}^{(\ell)}_{v, \Leftrightarrow} &=
    \psi^{(\ell)}_{\Leftrightarrow}( \{ \mathbf{h}_u^{(\ell)} :u \in \mathcal{N}_{\textrm{high} \Leftrightarrow \textrm{high}}(v) \}), \\
    \mathbf{m}^{(\ell)}_{v, \twoheadrightarrow} &=
    \psi^{(\ell)}_{\twoheadrightarrow}( \{ f(\mathbf{h}_u^{(\ell)}) :u \in \mathcal{N}_{\textrm{low} \twoheadrightarrow \textrm{high}}(v) \}),\\
    \mathbf{m}^{(\ell)}_{v, \rightarrowtail} &=
    \psi_{\rightarrowtail}^{(\ell)}( \{ g(\mathbf{h}_v^{(\ell)},\mathbf{h}_u^{(\ell)}) :u \in \mathcal{N}_{\textrm{high} \rightarrowtail \textrm{low}}(v) \}), \\
    \mathbf{h}_{v}^{(\ell+1)} &= \phi (\mathbf{h}_{v}^{(\ell)}, \mathbf{m}^{(\ell)}_{v, \leftrightarrow}, \mathbf{m}^{(\ell)}_{v, \Leftrightarrow}, \mathbf{m}^{(\ell)}_{v, \twoheadrightarrow}, \mathbf{m}^{(\ell)}_{v, \rightarrowtail}).\label{eq:update}
\end{align}
Each aggregator $\psi^{(\ell)}_{\ast}$ is tailored to handle a specific type of interactions among nodes.
In Eq.~\eqref{eq:update}, the hidden vector of a node $v$ is updated from the four messages.
For instance, $\psi^{(\ell)}_{\twoheadrightarrow}(\cdot)$ and $\psi^{(\ell)}_{\rightarrowtail}(\cdot)$ are designed for interactions with different widths by using functions $f(\cdot)$ and $g(\cdot)$.

$f(\cdot)$ is a linear transformation that projects node features to the expanded dimensional space and is defined as:
\begin{align}
    f(\mathbf{h}_u^{(\ell)}) := \mathbf{W}_{f}^{(\ell)}\mathbf{h}^{(\ell)}_{u},\label{eq:f}
\end{align}
where $\mathbf{W}_{f}^{(\ell)} \in \mathbb{R}^{p_{\mathrm{high}}\times p}$.
For $\psi^{(\ell)}_{\rightarrowtail}(\cdot)$, we define $g(\cdot)$ as a dimension selector as follows:
\begin{align}
    g(\mathbf{h}_v^{(\ell)}, \mathbf{h}_u^{(\ell)}) &:= \mathrm{gather}\big( \mathbf{h}_{u}^{(\ell)}, \textrm{topk}(\mathbf{s}, p)\big),\label{eq:g} \\
    \mathbf{s} &= \softmax \left(\sigmoid \big(\eta(\mathbf{h}_{u}^{(\ell)} \oplus \mathbf{h}_{v}^{(\ell)}) \big)\right),
\end{align}
where $\eta$ is a non-linear neural network that computes a score vector $\mathbf{s}\in\mathbb{R}^{p_{\mathrm{high}}}$ that gives the relative significance of each dimension for each message. This score vector is then used to select the top $p$ dimensions for the message aggregation.

In our \textbf{PANDA}, it is important to emphasize that nodes with low and high-dimensional features are processed in their respective dimensional spaces throughout the network layers. This means that until the final layer, these nodes maintain their distinct dimensionalities.

\paragraph{Instance of our framework.}
To better understand our framework, we show how to integrate \textbf{PANDA} into two classical MPNNs: GCN~\citep{kipf2017GCN} and GIN~\citep{xu2018GIN}. We will use these variations for our experiments. For the PANDA-GCN variant, the layer-update is expressed as follows for low and high-dimensional nodes. First, if the node $v$ is low-dimensional:
\begin{align}\begin{split}
    \mathbf{h}^{(\ell+1)}_{v} &= \mathbf{h}^{(\ell)}_{v} + \sigma \Big( \sum_{\mathclap{ u \in \mathcal{N}_{\textrm{low} \leftrightarrow \textrm{low}}(v) }}\;\;\;\;\;\; \frac{1}{\sqrt{d_u d_{v}}} \mathbf{W}^{(\ell)}_{\text{low}} \mathbf{h}^{(\ell)}_{u} \\
    &+\sum_{\mathclap{ u \in \mathcal{N}_{\textrm{low} \twoheadrightarrow \textrm{high}}(v)}}\;\;\;\;\;\; \frac{1}{\sqrt{d_u d_{v}}} \mathbf{W}^{(\ell)}_{\text{low}} g(\mathbf{h}^{(\ell)}_{v}, \mathbf{h}^{(\ell)}_{u}) \Big),
    \end{split}\label{eq:panda-gcn-low}
\end{align}
then if node $v$ is high-dimensional:
\begin{align}\begin{split}
    \mathbf{h}^{(\ell+1)}_{v} &= \mathbf{h}^{(\ell)}_{v} + \sigma \Big( \sum_{\mathclap{ u \in \mathcal{N}_{\textrm{high} \Leftrightarrow \textrm{high}}(v)}}\;\;\;\;\;\;\; \frac{1}{\sqrt{d_{u} d_{v}}} \mathbf{W}^{(\ell)}_{\text{high}} \mathbf{h}^{(\ell)}_{u} \\
    &+ \sum_{\mathclap{ u \in \mathcal{N}_{\textrm{high} \rightarrowtail \textrm{low}}(v)}} \;\;\;\;\;\;\;\frac{1}{\sqrt{d_{u} d_{v}}} \mathbf{W}^{(\ell)}_{\text{high}} f(\mathbf{h}^{(\ell)}_{u}) \Big),
    \end{split}\label{eq:panda-gcn-high}
\end{align}
where $\sigma$ is a ReLU activation function, 
$\mathbf{W}^{(\ell)}_{\mathrm{low}}\in \mathbb{R}^{p \times p}$ and $\mathbf{W}^{(\ell)}_{\mathrm{high}}\in \mathbb{R}^{p_{\mathrm{high}} \times p_{\mathrm{high}}}$ are the weight matrices. 
The messages are normalized by their degrees, $d_v$ and $d_u$. 
For the PANDA-GIN variant, we describe the layer-update in \Cref{app:gin}.
\paragraph{Implementation.}
For implementation, the hidden vectors of different widths are separated and entered into the layer. These low and high-dimensional nodes are output with the same dimension in the last layer and integrated again.

\subsection{Properties of PANDA}
\paragraph{Directivity.}
Nodes with an equal width remain undirected, while edges among nodes with different widths can be directed. $\psi^{(\ell)}_{\twoheadrightarrow}$ and $\psi^{(\ell)}_{\rightarrowtail}$ have independent sets of parameters --- Eqs.~\eqref{eq:panda-gcn-low} and~\eqref{eq:panda-gcn-high} have different weight parameters for low-dimensional and high-dimensional nodes.
By ensuring the directivity only for edges with different widths, our method can be more expressive than its undirected counterpart~\citep[Theorems 4.1 and 4.2]{rossi2023dir}.

\paragraph{Relational graph.}
Our \textbf{PANDA} can be considered as a relational graph convolutional network (R-GCN)~\citep{schlichtkrull2018relational} applied to an augmented relational graph that incorporates two types of relation.
Not only is message propagation performed according to the relationship of the neighbour set, but also different weight matrices are used for low-dim nodes and high-dim nodes, as mentioned earlier. In this sense, our \textbf{PANDA} is similar to R-GCN. However, R-GCN is not a model designed to alleviate over-squashing.

\paragraph{Computational complexity.}
Compared to existing rewiring methods, \textbf{PANDA} adds complexity from centrality calculation and four different message passings instead of graph rewiring. The specific complexity depends on which algorithm is used to calculate the centrality. For example, the time complexity of degree centrality is $\mathcal{O}(|\mathcal{E}|)$, and for betweenness centrality it is $\mathcal{O}(|\mathcal{V}|^3)$. We will discuss the empirical runtime of \textbf{PANDA} in \Cref{app:runtime}.

\section{Discussions}\label{sec:why}
This section discusses why our \textbf{PANDA} alleviates over-squashing from empirical perspectives, i.e., feature sensitivity, signal propagation, and Dirichlet energy.

\begin{figure}[t]
    \centering
    \subfigure[Sensitivity w.r.t. $p$]{\includegraphics[width=0.49\columnwidth]{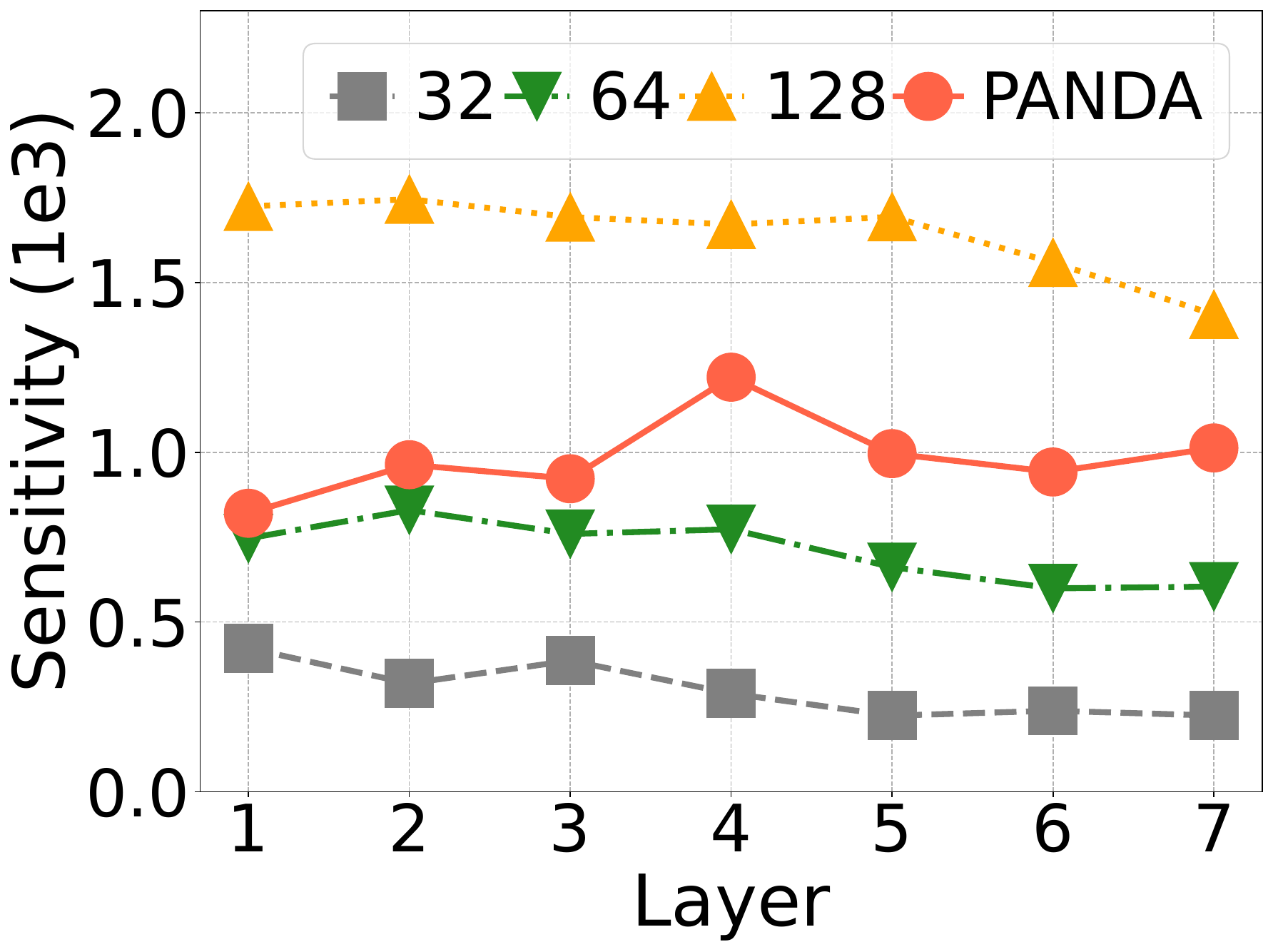}\label{fig:sens-a}}
    \subfigure[Sensitivity w.r.t. methods]{\includegraphics[width=0.49\columnwidth]{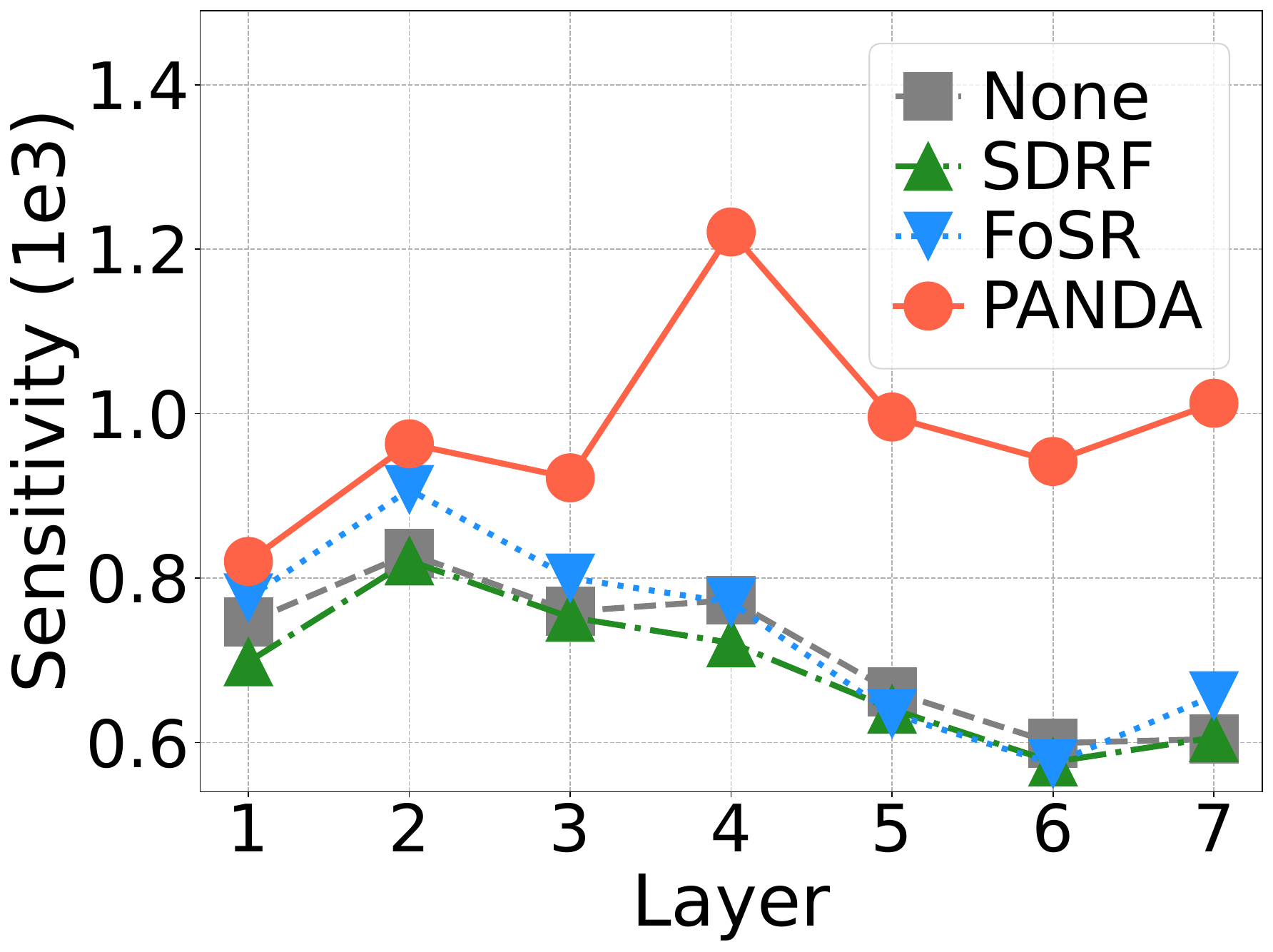}\label{fig:sens-b}}
    \caption{Empirical sensitivity across layers for GCN on \textsc{Mutag}. More results on other datasets are in \Cref{app:vis-why}.}
    \label{fig:sens}
\end{figure}

\paragraph{Empirical sensitivity analysis.}
In order to verify that our method actually improves the feature sensitivity among nodes, we analyze the sensitivity given by Eq.~\eqref{eq:sens} on benchmark datasets.
As shown in \Cref{fig:sens}, the sensitivity improves as the width $p$ increases. Our method benefits from selectively having 64 and 128-dimensional nodes, yielding sensitivity that lies between these two dimensions --- 16.7\% of nodes have 128 dimensions in \Cref{fig:sens}. Compared to other rewiring techniques, \textbf{PANDA} shows higher sensitivity that is maintained even in deeper layers, while others exhibit decreasing trends.

\paragraph{Signal propagation w.r.t. effective resistance.}
\citet{giovanni2023oversquashing} provide an empirical evidence that the information propagation of general GNNs is inversely proportional to the total effective resistance $R_{tot}$, which motivates us to check if our \textbf{PANDA} maintains the magnitude of signal flow in a graph with high $R_{tot}$. 
We further analyze if the signal flow improves after applying various rewiring methods. The detailed setting on the analysis is in \Cref{app:signalpropdetail}.

\begin{figure}[t!]
    \centering
    \subfigure[GCN on \textsc{Reddit-Binary}]{\includegraphics[width=0.49\columnwidth]{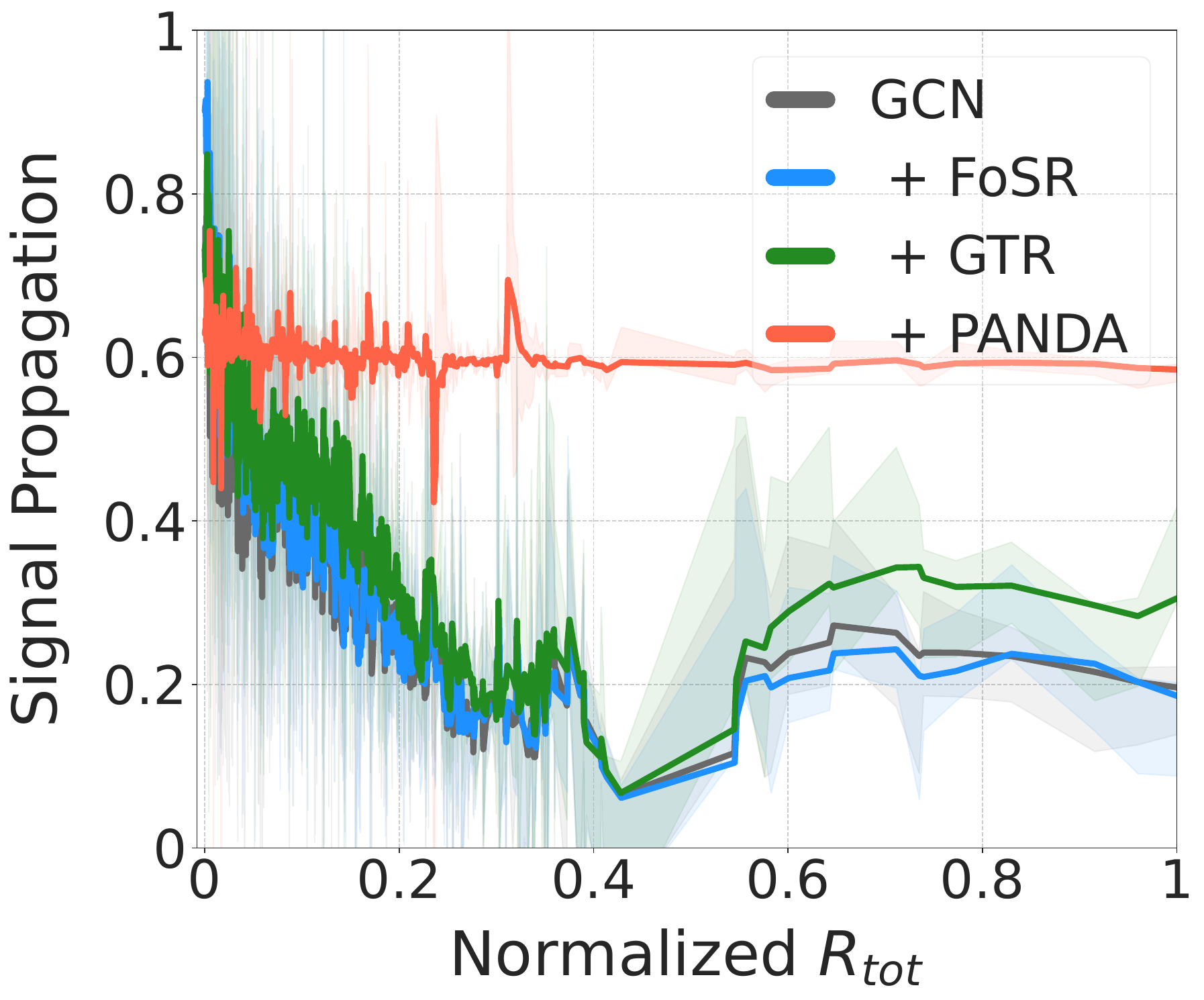}}
    \subfigure[GIN on \textsc{Reddit-Binary}]{\includegraphics[width=0.49\columnwidth]{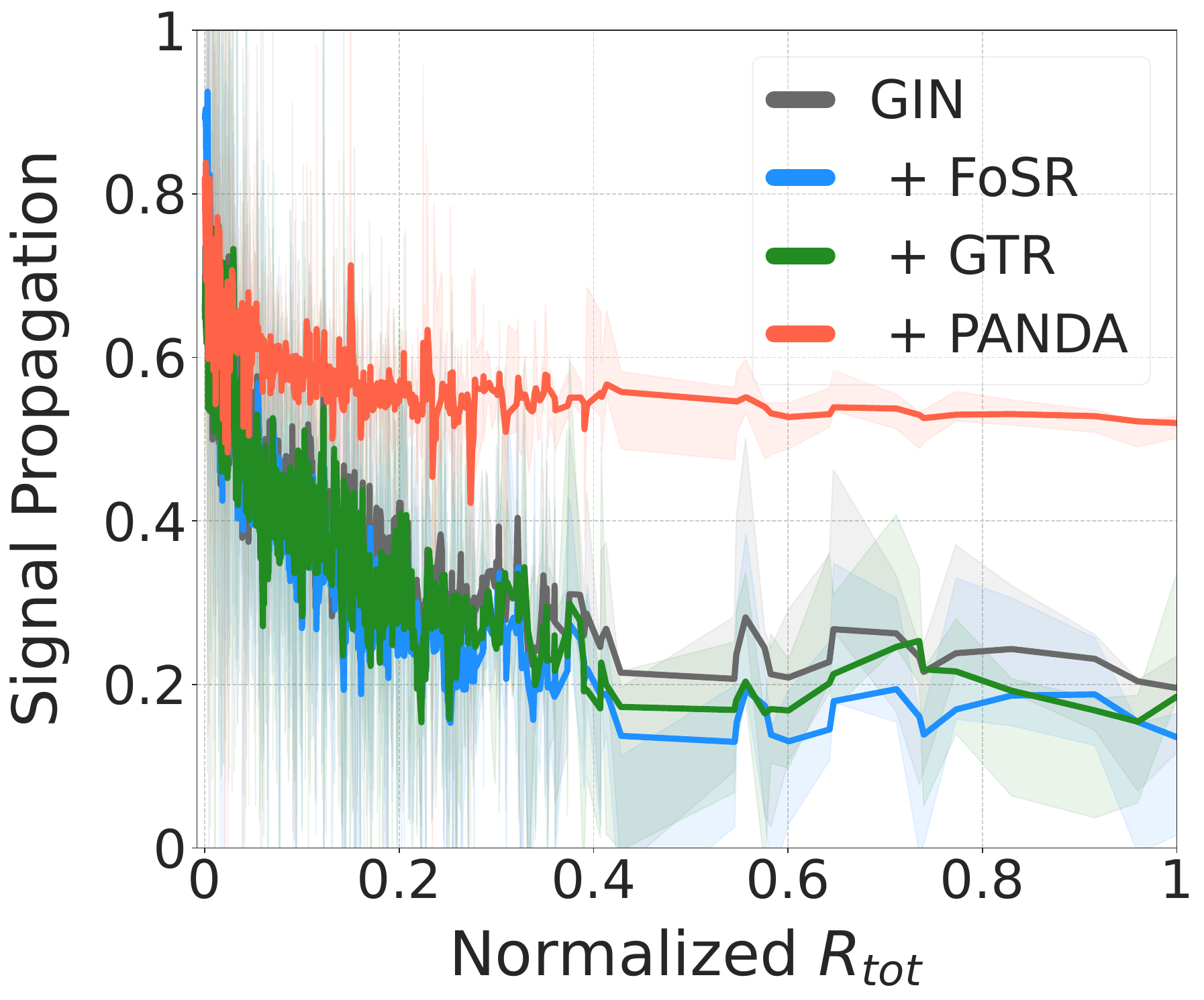}}
    \caption{The amount of signal propagated across the graph w.r.t. the normalized total effective resistance ($R_{tot}$). More results in other datasets are in \Cref{app:vis-why}.}
    \label{fig:signalweffres}
\end{figure}

In \Cref{fig:signalweffres}, all methods except \textbf{PANDA} present decaying signal propagation trends as the resistance of graphs increases. In contrast, \textbf{PANDA} shows consistent signal propagation even for graphs with higher effective resistance. 
To verify the linear relationship between total effective resistance and signal propagation, we also quantify the correlation coefficient between them in \Cref{fig:resistance}.
This result demonstrates the powerful effect of the width expansion in mitigating the over-squashing problem. Our \textbf{PANDA} maintains continuous information flows even under high bottleneck conditions, which cannot be overcome even by the direct modification of graph topology through various rewiring strategies.

\begin{figure}[t!]
    \centering
    \subfigure[\textsc{Proteins}]{\includegraphics[width=0.48\columnwidth]{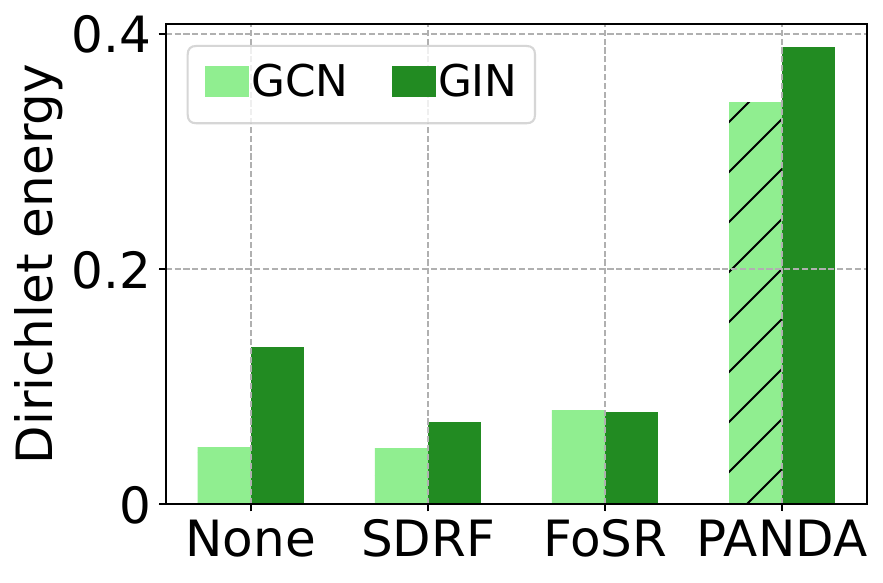}}
    \subfigure[\textsc{Mutag}]{\includegraphics[width=0.48\columnwidth]{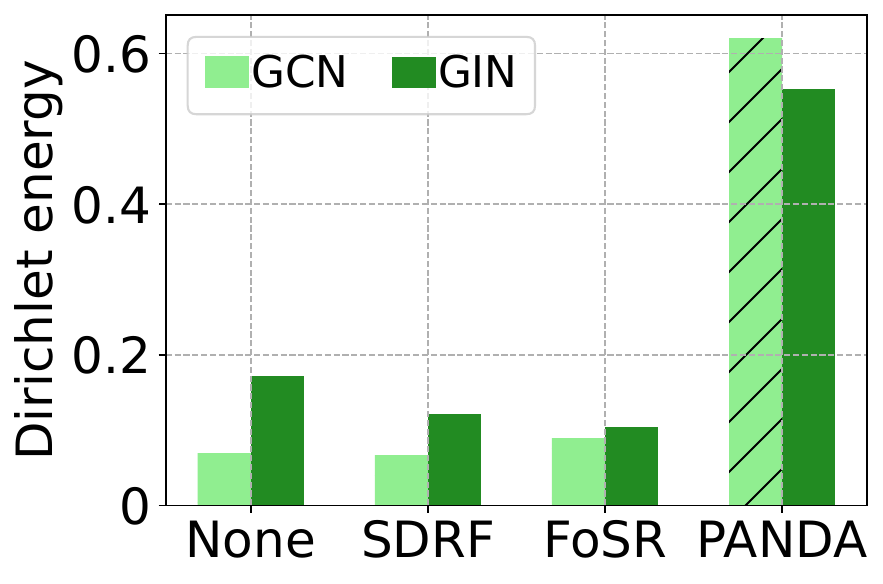}}
    \caption{Dirichlet energy of baselines and PANDA. More results in other datasets are in \Cref{app:vis-why}.}
    \label{fig:energy}
\end{figure}

\begin{table*}[ht!]
    \small
    \setlength{\tabcolsep}{3pt}
    \centering
    \caption{Results of PANDA and baselines for GCN and GIN. We show the best three in \BEST{red} (first), \SECOND{blue} (second), and \THIRD{purple} (third).}
    \label{tab:main}
    \begin{tabular}{l cccccc}\toprule
        Method & \textsc{Reddit-Binary} & \textsc{IMDB-Binary} & \textsc{Mutag} & \textsc{Enzymes} & \textsc{Proteins} & \textsc{Collab} \\ \midrule
        GCN (None) & 68.255\std{1.098} & 49.770\std{0.817} & 72.150\std{2.442} & \THIRD{27.667\std{1.164}} & 70.982\std{0.737} & 33.784\std{0.488} \\ 
        \;+ Last Layer FA & 68.485\std{0.945} & 48.980\std{0.945} & 70.050\std{2.027} & 26.467\std{1.204} & 71.018\std{0.963} & 33.320\std{0.435} \\ 
        \;+ Every Layer FA & 48.490\std{1.044} & 48.170\std{0.801} & 70.450\std{1.960} & 18.333\std{1.038} & 60.036\std{0.925} & \THIRD{51.798\std{0.419}} \\ 
        \;+ DIGL & 49.980\std{0.680} & 49.910\std{0.841} & 71.350\std{2.391} & 27.517\std{1.053} & 70.607\std{0.731} & 15.530\std{0.294} \\ 
        \;+ SDRF & \THIRD{68.620\std{0.851}} & 49.400\std{0.904} & 71.050\std{1.872} & \SECOND{28.367\std{1.174}} & 70.920\std{0.792} & 33.448\std{0.472} \\ 
        \;+ FoSR & \SECOND{70.330\std{0.727}} & 49.660\std{0.864} & \SECOND{80.000\std{1.574}} & 25.067\std{0.994} & \SECOND{73.420\std{0.811}} & 33.836\std{0.584} \\ 
        \;+ BORF & Time-out & \THIRD{50.100\std{0.900}} & 75.800\std{1.900} & 24.700\std{1.000} & 71.000\std{0.800} & Time-out \\ 
        \;+ GTR  & 68.990\std{0.610} & 49.920\std{0.990} & \THIRD{79.100\std{1.860}} & 27.520\std{0.990} & \THIRD{72.590\std{2.480}} & 33.050\std{0.400} \\
        \;+ CT-Layer & 51.580\std{1.019} & \SECOND{50.320\std{0.944}} & 75.899\std{3.024} & {17.383\std{1.030}} & 60.357\std{1.060} & \SECOND{52.146\std{0.415}} \\
        \cmidrule(lr){1-7}
        \;+ \textbf{PANDA} & \BEST{80.690\std{0.721}} & \BEST{63.760\std{1.012}} & \BEST{85.750\std{1.396}} & \BEST{31.550\std{1.230}} & \BEST{76.000\std{0.774}} & \BEST{68.400\std{0.452}} \\ 
        \midrule\midrule
        GIN (None) & 86.785\std{1.056} & 70.180\std{0.992} & 77.700\std{0.360} & 33.800\std{0.115} & 70.804\std{0.827} & 72.992\std{0.384} \\
        \;+ Last Layer FA & \SECOND{90.220\std{0.475}} & 70.910\std{0.788} & \SECOND{83.450\std{1.742}} & \BEST{47.400\std{1.387}} & 72.304\std{0.666} & \SECOND{75.056\std{0.406}} \\
        \;+ Every Layer FA & 50.360\std{0.684} & 49.160\std{0.870} & 72.550\std{3.016} & 28.383\std{1.052} & 70.375\std{0.910} & 32.984\std{0.390} \\
        \;+ DIGL & 76.035\std{0.774} & 64.390\std{0.907} & 79.700\std{2.150} & 35.717\std{1.198} & 70.759\std{0.774} & 54.504\std{0.410} \\
        \;+ SDRF & 86.440\std{0.590} & 69.720\std{1.152} & 78.400\std{2.803} & \THIRD{35.817\std{1.094}} & 69.813\std{0.792} & 72.958\std{0.419} \\
        \;+ FoSR & \THIRD{87.350\std{0.598}} & 71.210\std{0.919} & 78.400\std{2.803} & 29.200\std{1.367} & \SECOND{75.107\std{0.817}} & \THIRD{73.278\std{0.416}} \\
        \;+ BORF & Time-out & \SECOND{71.300\std{1.500}} & \THIRD{80.800\std{2.500}} & 35.500\std{1.200} & \THIRD{74.200\std{0.800}} & Time-out \\
        \;+ GTR  & 86.980\std{0.660} & \THIRD{71.280\std{0.860}} & 77.600\std{2.840} & 30.570\std{1.420} & 73.130\std{0.690} & 72.930\std{0.420} \\
        \;+ CT-Layer & 54.589\std{1.757} & 50.000\std{0.974} & 56.850\std{4.253} & {16.583\std{0.907}} & 61.107\std{1.184} & 52.304\std{0.605} \\
        \cmidrule(lr){1-7}
        \;+ \textbf{PANDA} & \BEST{91.055\std{0.402}} & \BEST{72.560\std{0.917}} & \BEST{88.750\std{1.570}} & \SECOND{46.200\std{1.410}} & \BEST{75.759\std{0.856}} & \BEST{75.110\std{0.210}} \\ \bottomrule
    \end{tabular}
\end{table*}

\begin{table*}[th!]
    \small
    \setlength{\tabcolsep}{3pt}
    \centering
    \caption{PANDA-GCN v.s. R-GCN}
    \label{tab:relational}
    \begin{tabular}{l cccccc}\toprule
        Method & \textsc{Reddit-Binary} & \textsc{IMDB-Binary} & \textsc{Mutag} & \textsc{Enzymes} & \textsc{Proteins} & \textsc{Collab} \\ \midrule
        R-GCN & 49.850\std{0.653} & 50.012\std{0.917} & 69.250\std{2.085} & 28.600\std{1.186} & 69.518\std{0.725} & 33.602\std{1.047} \\
        \textbf{PANDA-GCN} & \textbf{80.690\std{0.721}} & \textbf{63.760\std{1.012}} & \textbf{85.750\std{1.396}} & \textbf{31.550\std{1.230}} & \textbf{76.000\std{0.774}} & \textbf{68.400\std{0.452}}\\ \midrule
    \end{tabular}
\end{table*}

\paragraph{Trade-off between rewiring and over-smoothing.}
As mentioned in \Cref{pre:tradeoff}, graph rewiring can cause over-smoothing. 
Since \textbf{PANDA} does not modify the original graph connectivity, we analyze whether our method can avoid the over-smoothing problem induced by graph rewiring using the Dirichlet energy as a measure of over-smoothing. The theoretical relation between Dirichelt energy and over-smoohting is given in \Cref{app:dirismoothing}. \Cref{fig:energy} compares the Dirichlet energies of the final hidden representations learned from GCN and GIN having both a rewiring technique and \textbf{PANDA}. The result shows that \textbf{PANDA} yields a higher Dirichlet energy for both GCN and GIN compared to others, highlighting the strength of our model, mitigating the over-squashing while preserving the original graph connectivity.

\section{Experiments}\label{sec:exp}
In this section, we empirically verify that \textbf{PANDA} can significantly improve the performance of GNN over other rewiring methods.
We experiment with graph classification and node classification, as well as tasks on Long-Range Graph Benchmark (LRGB)~\citep{dwivedi2022LRGB}. We cover the experiments on TUDataset~\citep{Morris2020TUDataset} in the main content and the node classification task and LRGB dataset in \Cref{app:node} and \Cref{app:lrgb}.

\paragraph{Datasets.}
For graph classification, we report our method using the following benchmarks: \textsc{Reddit-Binary}, \textsc{IMDB-Binary}, \textsc{Mutag}, \textsc{Enzymes}, \textsc{Proteins}, \textsc{Collab} from the TUDataset~\citep{Morris2020TUDataset}. 
A detail of datasets is available in \Cref{app:dataset}.

\paragraph{Experimental setting.}
For graph classification, we compare \textbf{PANDA} to no graph rewiring and 7 other state-of-the-art rewiring methods: DIGL~\citep{gasteiger2019digl}, Fully adjacent layer (FA)~\citep{alon2021oversquashing}, SDRF~\citep{topping2022riccurvature}, FoSR~\citep{karhadkar2023fosr}, BORF~\citep{karhadkar2023fosr}, GTR~\citep{black2023gtr}, and CT-Layer~\citep{arnaiz2022diffwire}.
In our experiments, we prioritize fairness and comprehensiveness rather than aiming to obtain the best possible performance for each dataset.
For backbone GNNs, we use GCN, GIN~\cite{xu2018GIN}, R-GCN~\cite{schlichtkrull2018relational}, and R-GIN~\cite{brockschmidt2020gnn}. 
We accumulate the result in 100 random trials and report the mean test accuracy, along with the 95\% confidence interval.
Further experiment details are in \Cref{app:exp_detail}.

\begin{table*}
    \small
    \setlength{\tabcolsep}{3pt}
    \centering
    \caption{Performance comparison by various centrality measures for PANDA-GCN. The results of PANDA-GIN are in \Cref{app:centrality-result}.}
    \label{tab:centrality-gcn}
    \begin{tabular}{l cccccc}\toprule
        $C(\mathcal{G})$ & \textsc{Reddit-Binary} & \textsc{IMDB-Binary} & \textsc{Mutag} & \textsc{Enzymes} & \textsc{Proteins} & \textsc{Collab} \\ \cmidrule(lr){1-7}
        Degree & \BEST{80.690\std{0.721}} & \THIRD{62.100\std{1.043}} & 85.200\std{1.568} & \SECOND{31.117\std{1.258}} & \SECOND{75.375\std{0.800}} & \SECOND{68.162\std{0.471}}\\
        Betweenness & \THIRD{80.000\std{0.659}} & 59.630\std{1.152} & \BEST{85.750\std{1.396}} & 29.600\std{1.208} & 74.589\std{0.791} & \THIRD{67.844\std{0.547}} \\
        Closeness & 79.700\std{0.664} & \SECOND{61.160\std{0.992}} & 84.700\std{1.554} & \THIRD{29.967\std{1.231}} & \BEST{76.000\std{0.774}} & \BEST{68.400\std{0.452}} \\ 
        PageRank & \SECOND{80.340\std{0.826}} & \BEST{63.760\std{1.012}} & \THIRD{85.450\std{1.569}} & \BEST{31.550\std{1.230}} & 74.098\std{0.851} & 67.540\std{0.500} \\  
        Load & 79.500\std{0.732} & 59.840\std{1.153} & \SECOND{85.700\std{1.549}} & 28.167\std{1.090} & \THIRD{74.188\std{0.814}} & 67.802\std{0.506} \\ \bottomrule
    \end{tabular}
\end{table*}

\paragraph{Results.}
\Cref{tab:main} shows the results of different methods applied to the GCN and GIN models across benchmark datasets. 
Our \textbf{PANDA} method shows the highest accuracy across most of the datasets for both models, significantly outperforming the baseline (``None'') and other rewiring methods. 
Surprisingly, For \textsc{Reddit-Binary}, our \textbf{PANDA} shows a substantial 22.30\% improvement over the baseline (None) method.
In particular, in the case of BORF, it is impracticable to evaluate for \textsc{Reddit-Binary} and \textsc{Collab} due to the rewiring time.
In the case of GIN, our \textbf{PANDA} also leads to the highest accuracies, with 91.055 on \textsc{Reddit-Binary} and 88.75 on \textsc{Mutag}. These results underscore the effectiveness of our \textbf{PANDA} in enhancing the performance of GNN.
In \Cref{tab:main-rgnn} of \Cref{app:relational}, we also show the experimental results for R-GCN and R-GIN.

\paragraph{Performance comparison of PANDA-GCN and R-GCN.}
Considering the similarities between PANDA-GCN and R-GCN discussed in the previous section, we empirically evaluate their performance. 
As shown in \Cref{tab:relational}, PANDA-GCN outperforms R-GCN in all datasets.
Compared to R-GCN, which is not designed to mitigate over-squashing, these results show the efficacy of \textbf{PANDA}.

\paragraph{Effect of centrality metrics.}
It is natural to ask how different the performance of \textbf{PANDA} varies depending on the centrality metrics.
\Cref{tab:centrality-gcn} compares the results obtained using different kinds of centrality metrics.
In \textsc{IMDB-Binary}, PageRank centrality is better than other metrics, indicating that larger node widths are effective in receiving messages from influential nodes. 
Closeness centrality shows the best performance on \textsc{Proteins} and \textsc{Collab}.
This indicates that it is effective to expand the width of nodes that require many connections to connect to other distant nodes.

\begin{figure}[t]
    \centering
    \subfigure[\textsc{Proteins}]{\includegraphics[width=0.48\columnwidth]{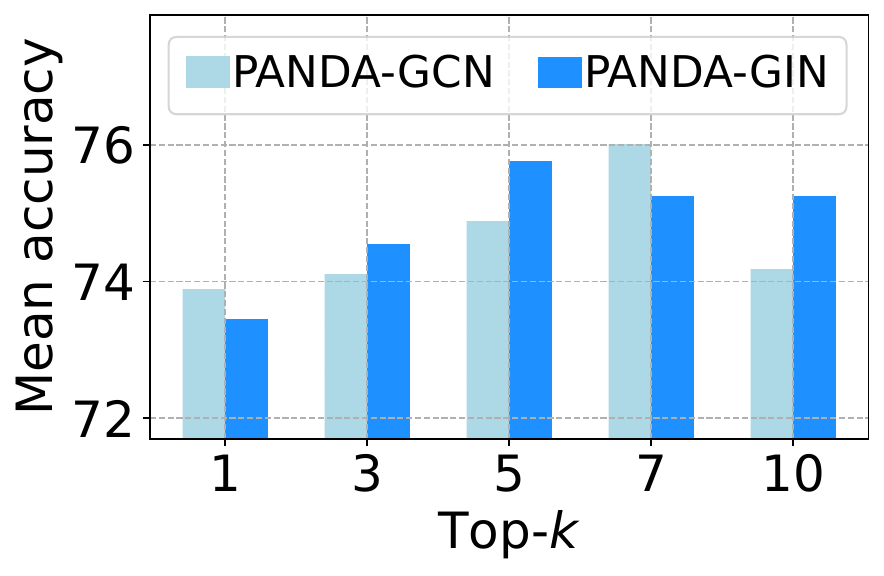}}
    \subfigure[\textsc{Mutag}]{\includegraphics[width=0.48\columnwidth]{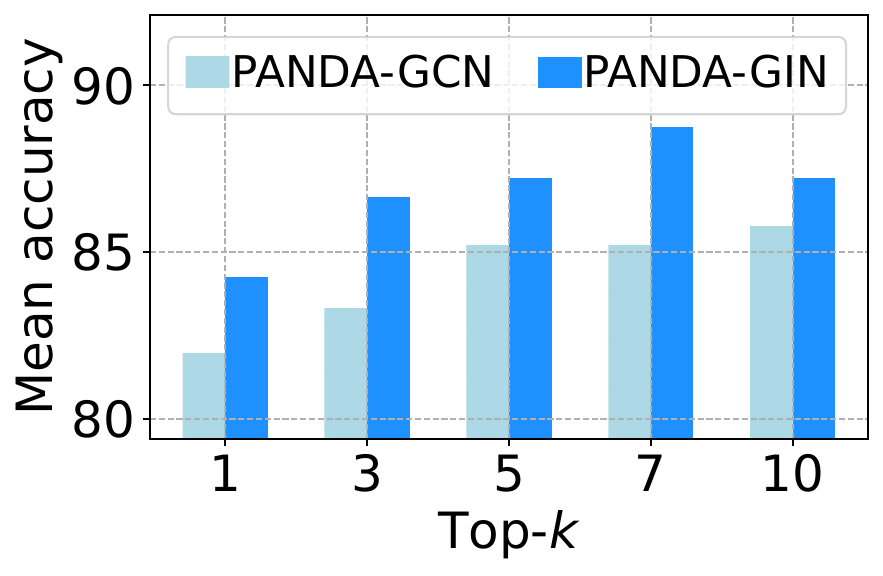}}
    \caption{Sensitivity on top-$k$. More results are in \Cref{app:vis-exp}.}
    \label{fig:topk}
\end{figure}

\paragraph{Sensitivity on top-$k$ nodes.}
\Cref{fig:topk} shows a sensitivity study w.r.t. top-$k$ nodes.
For \textsc{Proteins}, both PANDA-GCN and PANDA-GIN show an increase in mean accuracy as the top-$k$ value increased from 3 to 7. However, the accuracy does not increase after the $k$ value is 7.
For \textsc{Mutag}, PANDA-GCN and PANDA-GIN have the highest accuracy for $k$ values of 10 and 7, respectively.

\begin{figure}[t]
    \centering
    \subfigure[\textsc{Proteins}]{\includegraphics[width=0.48\columnwidth]{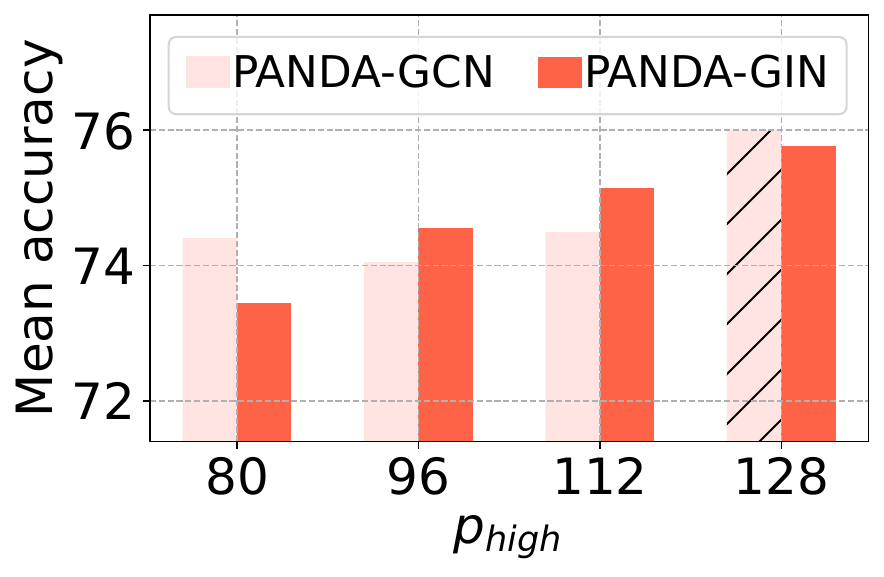}}
    \subfigure[\textsc{Mutag}]{\includegraphics[width=0.48\columnwidth]{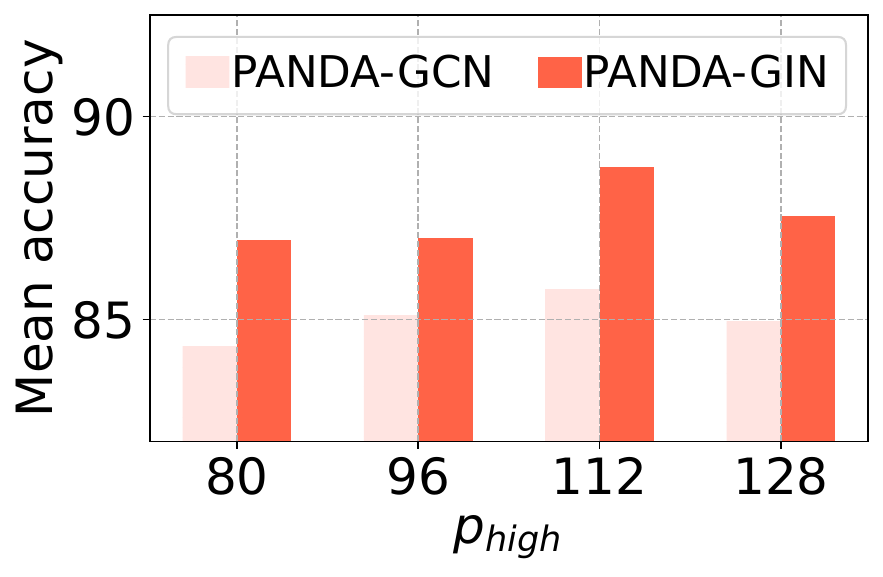}}
    \caption{Sensitivity on $p_{\mathrm{high}}$. More results are in \Cref{app:vis-exp}.}
    \label{fig:p_high}
\end{figure}

\paragraph{Sensitivity on $p_{\mathrm{high}}$.}
\Cref{fig:p_high} shows a sensitivity study of the \textbf{PANDA} w.r.t. $p_{\mathrm{high}}$.
In \textsc{Proteins}, using PANDA-GCN and PANDA-GIN with larger $p_{\mathrm{high}}$ improves performance. In \textsc{Mutag}, on the other hand, both models achieve their best performance when using a $p_{\mathrm{high}}$ value of 112.

\section{Related Work}\label{sec:related_work}
This section introduces related work to alleviate the over-squashing problem. Before the link between over-squashing and graph curvature was established by \citep{topping2022riccurvature}, DIGL was developed to sparsify the adjacency matrix using the graph heat kernel and personalized PageRank~\citep{gasteiger2019digl}. 
After \citet{alon2021oversquashing} empirically observed the over-squashing problem, research is active to find indicators for over-squashing and propose rewiring methods.
\citet{topping2022riccurvature} initially connect over-squashing with graph Ricci curvature, showing that edges with highly negative Ricci curvature contribute to the over-squashing and proposes an SDRF.
Since then, rewiring methods with indicators based on curvatures have been studied~\citep{nguyen2023borf,shi2023curvature}.
In addition to using curvatures as indicators of the over-squashing, \citet{black2023gtr} measure the over-squashing through the concept of effective resistance.
\citet{banerjee2022rlef} propose rewiring methods to handle the original topology of the graph to prevent it from being disconnected~\citep{banerjee2022rlef}.
To prevent over-smoothing, \citet{karhadkar2023fosr} propose a rewiring method that optimizes the spectral gap.
\citet{deac2022expander} propose to interleave message propagation on the original graph with message passing on an expander graph to alleviate over-squashing.
\citet{gutteridge2023drew} propose a layer-dependent rewiring method that provides a variety of rewiring graphs for feature propagation.
\citet{barbero2023laser} propose a LASER to rewire the graph to better preserve locality.
\citet{finkelshtein2023cooperative} propose a CO-GNN with flexible and dynamic message passing that can perform effective graph rewiring at each layer of GNN.
Recently, \citet{giovanni2023oversquashing} analyze the factors that contribute to over-squashing, and \citet{southern2024understanding} analyze the role of virtual nodes in over-squashing.
Also, rewiring methods, as well as more advanced and flexible message-passing paradigms, are being studied~\cite{errica2023adaptive,barbero2023laser,park2023nonbacktracking,qian2024probabilistically,behrouz2024graphmamba,qiu2024gump}.

Most studies to mitigate over-squashing focus on rewiring methods that change the topology. However, PANDA is different in that it addresses the over-squashing problem by moving away from the rewiring method.

\section{Concluding Remarks}
We presented \textbf{PANDA}, a novel message passing framework, effectively addressing the over-squashing problem in GNNs without rewiring edges. By selectively expanding the width of high-centrality nodes, \textbf{PANDA} promotes signal propagation to alleviate over-squashing. Our empirical evaluations show that \textbf{PANDA} outperforms existing graph rewiring methods in graph classification. This research contributes a pioneering approach to selectively expanding node widths in message passing and lays the groundwork for the future exploration of width-aware strategies in GNNs. 
\paragraph{Limitations and future work.} We only focus on redesigning the GNN based on sensitivity (Eq.~\eqref{eq:sens}) for the width of the nodes that contribute to over-squashing.
Despite our empirical evidence for mitigating over-squashing, more research is needed to prove the effects of higher widths on over-squashing.
For future work, we will aim to reduce the complexity of our method and explore a much wider range of tasks to study the pros and cons of higher widths.

\clearpage

\section*{Impact Statement}
Our proposed \textbf{PANDA} is designed as a general graph representation learning method. Therefore, our proposed method has no discernible negative impact on society. Nonetheless, adverse or malicious applications of the proposed algorithms in various domains, including drug discovery and healthcare, may lead to undesirable effects.
\section*{Acknowledgements}
This work was partly supported by Samsung Electronics Co., Ltd. (No. G01240136, KAIST Semiconductor Research Fund (2nd), 5\%), the Korea Advanced Institute of Science and Technology (KAIST) grant funded by the Korea government (MSIT) (No. G04240001, Physics-inspired Deep Learning, 5\%), Institute for Information \& Communications Technology Planning \& Evaluation (IITP) grants funded by the Korea government (MSIT) (No. RS-2020-II201361, Artificial Intelligence Graduate School Program (Yonsei University), 10\%), and an ETRI grant funded by the Korean government (No. 24ZB1100, Core Technology Research for Self-Improving Integrated Artificial Intelligence System, 80\%).

\bibliography{reference}
\bibliographystyle{icml2024}

\newpage
\appendix
\onecolumn
{\Large \textbf{Supplementary Materials for ``PANDA: Expanded Width-Aware Message Passing Beyond Rewiring''}}

\section{Effective Resistance and Signal Propagation}\label{app:effresdef}
\subsection{Effective Resistance}
\paragraph{Total effective resistance.}
The resistance between the nodes $u$ and $v$ in the graph is given by
\begin{align}
    R_{u,v}  =  (\mathbf{1}_u - \mathbf{1}_v)^{\mathsf{T}} \mathbf{L}^{+} (\mathbf{1}_u - \mathbf{1}_v),
\end{align}
where $\mathbf{1}_v$ and $\mathbf{1}_u$ are indicator vectors for node $u$ and $v$, respectively. Total effective resistance, $R_{tot}$, is defined as the sum of effective resistance between all pairs of nodes~\citep{ghosh2008minimizing,black2023gtr}:
\begin{align}
    R_{tot} \,=\, \sum_{u>v} R_{u,v} \,=\, n\cdot \mathrm{Tr}(\mathbf{L}^{+}) \,=\, n\sum_{i}^{n}\frac{1}{\lambda_{i}},\label{eq:r_total}
\end{align}
where $\lambda_{i}$ is the $i$-th eigenvalues of $\mathbf{L}$ and $\mathbf{L}^{+}$ is the pseudoinverse of $\mathbf{L}$.

\subsection{Signal Propagation}\label{app:signalpropdetail}
\paragraph{Signal propagation w.r.t. effective resistance.}
Here, we provide experimental details for measuring signal propagation w.r.t. normalized total effective resistance of the graphs. First, fix a source node $v\in \mathcal{V}$ and assign $p$-dimensional feature vector to it, while assign the rest of the nodes zero vectors. 
Then, the amount of signal that has been propagated over the graph by the randomly initialized model with $\ell$ layers is given by
\begin{align}
    \mathbf{h}_{\odot}^{(\ell)} = \frac{1}{p \max_{u\neq v} k_{\mathcal{G}}(u, v)}  \sum_{t=1}^{p} \sum_{u\neq v} \frac{\mathbf{h}_{u}^{(\ell),t}}{\|\mathbf{h}_{u}^{(\ell),t}\|} k_{\mathcal{G}}(u, v),
\end{align} 
where $\mathbf{h}_{u}^{(\ell), t}$ is the $t$-th feature of $p$-dimensional feature vector of node $u$ at layer $\ell$ and $k_{\mathcal{G}}(u, v)$ is the distance between two nodes $u$ and $v$, computed as a shortest path.
Every unitary signal $\mathbf{h}_{u}^{(\ell), t} / \| \mathbf{h}_{u}^{(\ell),t}\|$ propagated across the graph $G$ from the source node $v$ is weighted by the normalized propagation distance $k_{\mathcal{G}}(u, v) / \max_{u\neq v}d_{G}(u, v)$ for all nodes $u \neq v$ and then averaged over entire $p$ output channels. 10 nodes are randomly sampled from each graph and total effective resistance of the graph is estimated for each source node. The final $\mathbf{h}_{\odot}^{(\ell)}$ and total resistance of the graph are calculated as the mean of all 10 nodes. The experiment is repeated for every graph in the dataset and the signal propagation measured for each graph is plotted w.r.t. the normalized total effective resistance of the corresponding graph.

\section{Proposition on Enhanced Sensitivity to Alleviate Over-Squashing}\label{app:proposition}
We provide a proposition based on Eq.~\eqref{eq:sens} that the introduction can alleviate the over-squashing.

\begin{proposition}[Informal]\label{proposition:dimension}
Given the sensitivity bound in Eq.~\eqref{eq:sens}, we consider two scenarios: one with a standard width $p$ and the other with an increased width $p_{\mathrm{high}}$, where $p_{\mathrm{high}} > p$. 
For nodes with increased width, the sensitivity bound is augmented, demonstrating a potentially higher sensitivity to input features:
\begin{align}
\left \| \frac{\partial \mathbf{h}_v^{(\ell)}}{\partial \mathbf{h}_u^{(0)}} \right \|_1 \leq (zwp)^{\ell} \left(\mathbf{S}^{\ell}\right)_{vu} < (zwp_{\mathrm{high}})^{\ell} \left(\mathbf{S}^{\ell}\right)_{vu}.\label{eq:sens-high}
\end{align} 
This increased sensitivity can potentially reduce over-squashing, facilitating better signal propagation.
\end{proposition}
\Cref{proposition:dimension} provides the theoretical foundation for our approach. We are inspired by the insights gained from increased upper bound with increased width, but we emphasize that we do not increase the width of all nodes, but rather \emph{selectively}.

\section{Graph Centrality Metrics}\label{app:centrality}

We next introduce five graph centrality metrics:
\begin{itemize}
    \item Degree centrality~\citep{borgatti2011analyzing} measures the number of direct connections a node has to other nodes in the network, e.g., the degree of nodes. The centrality value of node v is the fraction of nodes connected to it. This centrality measures a local characteristic and does not take into account global topology or neighbor connectivity.
    \item PageRank centrality~\citep{page1999pagerank} measures the importance of a node based on the importance of its neighbors.
    It is an iterative algorithm that assigns a node a high score if it is connected to many nodes that themselves have high scores. High PageRank centrality indicates that a node is not only connected to many nodes but also to nodes that are themselves central or important within the network.
    This centrality uses a damping factor to control neighbors' effect on your node while measuring its importance. We set the damping factor as 0.85.
    \item Closeness centrality~\citep{freeman1978centralitycls} is a metric that measures how close a particular node is to all other nodes within a network. This metric is calculated based on the shortest path length between nodes in a network and indicates how efficiently a node is connected to other nodes in the network on average. 
    Closeness centrality of a node $v$ is the reciprocal of the average shortest path distance to $u$ over all $n-1$ reachable nodes.
    Nodes with high closeness centrality are located relatively close to all other nodes in the network.
    Because a node with high closeness centrality plays a central role through the shortest paths to many other nodes, the amount of information focused on this node can be very large. 
    \item Betweenness centrality~\citep{freeman1977betweenness} measures the extent to which a node lies on the shortest paths between other nodes. It highlights nodes that serve as bridges or points of control within the network. A node with high betweenness centrality significantly influences the flow of information or resources within the network, as it can affect the transfer between other nodes by facilitating or constraining it. For this reason, in studies that aim to measure over-squashing~\cite{topping2022riccurvature}, it is also used to measure the bottleneck of the graph.
    The centrality score of a node is calculated by the number of shortest paths that include the node.
    However, this metric is notorious for its high complexity to calculate all-pair shortest paths.
    \item Load centrality~\citep{goh2001universal}  is similar to betweenness centrality but considers the number of times a node is traversed by all shortest paths in the network. It is a measure of the load or traffic that a node will handle. Nodes with high load centrality are critical for the flow of information or resources, as they are likely to be bottlenecks or critical points of failure within the network. Like other centralities, it is useful for finding nodes where bottlenecks occur.
\end{itemize}

\section{Layer-update of PANDA-GIN}\label{app:gin}
In this section, we define the layer-update of PANDA-GIN. If the node $v$ is low-dimensional:
\begin{align}
\begin{split}
    \mathbf{h}^{(\ell+1)}_{v} &= \mathrm{MLP}^{(\ell)}_{\text{low}} \Big( (1+\epsilon) \mathbf{h}^{(\ell)}_{v} + \sum_{\mathclap{u\in\mathcal{N}_{\textrm{low} \leftrightarrow \textrm{low}} (v)}} \; \mathbf{h}^{(\ell)}_{u} + \;\;\;\;\; \sum_{\mathclap{u\in\mathcal{N}_{\textrm{high} \rightarrowtail \textrm{low}} (v)}}\; g(\mathbf{h}^{(\ell)}_{v}, \mathbf{h}^{(\ell)}_{u}) \Big),
\end{split}
\end{align}
where $\mathrm{MLP}^{(\ell)}_{\textrm{low}}$ is one or more linear layers separated by ReLU activation and $\epsilon$ is a weight parameter.
For the high-dimensional nodes, the layer-update is defined as follows:
\begin{align}
\begin{split}
    \mathbf{h}^{(\ell+1)}_{v} &= \mathrm{MLP}^{(\ell)}_{\text{high}} \Big( (1+\epsilon) \mathbf{h}^{(\ell)}_{v} + \sum_{\mathclap{u\in\mathcal{N}_{\textrm{high} \Leftrightarrow \textrm{high}} (v)}}\; \mathbf{h}^{(\ell)}_{u} + \;\;\;\;\; \sum_{\mathclap{u\in\mathcal{N}_{\textrm{low} \twoheadrightarrow \textrm{high}} (v)}}\; f(\mathbf{h}^{(\ell)}_{u}) \Big).
\end{split}
\end{align}

\section{Dirichlet Energy and Over-smoothing}\label{app:dirismoothing}
Let $\mathcal{G}$ be a graph with adjacency matrix $\mathbf{A}$ and normalized Laplacian $\widetilde{\mathbf{L}}=\mathbf{I}-\mathbf{D}^{-1/2}\mathbf{A}\mathbf{D}^{-1/2}$. Given a vector field $\mathbf{H}\,\in\, \mathbb{R}^{n \times p}$ defined on nodes in $\mathcal{G}$, Dirichlet energy is defined as 
\begin{align}
    E_{\mathrm{Dir}}(\mathbf{H}) := \mathrm{Tr}(\mathbf{H}^{\mathsf{T}}\widetilde{\mathbf{L}}\mathbf{H}) = \frac{1}{2}\sum_{v,u,w}\mathbf{A}_{v,u}{\biggl(\frac{\mathbf{H}_{v,u}}{\sqrt{d_{v}}} - \frac{\mathbf{H}_{u,w}}{\sqrt{d_{u}}}\biggr)}^{2}
\end{align} 
 
Essentially, the Dirichlet energy quantifies how much a function on the graph deviates from being constant between connected node pairs, which thereby indicates how non-smooth the signals on a graph are~\citep{chung1997spectgraph, arnaiz2022diffwire}. Hence a metric of Dirichlet energy has been widely applied to measure the amount of over-smoothing of the graph representations.

\section{Experimental Details for Graph Classification}\label{app:exp_detail}
\subsection{Dataset Statisctics}\label{app:dataset}
\paragraph{General statistics for graph classification datasets.}
We consider the \textsc{Reddit-Binary} (2,000 graphs), \textsc{IMDB-BINARY} (1,000 graphs), \textsc{Mutag} (188 graphs), \textsc{Enzymes} (600 graphs), \textsc{Proteins} (1,113 graphs), and \textsc{Collab} (5.000 graphs) tasks from TUDatasets~\citep{Morris2020TUDataset}. These datasets were chosen by~\citet{karhadkar2023fosr}, under the claim that they require long-range interactions. 
\subsection{Detail of Baselines}\label{app:baselines}
We compare our method with the following baselines and try to focus on baselines that preprocess the input graph by adding edges.
\paragraph{DIGL.} Diffusion Improves Graph Learning (DIGL)~\citep{gasteiger2019digl} is a diffusion-based rewiring scheme that computes a kernel evaluation of the adjacency matrix, followed by sparsification. 
\paragraph{SDRF.} Stochastic Discrete Ricci Flow (SDRF)~\citep{topping2022riccurvature} surgically rewires a graph by adding edges to support other edges with low curvature, which are locations of bottlenecks. For SDRF, we include results for both the original method (configured to only add edges), and our relational method (again only add edges and include the added edges with their own relation). 
\paragraph{FA.} Fully adjacent (FA) layers~\citep{alon2021oversquashing} rewire the graph by adding all possible edges. We include results for rewiring only the last layer (last layer FA) and rewiring every layer (every layer FA).
\paragraph{FOSR.} First-order Spectral Rewiring (FoSR)~\citep{karhadkar2023fosr} is a graph rewiring method for preventing over-squashing based on iterative first-order maximization of the spectral gap.
\paragraph{BORF.} Batch Ollivier-Ricci Flow (BORF)~\citep{karhadkar2023fosr} is a novel curvature-based rewiring method designed to mitigate the over-smoothing and over-squashing issues simultaneously.
\paragraph{GTR.} Greedy Total Resistance (GTR)~\citep{black2023gtr} is a rewiring method based on total effective resistance.
\paragraph{CT-Layer.} Commute Time (CT)-Layer~\citep{arnaiz2022diffwire} is a method that learns the commute time and rewires the input graph accordingly.
In case of CT-Layer, we isolated the CT-Layer that consists of CT pooling and CT rewiring steps from the DiffWire~\citep{arnaiz2022diffwire} where the input graph is rewired and re-weighted based on the learnable commute time embeddings. Then we subsequently add series of GCN and GIN layers identical to the ones used in other rewiring methods and \textbf{PANDA}.   

\subsection{Evaluation protocol}\label{app:evaluation}
In order to faithfully compare the performance of the rewiring techniques, we follow the same setting as in~\citet{karhadkar2023fosr}.
Each configuration is evaluated using the validation set. 
The testset accuracy of the configuration with the best validation performance is then recorded. 
For each experiment, we accumulate the result in 100 random trials with a 80\%/10\%/10\% train/val/test split and report the mean test accuracy, along with the 95\% confidence interval.

\subsection{Hyperparameters}\label{app:hyperparam}
For each task and baseline model, we used the same settings of GNN and optimization hyperparameters across all methods to rule out hyperparameter tuning as a source of performance gain.
\Cref{tab:param-common} shows common hyperparameters. \Cref{tab:param-panda} shows the search range for hyperparameters of \textbf{PANDA}, and \Cref{tab:best-param} shows the best hyperparameters used by \textbf{PANDA}.

\begin{table}[h]
    \small
    \setlength{\tabcolsep}{5pt}
    \caption{Common hyperparameters}
    \label{tab:param-common}
    \centering
    \begin{tabular}{cc}\toprule
        \multicolumn{2}{c}{Common Hyperparameters}  \\\midrule
        Dropoout & 0.5  \\
        Number of layers & 4 \\
        Hidden dimension & 64 \\
        Learning rate & 0.001 \\
        Stopping patience & 100 epochs \\
        \bottomrule
    \end{tabular}
\end{table}

\begin{table}[h]
    \small
    \centering
    \setlength{\tabcolsep}{5pt}
    \caption{Hyperparameter search space of PANDA for benchmark datasets}
    \label{tab:param-panda}
    \begin{tabular}{cc}\toprule
        Hyperparameters & Search Space  \\\midrule
        $p_{\mathrm{high}}$ & \{80, 96, 112, 128\}  \\
        top $k$ & \{1, 3 ,5, 7, 10, 15, 20\} \\
        Centrality & \{Degree, Betweenness, Closeness, PageRank, Load \} \\
        \bottomrule
    \end{tabular}
\end{table}

\begin{table}[h!]
    \small
    \centering
    \caption{Best hyperparameter}
    \label{tab:best-param}
    \begin{tabular}{cc cccccc}\toprule
        Hyperparameter & Method & \textsc{Reddit-Binary} & \textsc{IMDB-Binary} & \textsc{Mutag} & \textsc{Enzymes} & \textsc{Proteins} & \textsc{Collab} \\ \midrule
        \multirow{4}{*}{Top-$k$} 
         & GCN   & 7  & 3 & 10 & 7 & 7 & 3 \\
         & GIN   & 10 & 3 & 10 & 10 & 5 & 10 \\
         & R-GCN & 5 & 5 & 5 & 10 & 10 & 3 \\
         & R-GIN & 15 & 5 & 5 & 15 & 5  & 10 \\
        \midrule
        \multirow{4}{*}{$p_{\mathrm{high}}$} 
         & GCN   & 80 & 80  & 112 & 128 & 128 & 128 \\
         & GIN   & 96 & 128 &  96 & 128 & 112 & 96 \\
         & R-GCN & 128 & 128 & 128 & 96  & 112 & 128 \\
         & R-GIN & 96 & 128 & 128 & 112 & 128 & 128 \\
        \midrule
        \multirow{4}{*}{$C(\mathcal{G})$} 
         & GCN & Degree & PageRank & Betweenness & PageRank & Closeness & Degree \\
         & GIN & PageRank & PageRank & PageRank & Closeness & Degree & Load \\
         & R-GCN & Degree & PageRank & PageRank & Betweenness & Betweenness & Degree \\
         & R-GIN & Degree & Degree & PageRank & Betweenness & Betweenness & Betweenness \\
        \bottomrule
    \end{tabular}
\end{table}

\subsection{Hardware specifications and libraries}
The following software and hardware environments were used for all experiments: \textsc{Ubuntu} 18.04 LTS, \textsc{Python} 3.7.13,  \textsc{PyTorch} 1.11.0, \textsc{PyTorch Geometric} 2.0.4, \textsc{Numpy} 1.21.6, \textsc{NetworkX} 2.6.3, \textsc{CUDA} 11.3, and \textsc{NVIDIA} Driver 465.19, and i9 CPU, and \textsc{NVIDIA RTX 3090}. We implemented our \textbf{PANDA} message passing framework in \textsc{PyTorch Geometric}.

\clearpage

\section{Additional Visualizations in \Cref{sec:why}}\label{app:vis-why}
\begin{figure}[h!]
    \centering
    \subfigure[GCN on \textsc{Enzymes}]{\includegraphics[width=0.24\textwidth]{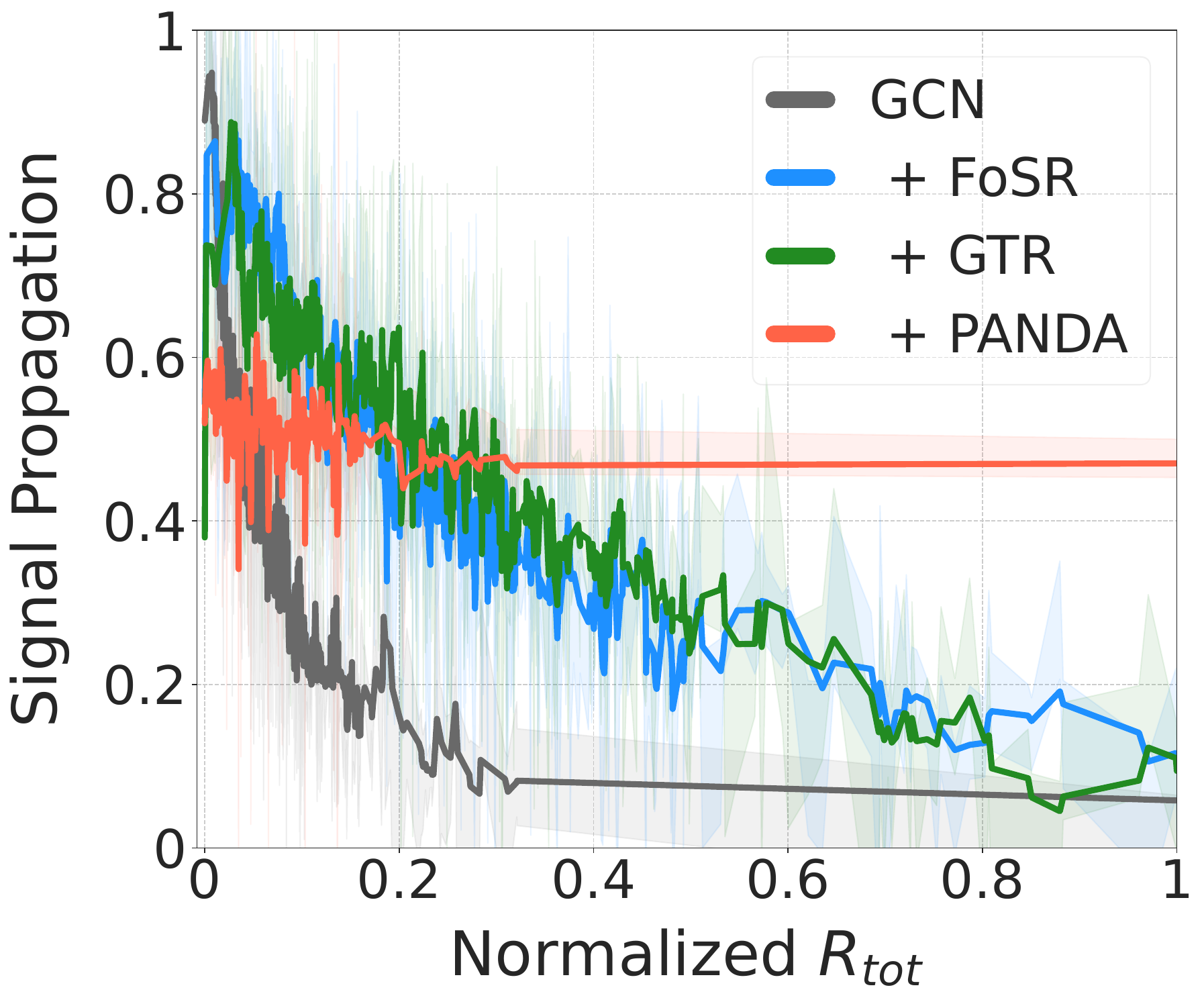}}
    \subfigure[GIN on \textsc{Enzymes}]{\includegraphics[width=0.24\textwidth]{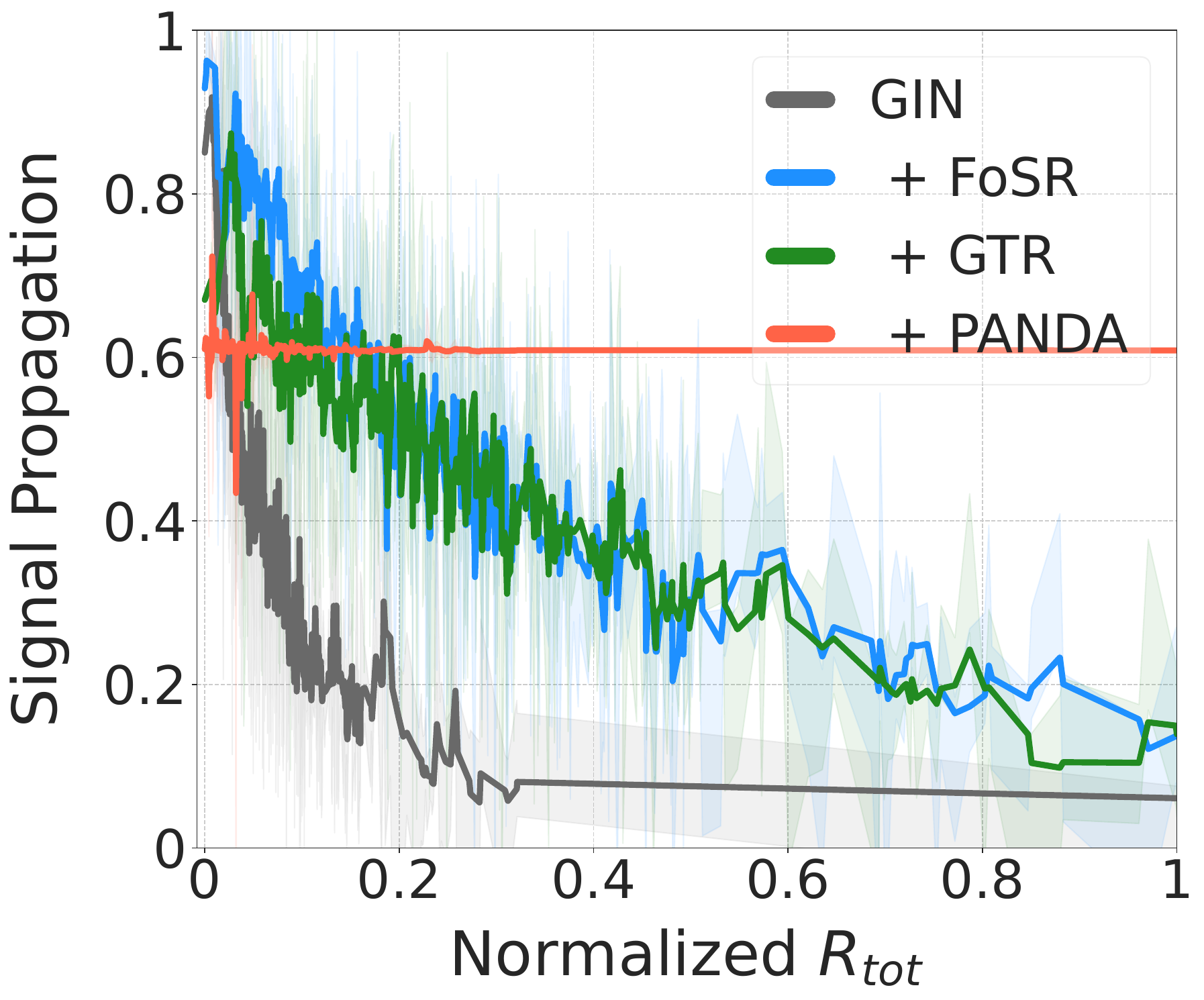}}
    \subfigure[GCN on \textsc{Mutag}]{\includegraphics[width=0.24\textwidth]{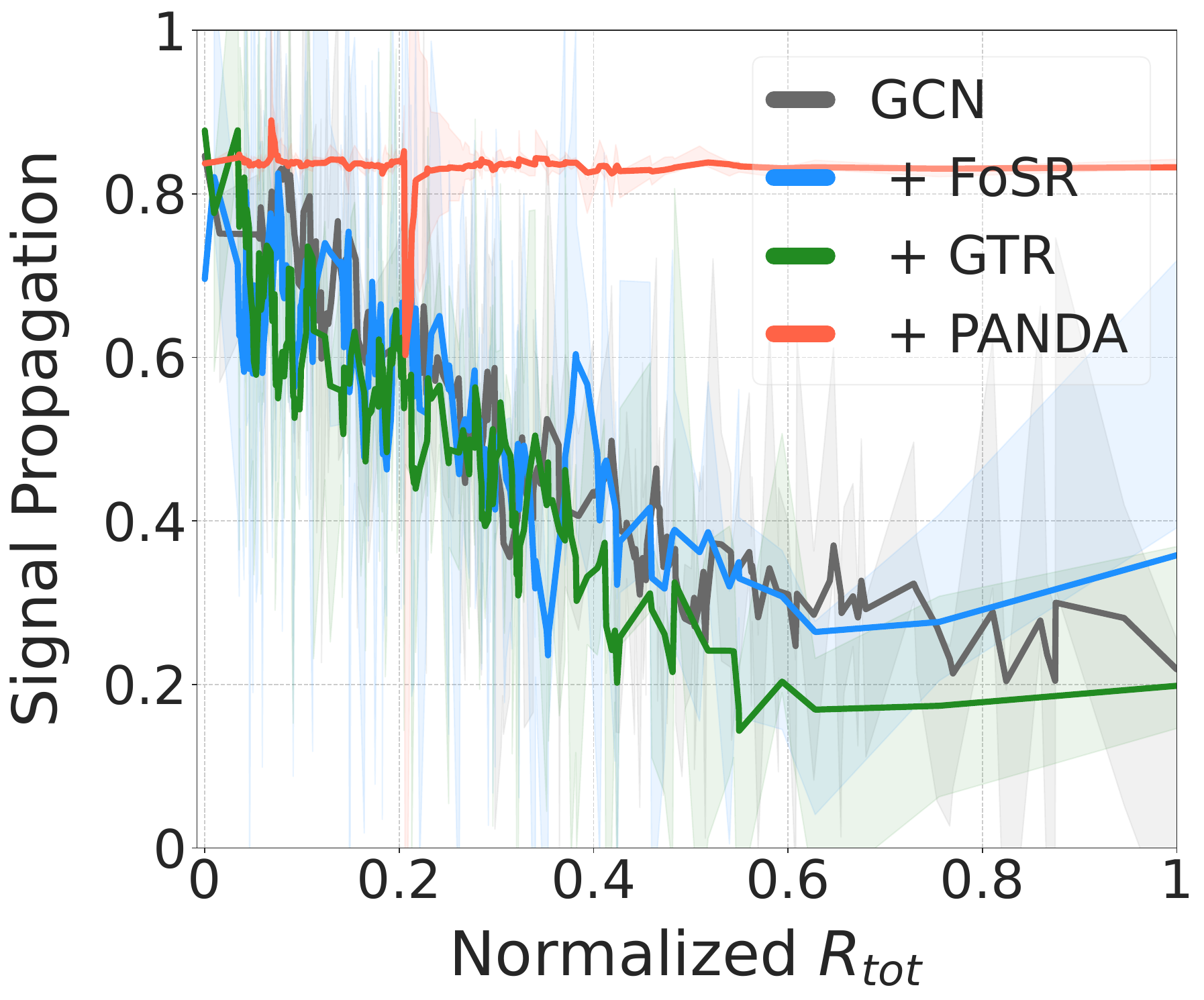}}
    \subfigure[GIN on \textsc{Mutag}]{\includegraphics[width=0.24\textwidth]{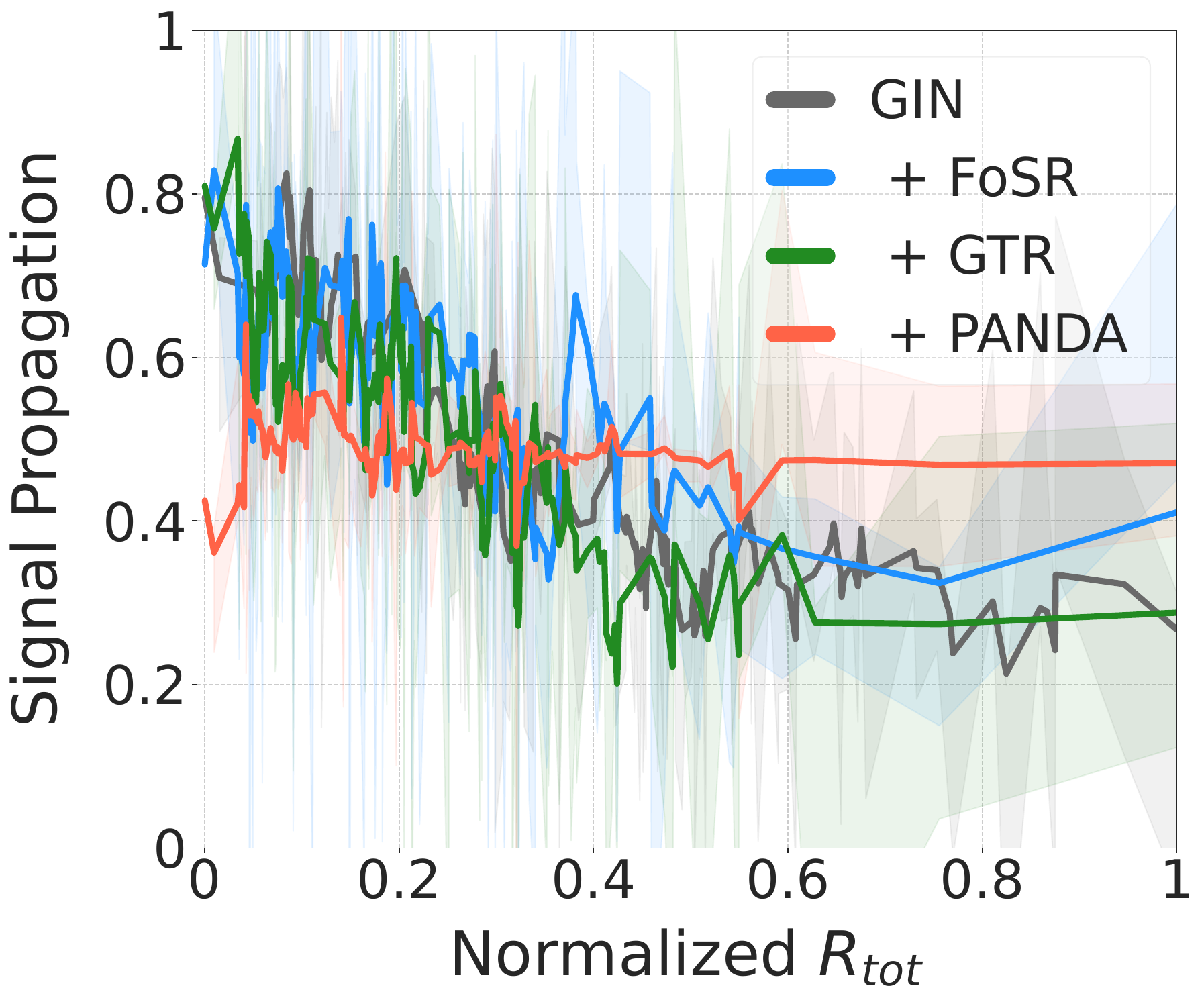}}
    \subfigure[GCN on \textsc{Proteins}]{\includegraphics[width=0.24\textwidth]{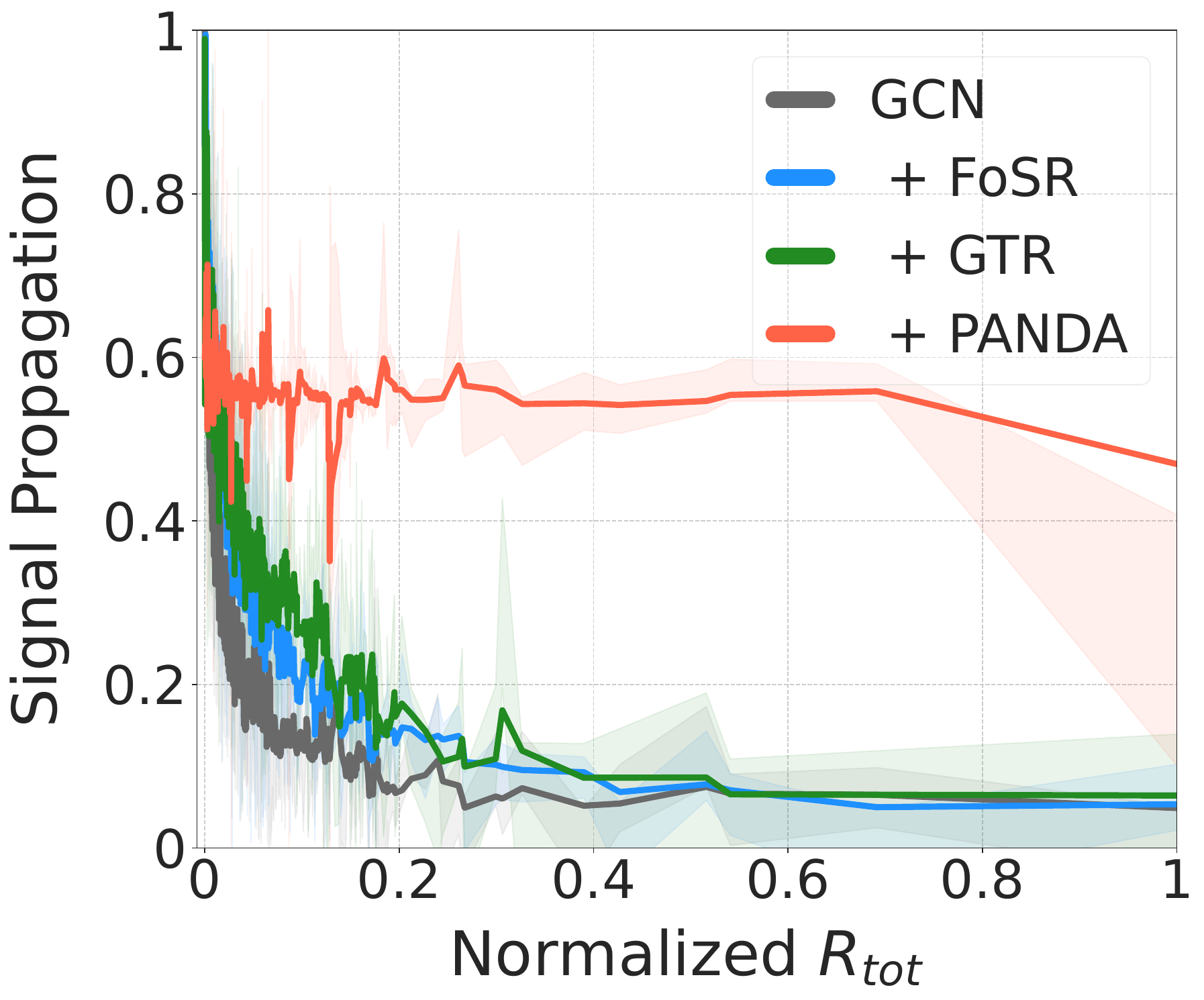}}
    \subfigure[GIN on \textsc{Proteins}]{\includegraphics[width=0.24\textwidth]{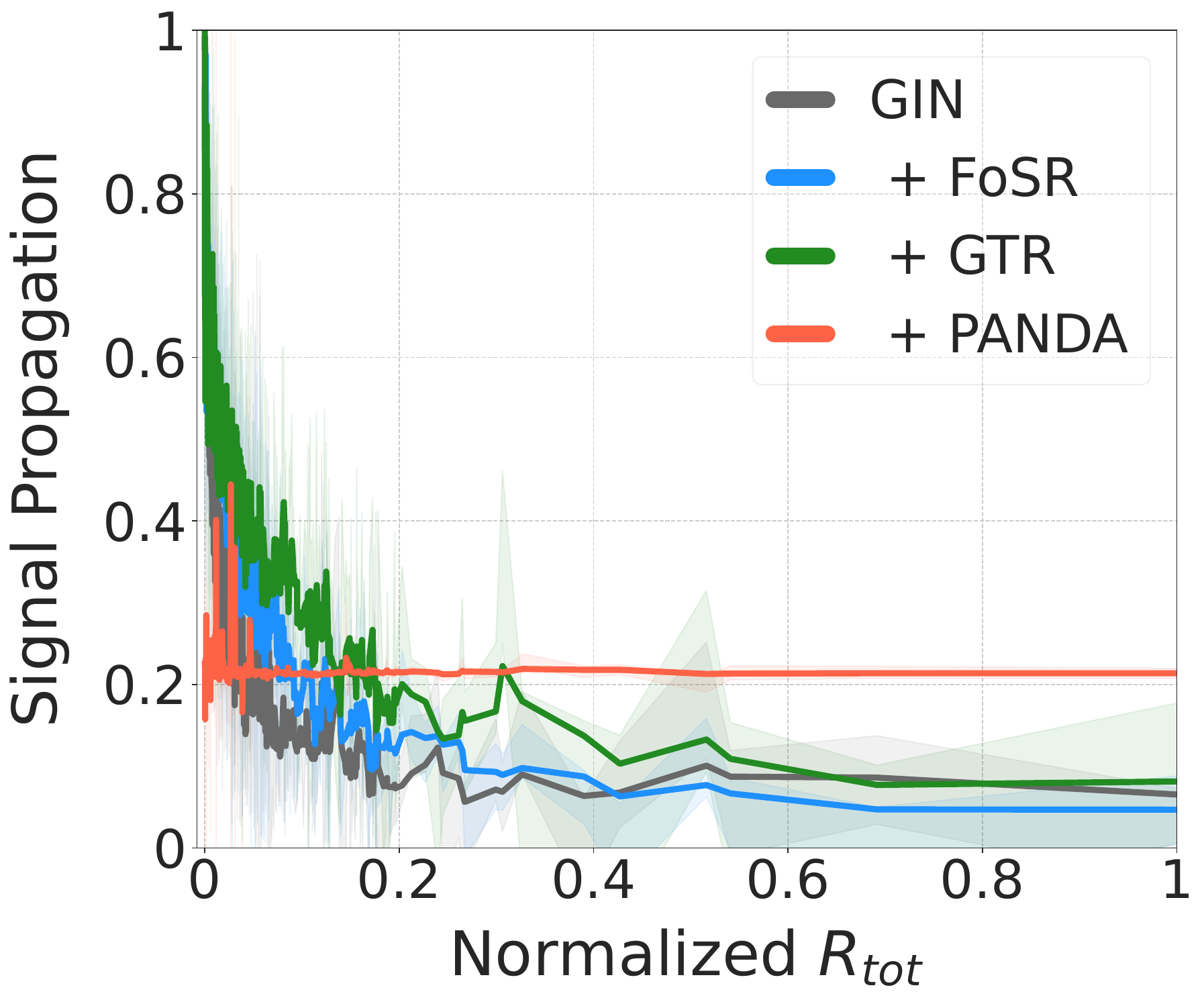}}
    \subfigure[GCN on \textsc{Reddit-Binary}]{\includegraphics[width=0.24\textwidth]{img/resistance/reddit_GCN_32.pdf}}
    \subfigure[GIN on \textsc{Reddit-Binary}]{\includegraphics[width=0.24\textwidth]{img/resistance/reddit_GIN_32.pdf}}
    \caption{The amount of signal propagated across the graph w.r.t. the normalized total effective resistance ($R_{tot}$) for all datasets.}
    \label{fig:signalweffres-app}
\end{figure}

\begin{figure}[h!]
    \centering
    \subfigure[\textsc{Reddit-Binary}]{\includegraphics[width=0.24\textwidth]{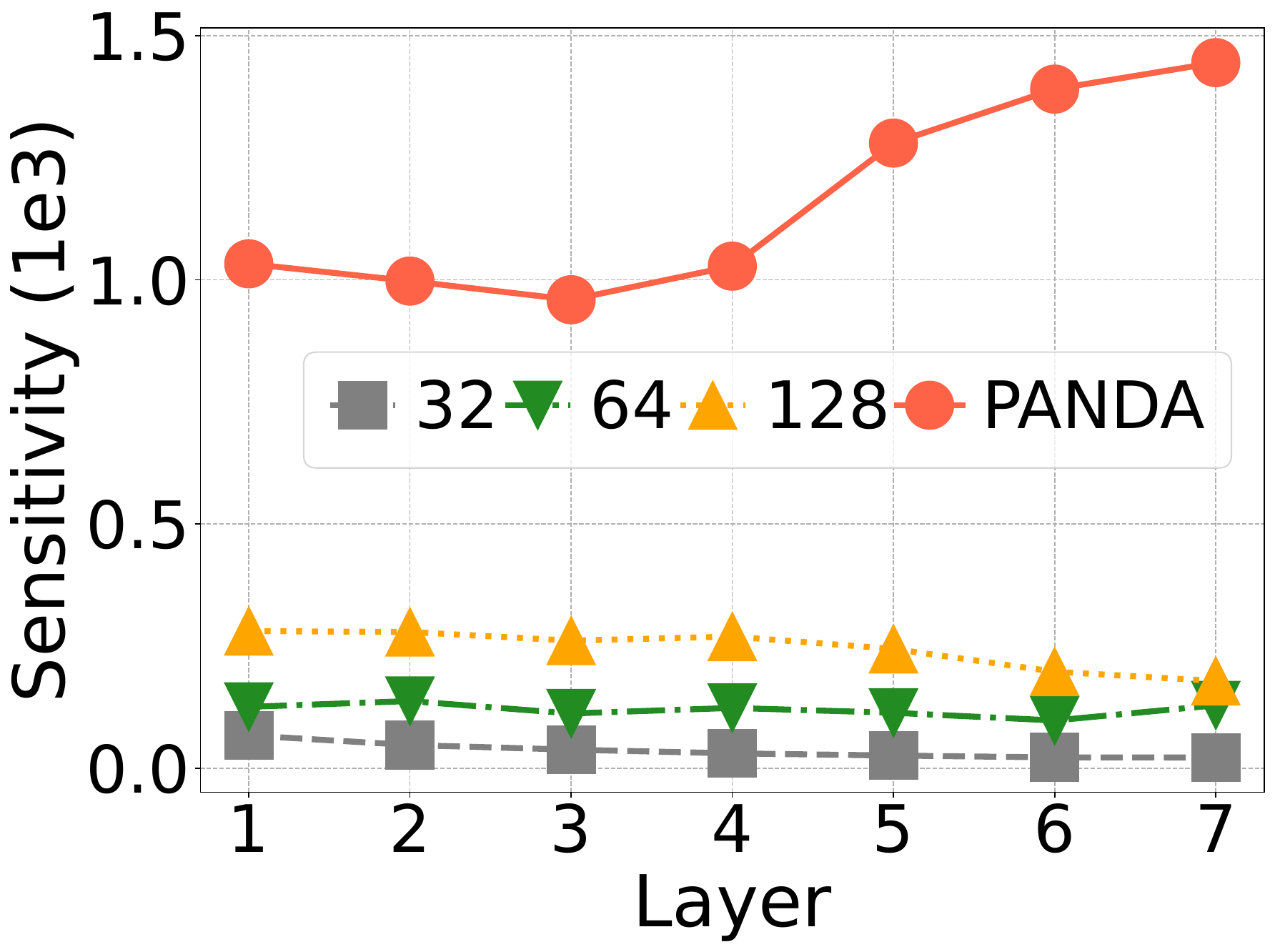}}
    \subfigure[\textsc{Reddit-Binary}]{\includegraphics[width=0.24\textwidth]{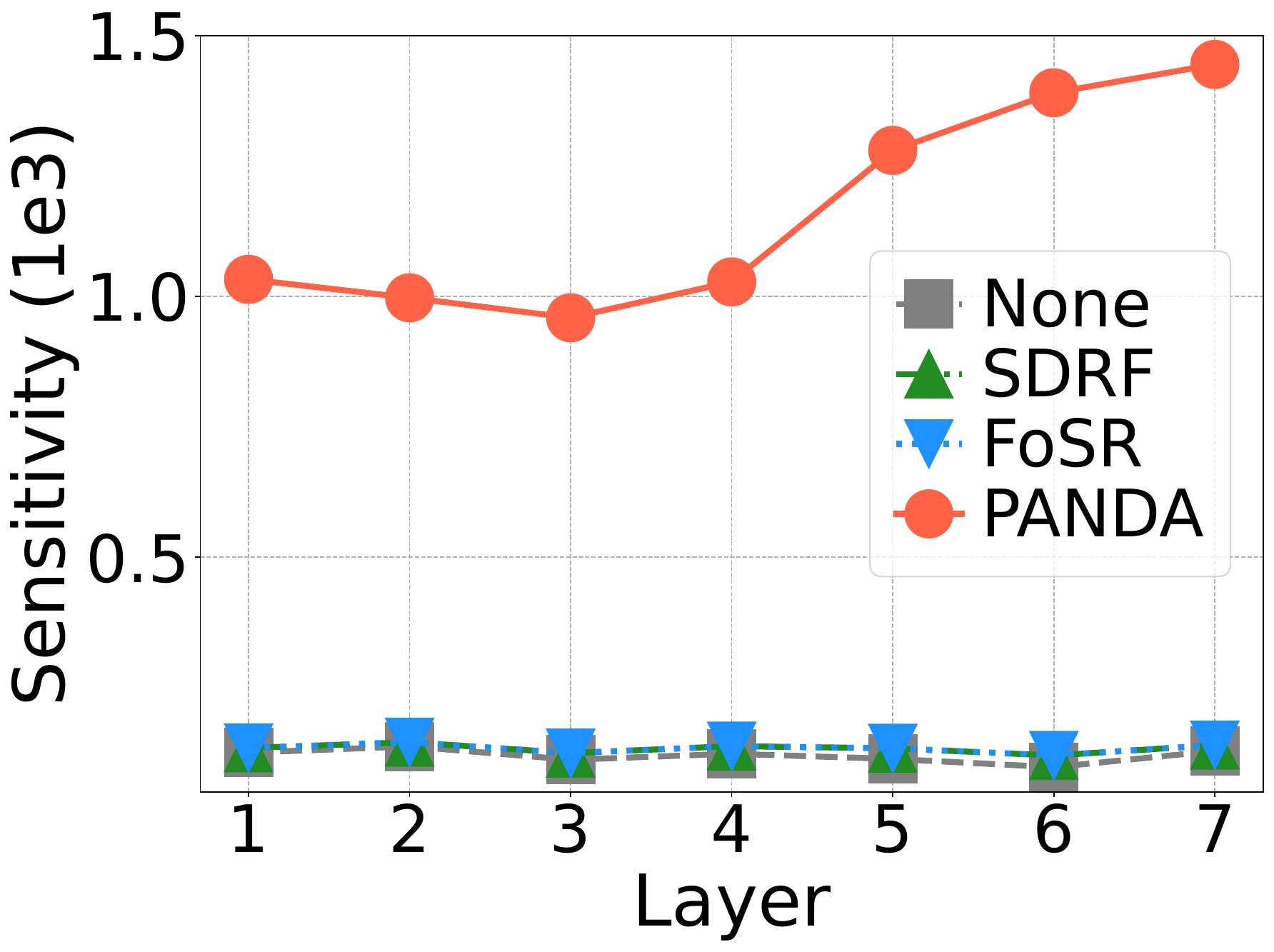}}
    \subfigure[\textsc{IMDB-Binary}]{\includegraphics[width=0.24\textwidth]{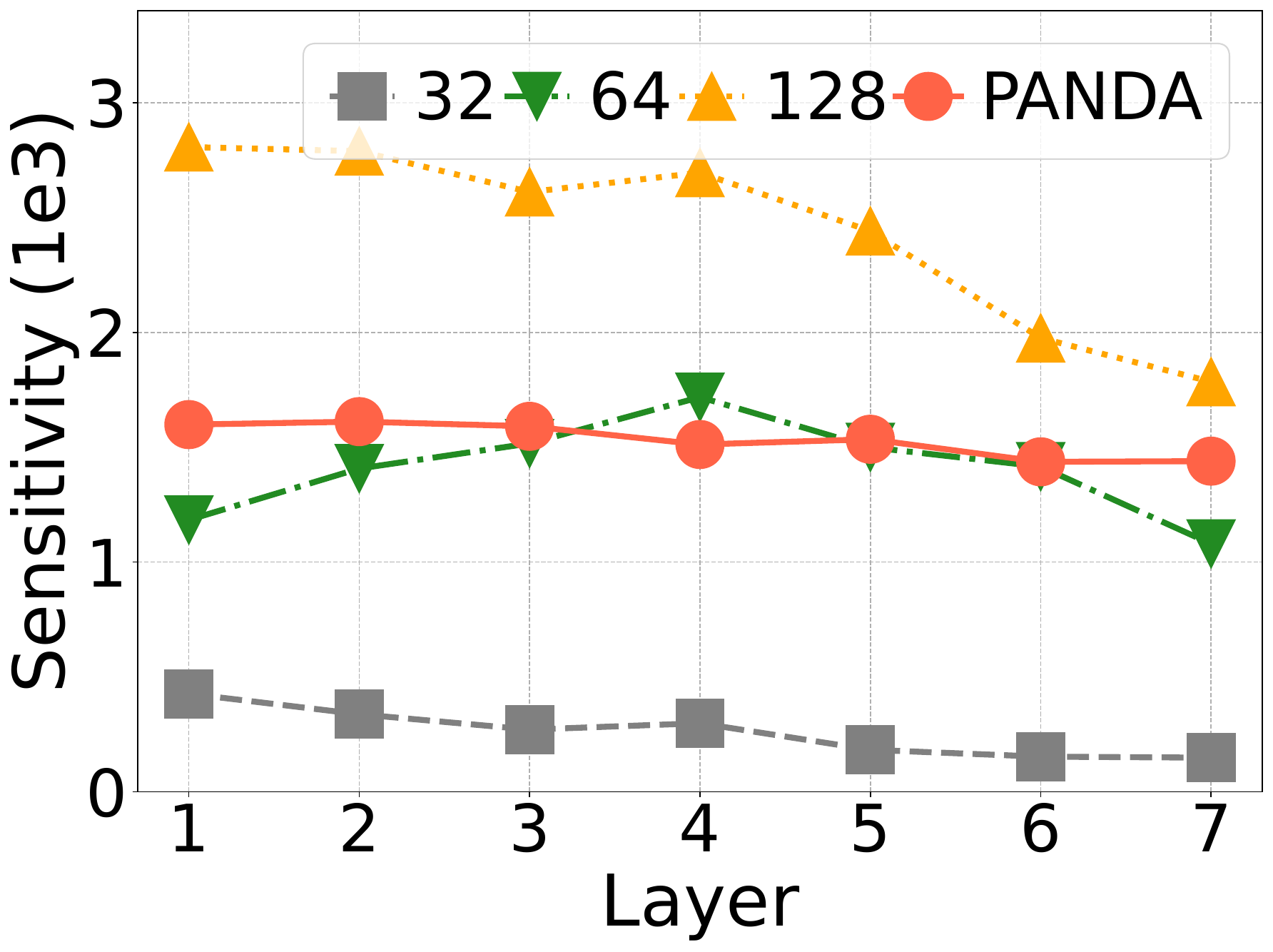}}
    \subfigure[\textsc{IMDB-Binary}]{\includegraphics[width=0.24\textwidth]{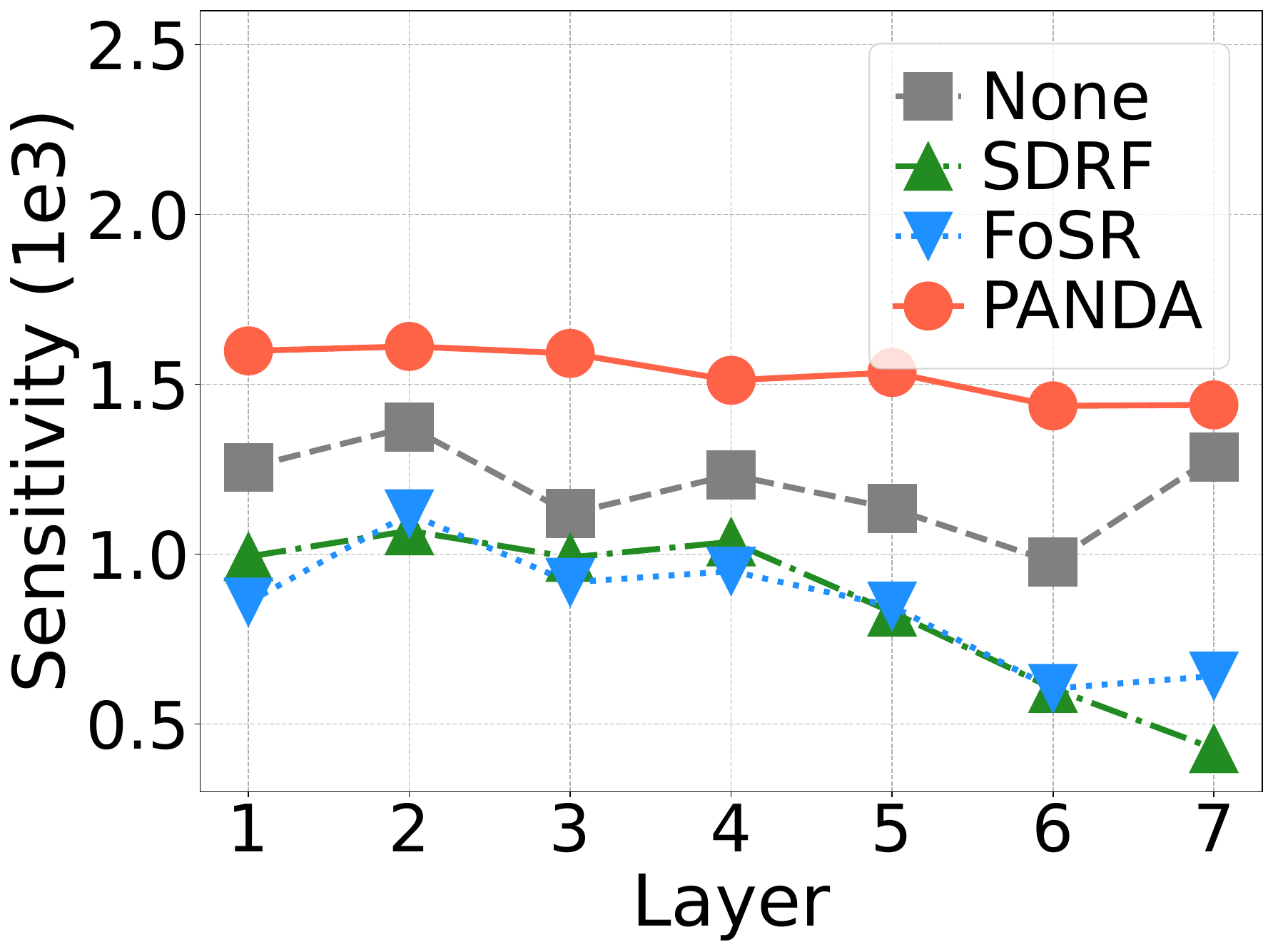}}
    \subfigure[\textsc{Mutag}]{\includegraphics[width=0.24\textwidth]{img/bound/sens_dim-mutag.pdf}}
    \subfigure[\textsc{Mutag}]{\includegraphics[width=0.24\textwidth]{img/bound/sens_method-mutag.pdf}}
    \subfigure[\textsc{Enzymes}]{\includegraphics[width=0.24\textwidth]{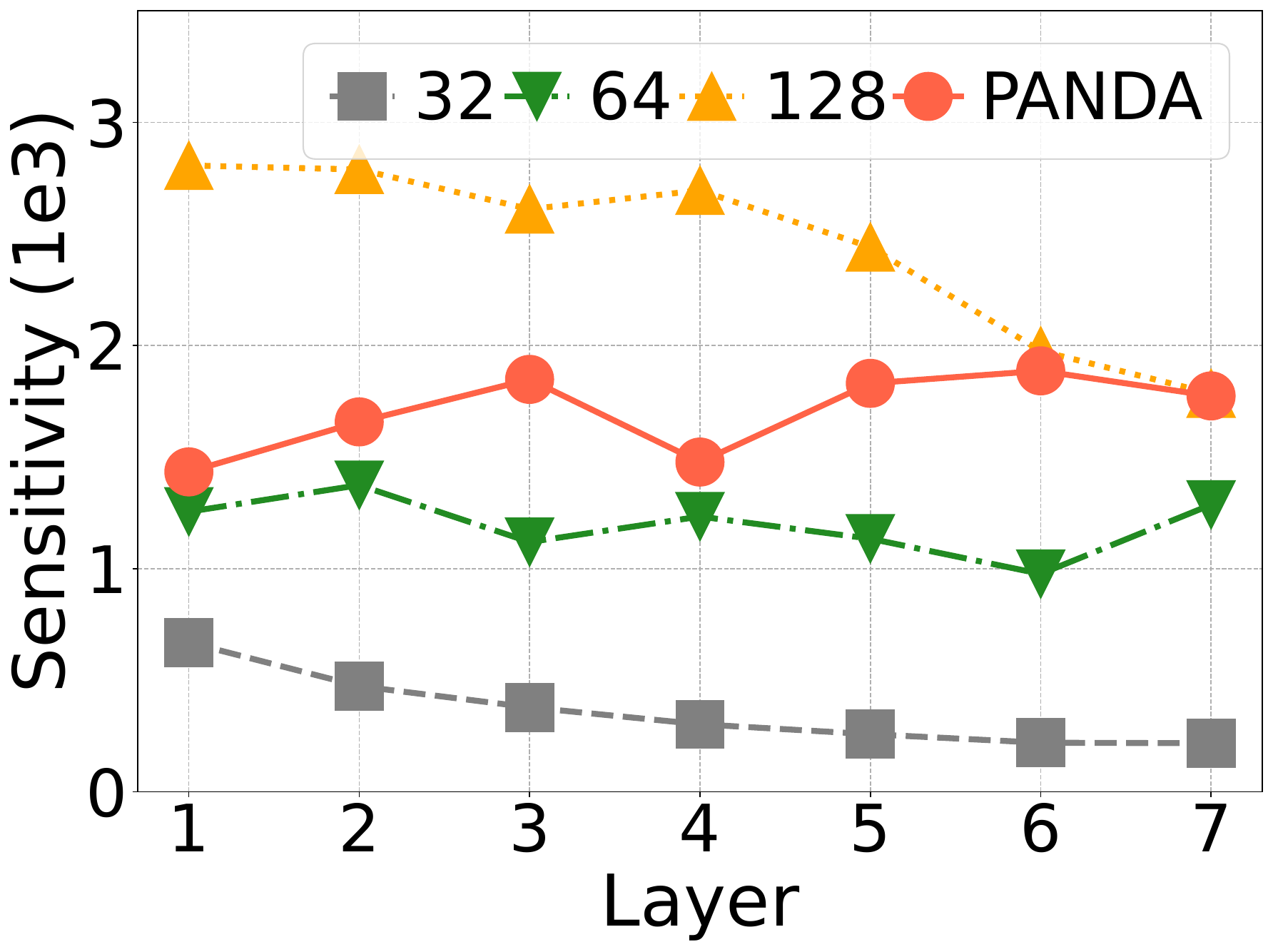}}
    \subfigure[\textsc{Enzymes}]{\includegraphics[width=0.24\textwidth]{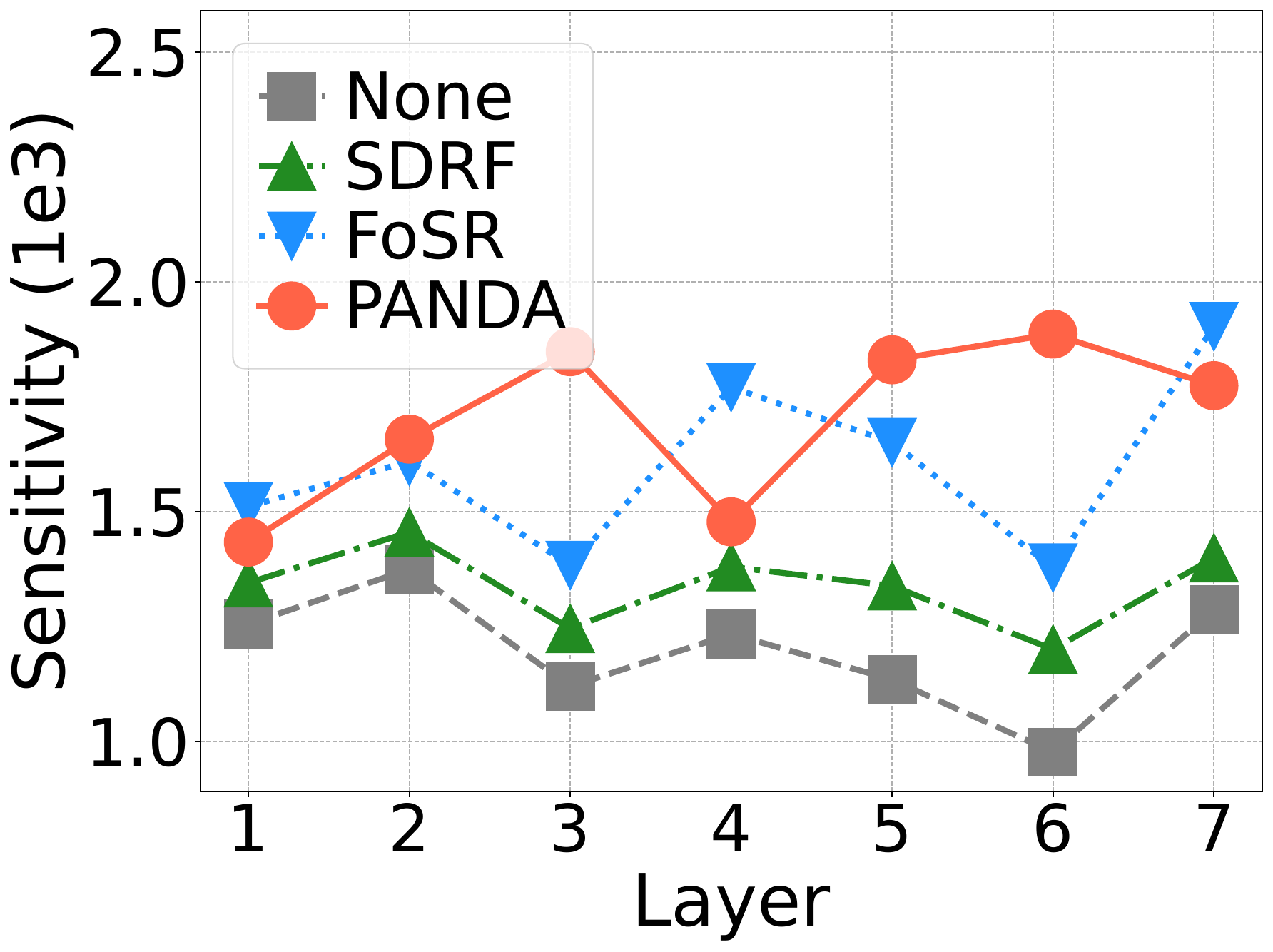}}
    \subfigure[\textsc{Proteins}]{\includegraphics[width=0.24\textwidth]{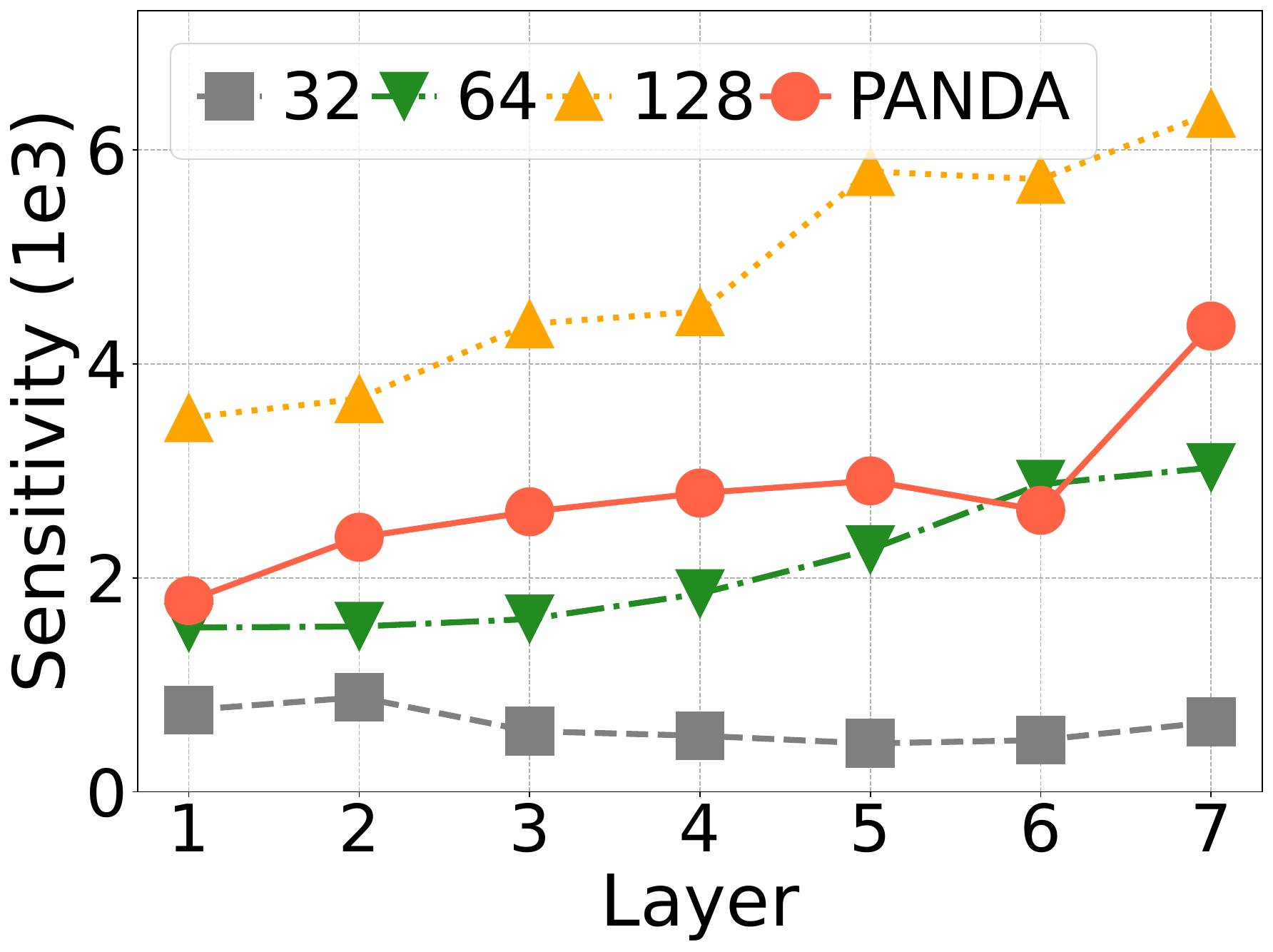}}
    \subfigure[\textsc{Proteins}]{\includegraphics[width=0.24\textwidth]{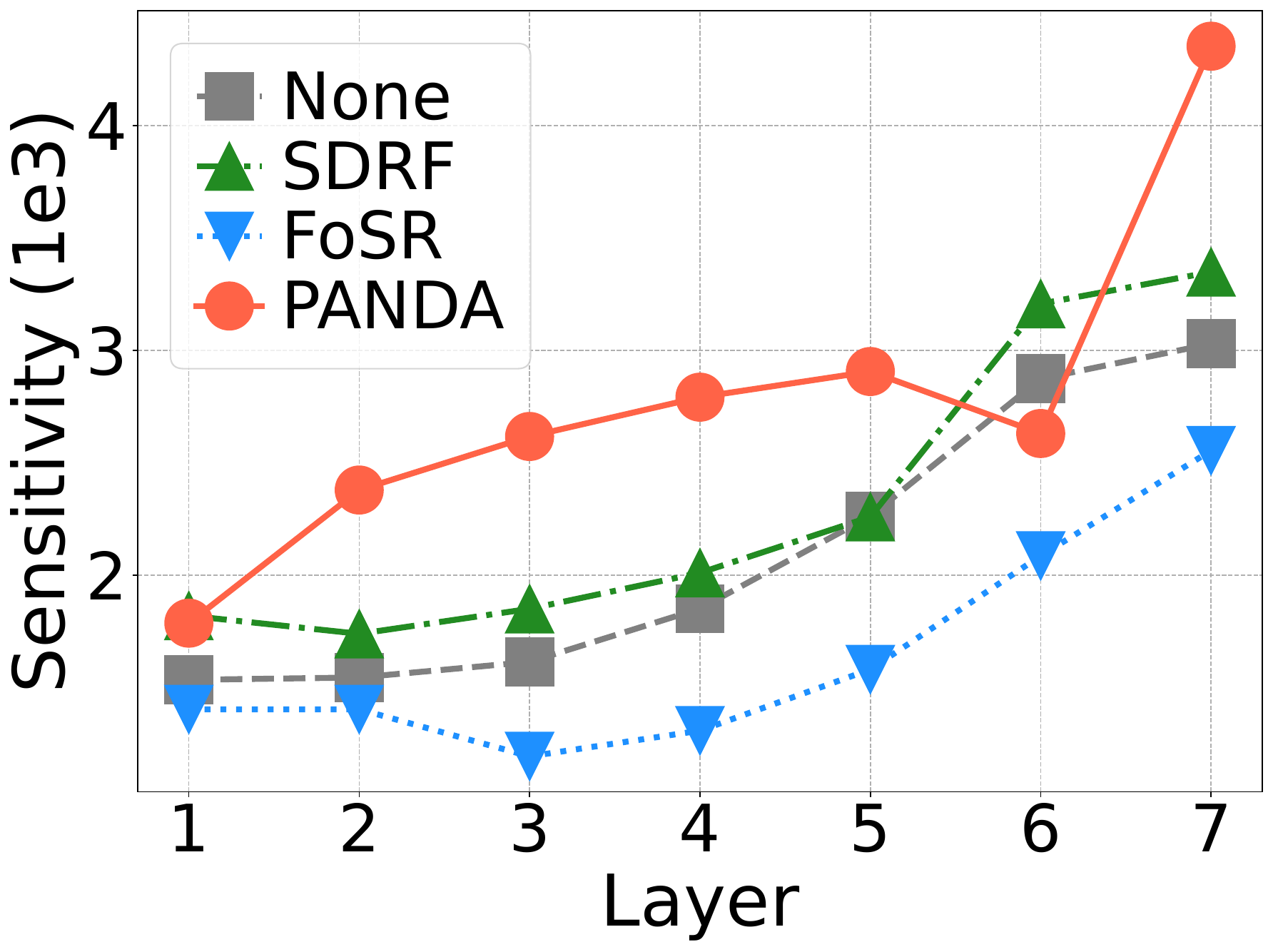}}
    \subfigure[\textsc{Collab}]{\includegraphics[width=0.24\textwidth]{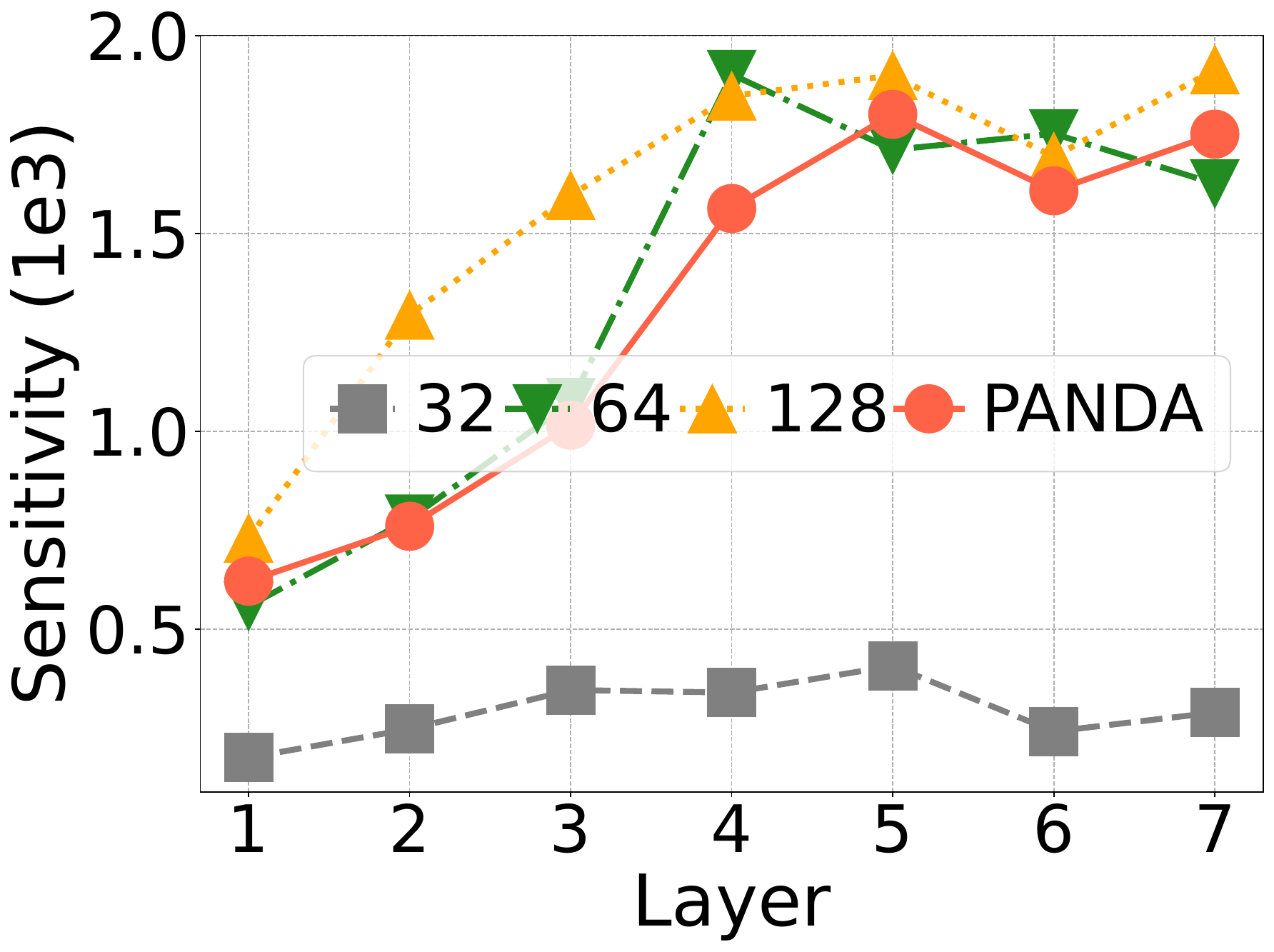}}
    \subfigure[\textsc{Collab}]{\includegraphics[width=0.24\textwidth]{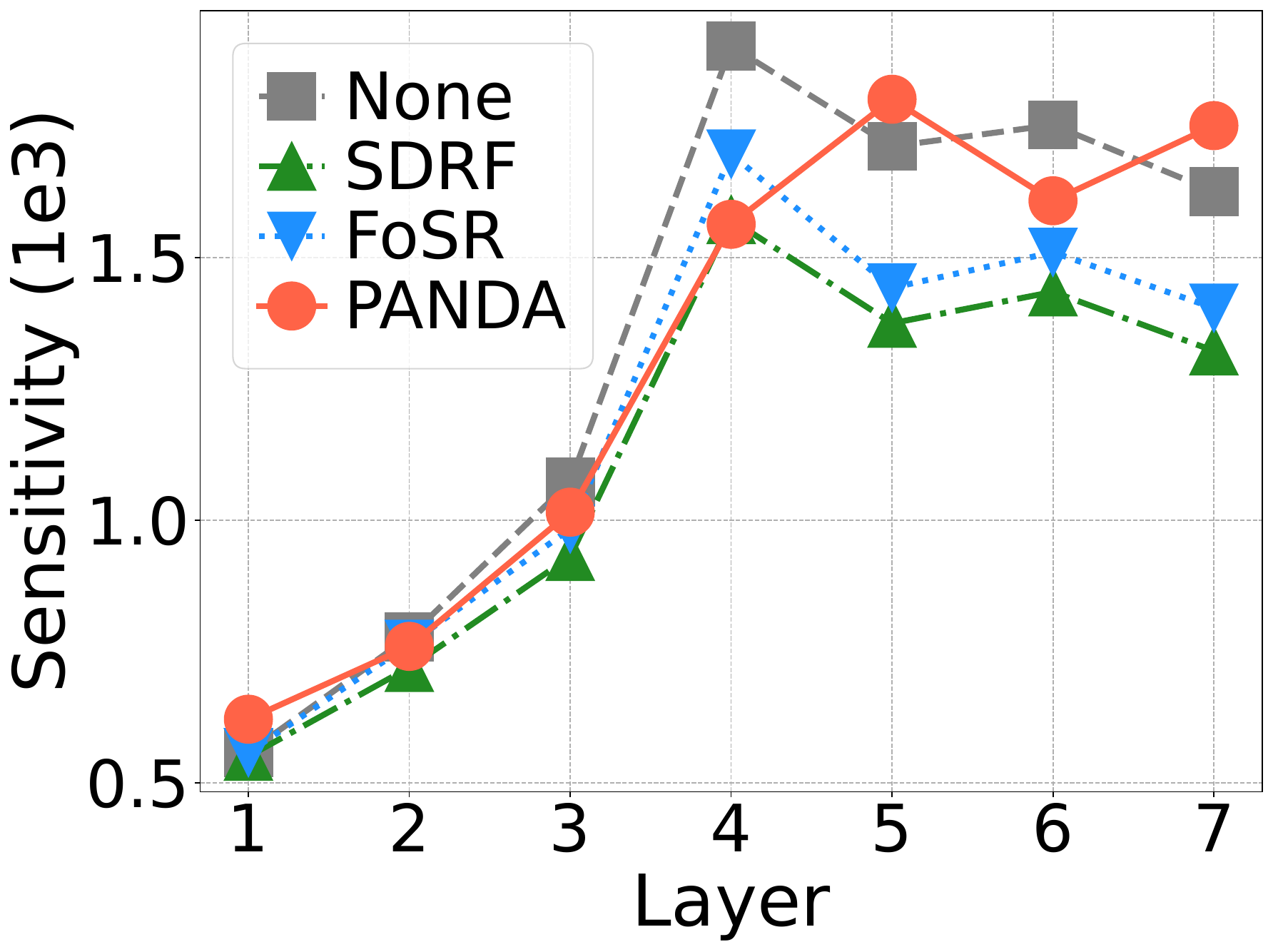}}
    \caption{Empirical sensitivity across layers for GCN on all datasets.}
    \label{fig:sens-app}
\end{figure}

\begin{figure}[h!]
    \centering
    \subfigure[\textsc{Reddit-Binary}]{\includegraphics[width=0.24\textwidth]{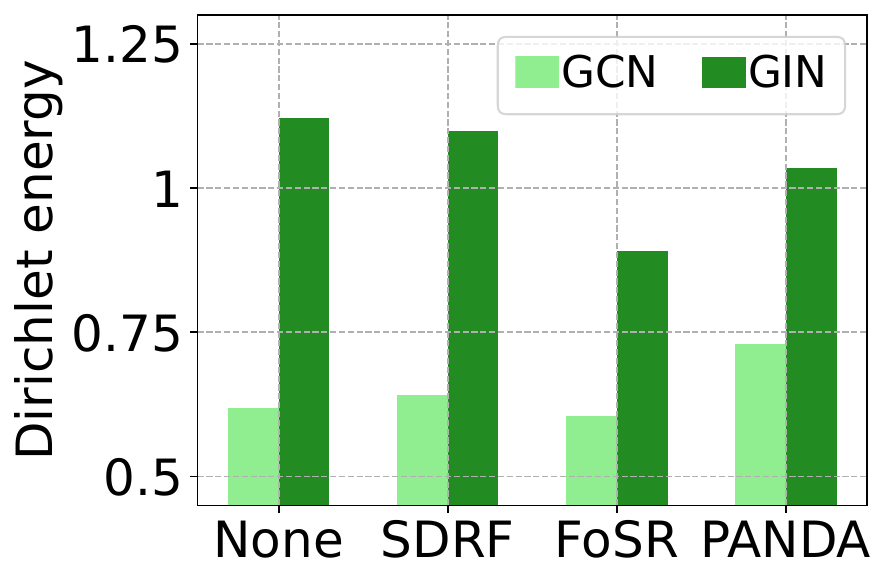}}
    \subfigure[\textsc{IMDB-Binary}]{\includegraphics[width=0.24\textwidth]{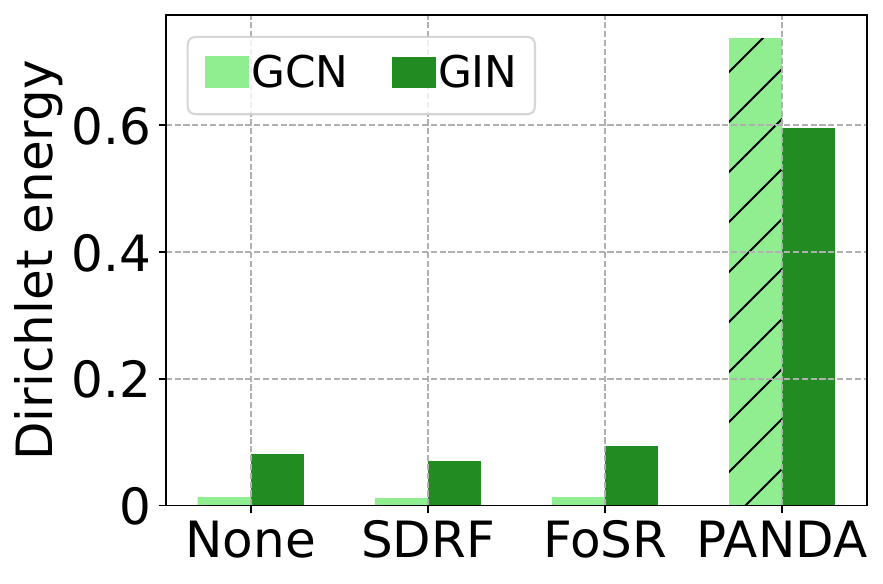}}
    \subfigure[\textsc{Mutag}]{\includegraphics[width=0.24\textwidth]{img/energy/mutag_energy.pdf}}
    \subfigure[\textsc{Proteins}]{\includegraphics[width=0.24\textwidth]{img/energy/proteins_energy.pdf}}
    \subfigure[\textsc{Enzymes}]{\includegraphics[width=0.24\textwidth]{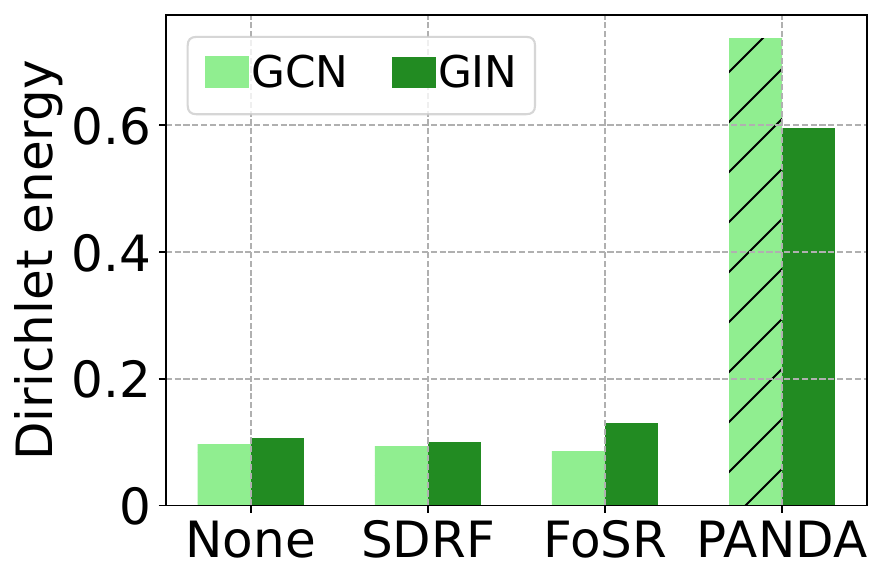}}
    \subfigure[\textsc{Collab}]{\includegraphics[width=0.24\textwidth]{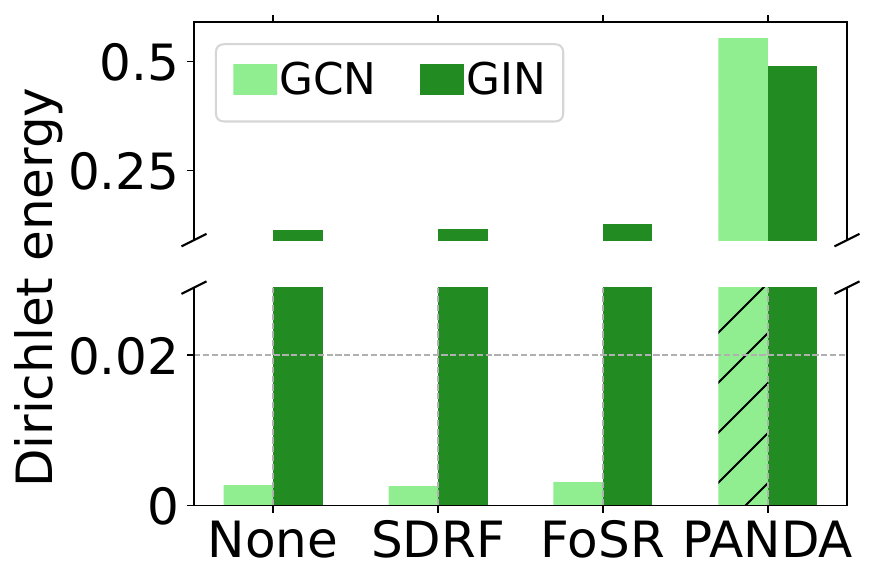}}
    \caption{Dirichlet energy of baselines and PANDA.}
    \label{fig:energy-app}
\end{figure}

\clearpage

\section{Empirical Computational Complexity}\label{app:runtime}
\subsection{Empirical Runtime Analysis}
We compare the runtime against the centrality metric used by PANDA and the rewiring method FoSR. As can be seen in \Cref{tab:runtime-centrality}, using centrality metrics for \textsc{IMDB-Binary}, \textsc{Mutag}, and \textsc{Proteins} is faster than the rewiring method FoSR. However, when the size of the graph such as \textsc{Reddit-Binary} or \textsc{Collab} is large, metrics such as betweenness, closeness, and load centrality that calculate the shortest path take a long time.

In \Cref{tab:runtime}, we also measure and compare the runtime of the forward pass of the original GCN and PANDA-GCN.
\textbf{PANDA} takes more time than the original GCN due to the amount of computation used in Eqs.~\eqref{eq:f} and ~\eqref{eq:g}.
However, in our design, it is meaningful to select the different dimensions of high-dimensional nodes that are learned to be more significant for each message (edge) in $g(\cdot)$ of Eq.~\eqref{eq:g}. 
In future work, we aim to design an efficient method for selecting the dimensions.

\begin{table*}[h!]
    \small
    \setlength{\tabcolsep}{3pt}
    \centering
    \caption{Empirical runtime comparison of rewiring methods and centrality metrics (in seconds)}
    \label{tab:runtime-centrality}
    \begin{tabular}{cr rrrrrr}\toprule
        \multicolumn{2}{c}{Method} & \textsc{Reddit-Binary} & \textsc{IMDB-Binary} & \textsc{Mutag} & \textsc{Enzymes} & \textsc{Proteins} & \textsc{Collab} \\ \midrule
        \multicolumn{2}{c}{FoSR}  & 11.97 & 3.36 & 3.43 & 3.28 & 3.47 & 6.55 \\ 
        \cmidrule(lr){1-8}
        \multirow{5}{*}{\textbf{PANDA}} 
        & Degree      & 3.33   & 0.24 & 0.03 & 0.14 & 0.27 & 21.05\\ 
        & Betweenness & 965.31 & 1.16 & 0.06 & 1.20 & 5.58 & 349.68\\ 
        & Closeness   & 238.32 & 0.38 & 0.43 & 0.40 & 1.53 & 55.60\\ 
        & PageRank    & 9.65   & 1.50 & 0.58 & 1.09 & 1.93 & 43.56\\ 
        & Load        & 953.12 & 1.03 & 0.11 & 1.21 & 5.17 & 331.34\\ 
        \bottomrule
    \end{tabular}
\end{table*}

\begin{table*}[h!]
    \small
    \setlength{\tabcolsep}{4pt}
    \centering
    \caption{Empirical runtime: PANDA-GCN forward pass duration (in milliseconds).}
    \label{tab:runtime}
    \begin{tabular}{c rrrrrr}\toprule
        Method & \textsc{Reddit-Binary} & \textsc{IMDB-Binary} & \textsc{Mutag} & \textsc{Enzymes} & \textsc{Proteins} & \textsc{Collab} \\ \midrule
        GCN & 11.5 & 10.6 & 7.32 & 19.2 & 12.1 & 38.5 \\
        \textbf{PANDA-GCN} & 41.3 & 33.9 & 27.3 & 68.4 & 45.5 & 98.7\\ \midrule
    \end{tabular}
\end{table*}
\section{Additional Experiments on Graph Classification}
\subsection{Relational GNNs with PANDA}\label{app:relational}
\Cref{tab:main-rgnn} shows the experimental results for R-GCN and R-GIN. Our \textbf{PANDA} method shows the highest accuracy on most datasets for both R-GCN and R-GIN, significantly outperforming the baseline (``none'') and other rewiring methods.
For example, in the RGIN model, \textbf{PANDA} is shown to be the most effective method, achieving an accuracy of 91.36 on \textsc{Reddit-Binary} and 88.2 on \textsc{Mutag}.
Surprisingly, for \textsc{Enzymes}, \textbf{PANDA} shows a significant 36.09\% improvement over the baseline (None) method. 
In R-GCN, GTR shows higher performance than \textbf{PANDA}.
However, overall, \textbf{PANDA} shows high effectiveness in R-GCN and R-GIN.
\begin{table}[h!]
    \small
    \setlength{\tabcolsep}{3pt}
    \centering
    \caption{Results of our method and rewiring methods for R-GCN and R-GIN. We show the best three in \BEST{red} (first), \SECOND{blue} (second), and \THIRD{purple} (third).}
    \label{tab:main-rgnn}
    \begin{tabular}{l cccccc}\toprule
        Method & \textsc{Reddit-Binary} & \textsc{IMDB-Binary} & \textsc{Mutag} & \textsc{Enzymes} & \textsc{Proteins} & \textsc{Collab} \\ \midrule
        R-GCN (None) & 49.850\std{0.653} & 50.012\std{0.917} & 69.250\std{2.085} & 28.600\std{1.186} & 69.518\std{0.725} & 33.602\std{1.047} \\
        \;+ Last Layer FA & 49.800\std{0.626} &50.650\std{0.964} & 70.550\std{1.810} & 28.233\std{1.138} & 69.527\std{0.815} & 34.732\std{1.194} \\
        \;+ Every Layer FA & 49.950\std{0.593} & 50.500\std{0.891} & 70.500\std{1.836} & 33.400\std{1.142} & 71.670\std{0.882} & 33.616\std{0.978} \\
        \;+ DIGL & 49.995\std{0.619} & 49.670\std{0.843} & 73.400\std{2.007} & 28.283\std{1.213} & 68.232\std{0.851} & 19.926\std{1.441} \\
        \;+ SDRF & 58.620\std{0.647} & 53.640\std{1.043} & 72.300\std{2.215} & 33.483\std{1.245} & 69.107\std{0.759} & 67.990\std{0.386} \\
        \;+ FoSR & \THIRD{76.590\std{0.531}} & \THIRD{64.050\std{1.123}} & \THIRD{84.450\std{1.517}} & \THIRD{35.633\std{1.151}} & \THIRD{73.795\std{0.692}} & \THIRD{70.650\std{0.482}} \\
        \;+ GTR & \SECOND{80.180\std{0.600}} & \SECOND{65.090\std{0.930}} & \SECOND{85.500\std{1.470}} & \SECOND{41.330\std{1.280}} & \SECOND{75.780\std{0.760}} & \BEST{74.340\std{0.410}} \\
        \cmidrule(lr){1-7}
        \;+ \textbf{PANDA} & \BEST{80.200\std{0.913}} & \BEST{66.790\std{1.088}} & \BEST{90.050\std{1.466}} & \BEST{43.900\std{1.176}} & \BEST{76.000\std{0.802}} & \SECOND{71.400\std{0.376}} \\ \midrule
        \midrule
        R-GIN (None) & 87.965\std{0.564} & 68.889\std{0.872} & 83.050\std{1.439} & 39.017\std{1.166} & 70.500\std{0.809} & 75.544\std{0.323} \\
        \;+ Last Layer FA & \THIRD{89.995\std{0.647}} & 69.710\std{1.025} & 80.600\std{1.639} & 48.183\std{1.401} & 70.304\std{0.844} & 75.434\std{0.491} \\
        \;+ Every Layer FA & 56.855\std{0.943} & 71.480\std{0.876} & 83.050\std{1.518} & \BEST{54.950\std{1.331}} & 71.045\std{0.909} & 75.432\std{0.475} \\
        \;+ DIGL & 74.425\std{0.723} & 63.930\std{0.947} & 81.450\std{1.488} & 37.600\std{1.198} & 71.312\std{0.757} & 54.714\std{0.416} \\
        \;+ SDRF & 86.825\std{0.523} & 70.210\std{0.806} & 82.700\std{1.782} & 39.583\std{1.333} & 70.696\std{0.815} & 76.480\std{0.388} \\
        \;+ FoSR & 89.665\std{0.416} & \SECOND{71.810\std{0.880}} & \SECOND{86.150\std{1.492}} & 45.550\std{1.258} & \THIRD{74.670\std{0.692}} & \THIRD{76.480\std{0.390}} \\
        \;+ GTR  & \SECOND{90.410\std{0.410}} & \THIRD{71.490\std{0.930}} & \THIRD{86.100\std{1.762}} & \THIRD{50.030\std{1.320}} & \SECOND{75.640\std{0.740}} & \SECOND{77.450\std{0.390}} \\
        \cmidrule(lr){1-7}
        \;+ \textbf{PANDA} & \BEST{91.360\std{0.372}} & \BEST{72.090\std{0.936}} & \BEST{88.200\std{1.513}} & \SECOND{53.100\std{1.344}} & \BEST{76.170\std{0.776}} & \BEST{77.800\std{0.355}} \\ \midrule
    \end{tabular}
\end{table}

\subsection{Impact of Centrality Metrics on PANDA-GIN}\label{app:centrality-result}
\Cref{tab:centrality-gin} compares results obtained using different kinds of centrality metrics in PANDA-GIN.
For PANDA-GIN, the PageRank centrality performs best on \textsc{Reddit-Binary}, \textsc{IMDB-Binary}, and \textsc{Mutag}. This shows that in social networks such as \textsc{Reddit-Binary} and \textsc{IMDB-Binary}, the dimension of the node that receives links from highly influential nodes must be large, as is the characteristic of PageRank.

\begin{table}[h!]
    \small
    \setlength{\tabcolsep}{3pt}
    \centering
    \caption{Results of our method and rewiring methods for PANDA-GIN. We show the best three in \BEST{red} (first), \SECOND{blue} (second), and \THIRD{purple} (third).}
    \label{tab:centrality-gin}
    \begin{tabular}{l cccccc}\toprule
        $C(\mathcal{G})$ & \textsc{Reddit-Binary} & \textsc{IMDB-Binary} & \textsc{Mutag} & \textsc{Enzymes} & \textsc{Proteins} & \textsc{Collab} \\ \cmidrule(lr){1-7}
        Degree & \THIRD{90.640\std{0.414}} & \SECOND{71.900\std{0.929}} & \THIRD{86.200\std{2.263}} & 37.450\std{1.376} & \BEST{75.759\std{0.856}} & \THIRD{74.950\std{0.402}} \\
        Betweenness & \SECOND{90.700\std{0.378}} & 71.390\std{0.857} & 86.150\std{1.999} & \THIRD{44.133\std{1.363}} & \SECOND{75.027\std{0.871}} & 74.644\std{0.491} \\  
        Closeness & 90.080\std{0.434} & 71.310\std{0.963} & \SECOND{88.400\std{1.612}} & \BEST{46.200\std{1.410}} & 74.741\std{0.845} & \SECOND{74.956\std{0.412}} \\  
        PageRank & \BEST{91.055\std{0.402}} & \BEST{72.560\std{0.917}} & \BEST{88.750\std{1.570}} & 41.483\std{1.237} & 74.223\std{0.702} & 74.500\std{0.464} \\  
        Load & 90.625\std{0.417} & \SECOND{71.900\std{0.967}} & 86.050\std{1.774} & \SECOND{44.150\std{1.455}} & \THIRD{74.812\std{0.878}} & \BEST{75.110\std{0.210}} \\ \bottomrule
    \end{tabular}
\end{table}

\subsection{Comparison of Centrality and Random Selection}
\Cref{tab:random} shows the results of an experiment by randomly expanding $k$ nodes according to the optimal $k$ number for each dataset.
As a result of the experiment, in the case of \textsc{Enzymes}, \textsc{Proteins}, and \textsc{Reddit}, there are clear differences. However, in \textsc{Mutag}, both results are comparable.
The graph statistics of \textsc{Mutag} are a minimum of 10 nodes, a maximum of 28 nodes, and an average of 17.93 nodes in a graph. Therefore, selecting some nodes randomly may not differ much from selecting nodes based on centrality. On the other hand, in the case of \textsc{Enzymes}, the average number of nodes is 32.36, and the maximum number of nodes is 126, so there is a non-trivial difference between them.

\begin{table}[h!]
    \small
    \setlength{\tabcolsep}{3pt}
    \centering
    \caption{Results of an experiment by randomly expanding nodes according to the optimal number for each dataset}
    \label{tab:random}
    \begin{tabular}{l cccccc}\toprule
        PANDA-GCN & \textsc{Reddit-Binary} & \textsc{IMDB-Binary} & \textsc{Mutag} & \textsc{Enzymes} & \textsc{Proteins} & \textsc{Collab} \\ \cmidrule(lr){1-7}
        Random & 72.000\std{0.850} & 54.000\std{1.200} & 85.350\std{1.525} & 26.300\std{1.280} & 73.520\std{0.795} & 54.435\std{0.548} \\
        Centrality & 80.690\std{0.721} & 63.760\std{1.012} & 85.750\std{1.396} & 31.550\std{1.230} & 76.000\std{0.774} & 68.400\std{0.452} \\ \bottomrule
    \end{tabular}
\end{table}

\clearpage
\section{Additional Visualizations in \Cref{sec:exp}}\label{app:vis-exp}
\begin{figure}[h!]
    \centering
    \subfigure[\textsc{Reddit-Binary}]{\includegraphics[width=0.24\textwidth]{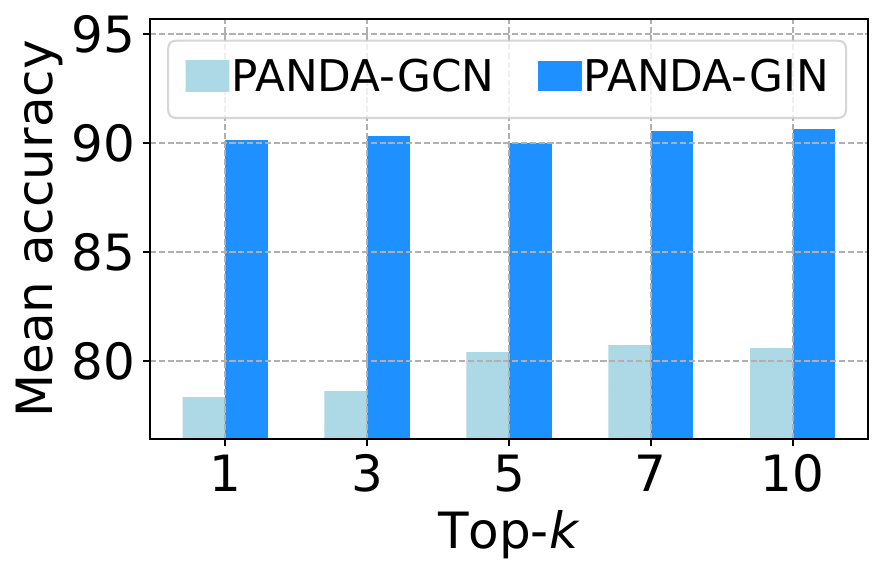}}
    \subfigure[\textsc{IMDB-Binary}]{\includegraphics[width=0.24\textwidth]{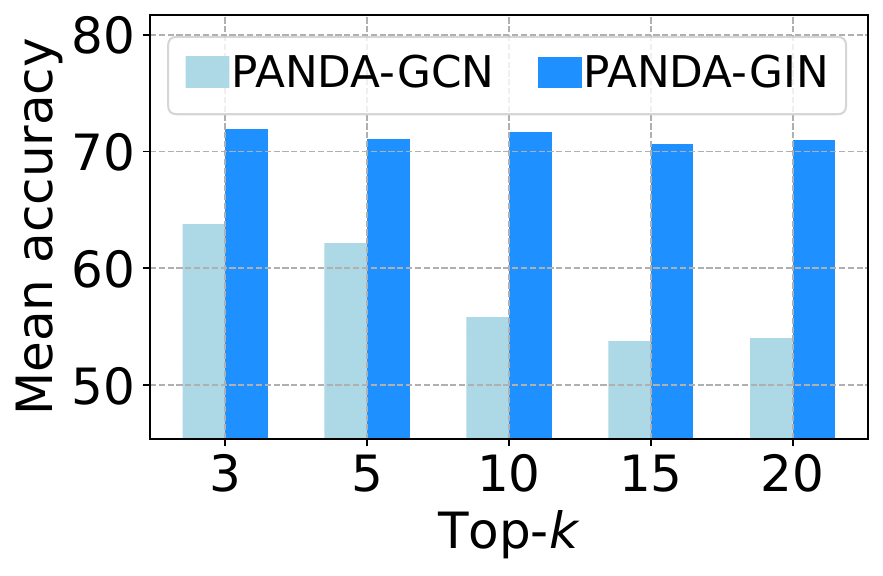}}
    \subfigure[\textsc{Mutag}]{\includegraphics[width=0.24\textwidth]{img/sensitivity/mutag_k.pdf}}
    \subfigure[\textsc{Enzymes}]{\includegraphics[width=0.24\textwidth]{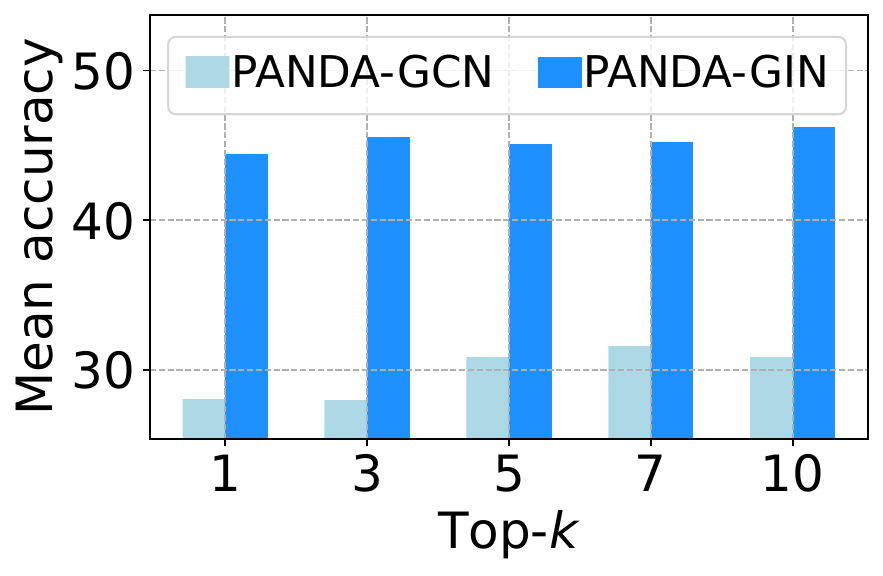}}
    \subfigure[\textsc{Proteins}]{\includegraphics[width=0.24\textwidth]{img/sensitivity/proteins_k.pdf}}
    \subfigure[\textsc{Collab}]{\includegraphics[width=0.24\textwidth]{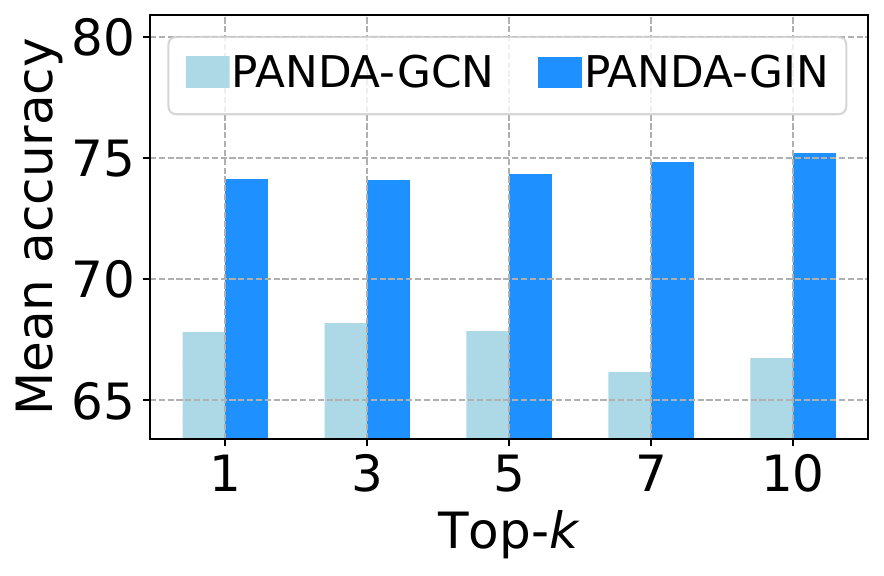}}
    \caption{Sensitivity on top-$k$ for all datasets.}
    \label{fig:topk-app}
\end{figure}

\begin{figure}[h!]
    \centering
    \subfigure[\textsc{Reddit-Binary}]{\includegraphics[width=0.24\textwidth]{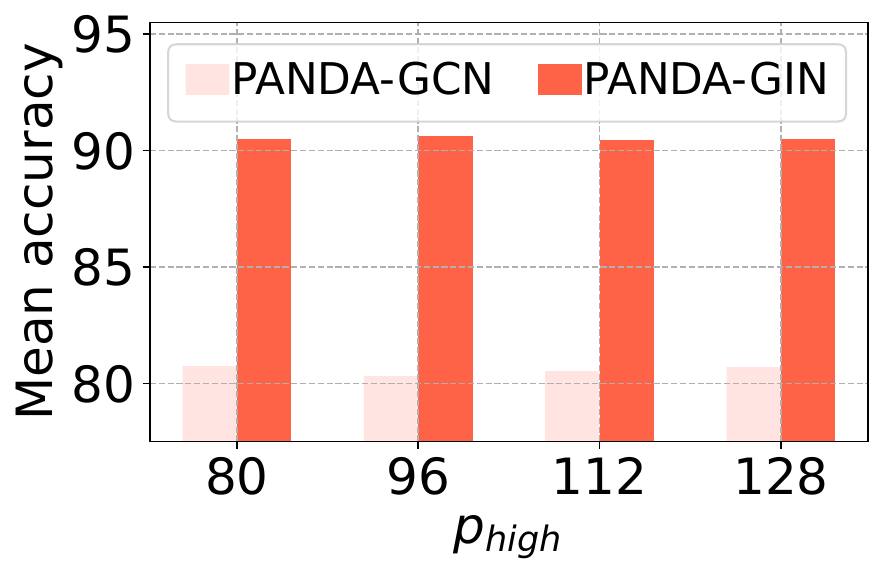}}
    \subfigure[\textsc{IMDB-Binary}]{\includegraphics[width=0.24\textwidth]{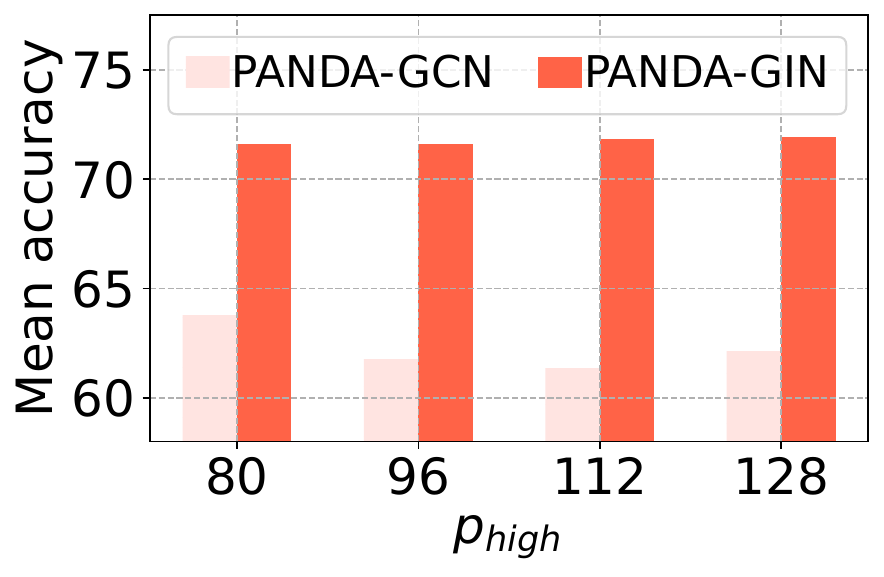}}
    \subfigure[\textsc{Mutag}]{\includegraphics[width=0.24\textwidth]{img/sensitivity/mutag_exp.pdf}}
    \subfigure[\textsc{Enzymes}]{\includegraphics[width=0.24\textwidth]{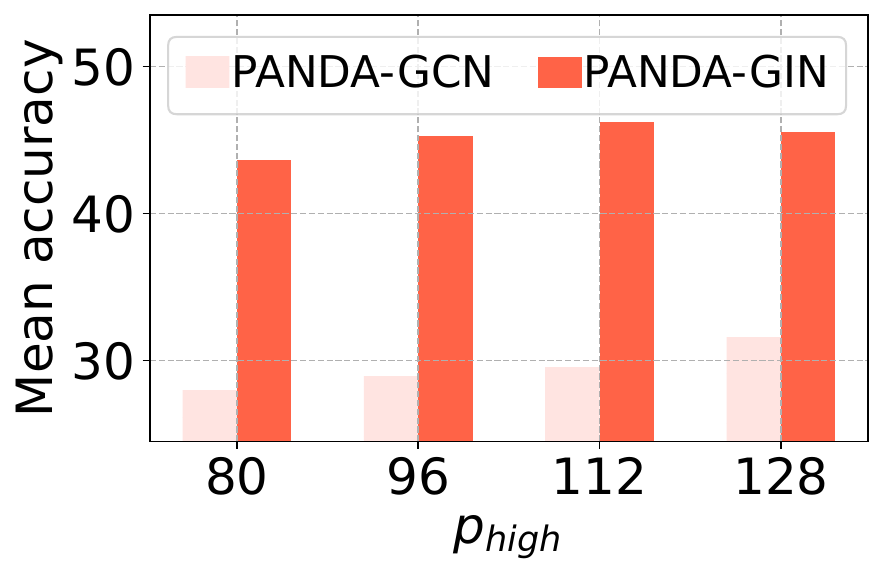}}
    \subfigure[\textsc{Proteins}]{\includegraphics[width=0.24\textwidth]{img/sensitivity/proteins_exp.pdf}}
    \subfigure[\textsc{Collab}]{\includegraphics[width=0.24\textwidth]{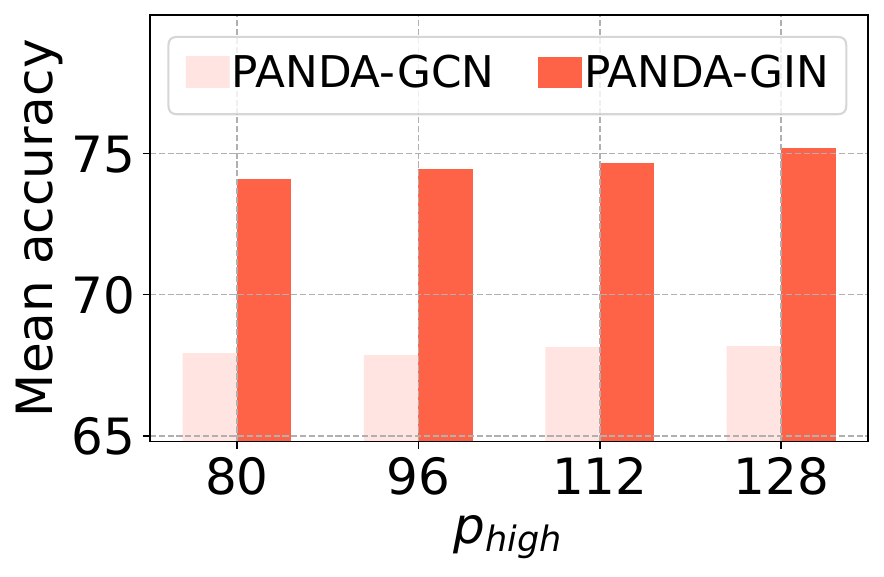}}
    \caption{Sensitivity on $p_{high}$ for all datasets.}
    \label{fig:exp-app}
\end{figure}

\clearpage

\section{Experiments on Node Classification}\label{app:node}
\paragraph{Datasets.}
For node classification, we use the following datasets: \textsc{Cora}, \textsc{Citeseer}~\cite{yang2016revisiting}, \textsc{Texas}, \textsc{Cornell}, \textsc{Wisconsin}~\cite{pei2020geomGCN}, and \textsc{Squirrel}~\cite{benedek2021musae}.

\paragraph{Experimental setting.}
Even if we allow different hidden sizes for the nodes, the output dimension of the last layer is the same for all nodes. Thus, it can also be applied to the task of node classification.
For node classification, we compare \textbf{PANDA} to no graph rewiring and 4 other state-of-the-art rewiring methods: DIGL~\citep{gasteiger2019digl}, SDRF~\citep{topping2022riccurvature}, FoSR~\citep{karhadkar2023fosr}, and BORF~\citep{karhadkar2023fosr}.
We use the same GCN settings across all methods to rule out hyper-parameter tuning as a source of performance improvement.
We accumulate results over 100 randomized trials, the same as \citet{nguyen2023borf}'s setup, and report the average test accuracy with 95\% confidence intervals.
We use the hyperparameters reported in \Cref{tab:best-param-node} as the best hyperparameters for \textbf{PANDA}.

\begin{table}[h!]
    \small
    \centering
    \caption{Best hyperparameter for node classification}
    \label{tab:best-param-node}
    \begin{tabular}{c cccccc}\toprule
        Hyperparameter & \textsc{Texas} & \textsc{Cornell} & \textsc{Wisconsin} & \textsc{Squirrel} & \textsc{Cora} & \textsc{Citeseer} \\ \midrule
        Top-$k$
         & 50  & 50 & 50 & 25 & 50 & 50 \\
        $p_{\mathrm{high}}$
         & 96 & 96  & 96 & 96 & 96 & 96 \\
        $C(\mathcal{G})$
         & Betweenness & Betweenness & Betweenness & Degree & Betweenness & Betweenness \\
        \bottomrule
    \end{tabular}
\end{table}

\paragraph{Results.}
As shown in \Cref{tab:node}, in almost all cases, \textbf{PANDA} shows higher accuracy than existing rewiring methods. 
In cases like \textsc{Cornell} and \textsc{Cora}, BORF still performs the best, but it is comparable to \textbf{PANDA}.
\textbf{PANDA} effectively improves the classification accuracy, especially in heterophilic graphs, such as \textsc{Texas}, \textsc{Squirrel}, and \textsc{Wisconsin}.
In the case of \textsc{Wisconsin}, it achieves higher average accuracy than both DIGL and BORF, with a 12.57\% improvement over DIGL.

\begin{table}[h]
    \small
    \setlength{\tabcolsep}{5pt}
    \centering
    \caption{Node classification accuracies of GCN with None, DIGL, SDRF, FoSR, BORF rewiring, and \textbf{PANDA} on various node classification datasets. We show the best three in \BEST{red} (first), \SECOND{blue} (second), and \THIRD{purple} (third).}
    \label{tab:node}
    \begin{tabular}{l cccccc}\toprule
        Method & \textsc{Texas} & \textsc{Cornell} & \textsc{Wisconsin} & \textsc{Squirrel} & \textsc{Cora} & \textsc{Citeseer} \\ \midrule
        GCN (None) & 44.2\std{1.5} & 41.5\std{1.8} & 44.6\std{1.4} & \THIRD{42.5\std{2.7}} & \THIRD{86.7\std{0.3}} & 72.3\std{0.3} \\ 
        \; + DIGL & \SECOND{53.6\std{1.5}} & \THIRD{44.5\std{1.2}} & \SECOND{51.7\std{1.3}} & 35.2\std{3.0} & 83.2\std{0.2} & \SECOND{74.5\std{0.3}}\\
        \; + SDRF & 43.9\std{1.6} & 42.2\std{1.5} & 46.2\std{1.2} & 42.4\std{2.5} & 86.3\std{0.3} & 72.6\std{0.3} \\ 
        \; + FoSR & \THIRD{46.0\std{1.6}} & 40.2\std{1.6} & 48.3\std{1.3} & \THIRD{42.5\std{1.9}} & 85.9\std{0.3} & 72.3\std{0.4} \\ 
        \; + BORF & 49.4\std{1.2} & \BEST{ 50.8\std{1.1}} & \THIRD{50.3\std{0.9}} & \SECOND{42.6\std{2.8}} & \BEST{87.5\std{0.2}} & \THIRD{73.8\std{0.2}}\\ 
        \cmidrule(lr){1-7}
        \; + \textbf{PANDA} & \BEST{55.4\std{1.6}} & \SECOND{47.5\std{1.4}} & \BEST{58.2\std{1.4}} & \BEST{43.1\std{0.9}} & \SECOND{87.0\std{0.3}} & \BEST{75.3\std{0.3}}\\ \bottomrule
    \end{tabular}
\end{table}

\section{Experiments on Long Range Graph Benchmark}\label{app:lrgb}
\paragraph{Datasets.}
We use the Peptides (15,535 graphs) dataset from the Long Range Graph Benchmark (LRGB)~\cite{dwivedi2022LRGB}.
There are two tasks related to peptides: i) \textsc{Peptides-func}, a peptide feature classification task, and ii) \textsc{Peptides-struct}, a peptide structure regression task. 

\paragraph{Experimental setting.}
We compare \textbf{PANDA} to no graph rewiring and 4 other state-of-the-art rewiring methods: DIGL~\citep{gasteiger2019digl}, SDRF~\citep{topping2022riccurvature}, FoSR~\citep{karhadkar2023fosr}, and BORF~\citep{karhadkar2023fosr}.
On the \textsc{Peptide-func} dataset, we use two experimental setups: the same experimental setup as \citet{nguyen2023borf} and the commonly used experimental setup of \citet{dwivedi2022LRGB} for a fair comparison between our method and the rewiring method. 
The setting for \citet{nguyen2023borf} uses a lower hidden dimension of 64 instead of 300, while the setting for \citet{dwivedi2022LRGB} uses a hidden dimension of 300.
In \textsc{Peptide-struct} dataset, we use the same experimental settings as those of \citet{dwivedi2022LRGB} after fixing the number of layers to 5. 
In particular, no additional positional features are used to confirm the effectiveness of the rewiring method and the effectiveness of \textbf{PANDA}.
We only replace the last layer of the backbone GCN with \textbf{PANDA} message passing for performance and efficiency.
\paragraph{Results.}
In \Cref{tab:lrgb}, high average precision (AP) and low mean absolute error (MAE) values are preferred. 
On the \textsc{Peptide-func} dataset, for the setting of \citet{nguyen2023borf}, FoSR and BORF slightly improve the average precision over GCN, but \textbf{PANDA} outperforms them. 
In the case of \citet{dwivedi2022LRGB}'s setting, existing rewiring methods show trivial improvement compared to GCN, while \textbf{PANDA} improves by 1.65\%.
Additionally, in \textsc{Peptides-struct}, FoSR achieves a performance improvement of 0.658\%, but \textbf{PANDA} achieves a performance improvement of 6.407\%. This shows that the \textbf{PANDA} message passing is effective even on LRGB datasets.

\begin{table}[h]
    \small
    \centering
    \caption{Results of GCN with None, SDRF, FoSR, BORF, and PANDA on \textsc{Peptides-func} and \textsc{Peptides-struct}. We show the best three in \BEST{red} (first), \SECOND{blue} (second), and \THIRD{purple} (third).}
    \label{tab:lrgb}
    \begin{tabular}{l ccc}\toprule
        \multirow{2}{*}{Method} & \multicolumn{2}{c}{\textsc{Peptides-func} (Test AP $\uparrow$)}  & \textsc{Peptides-struct} (Test MAE $\downarrow$) \\ \cmidrule(lr){2-3}\cmidrule(lr){4-4}
        & \citet{nguyen2023borf}'s setting & \citet{dwivedi2022LRGB}'s setting & \citet{dwivedi2022LRGB}'s setting \\ \midrule
        GCN (None) & 44.2\std{1.5}          & 59.30\std{0.23} & 0.3496\std{0.0013} \\ 
        \; + SDRF  & 43.9\std{1.6}          & \THIRD{59.47\std{1.26}} & \THIRD{0.3478\std{0.0013}} \\ 
        \; + FoSR  & \SECOND{46.0\std{1.6}} & \THIRD{59.47\std{0.35}} & \SECOND{0.3473\std{0.0007}} \\ 
        \; + BORF  & \THIRD{45.6\std{1.5}}  & \SECOND{59.94\std{0.37}} & 0.3514\std{0.0009} \\ 
        \cmidrule(lr){1-4}
        \; + \textbf{PANDA} & \BEST{55.4\std{1.6}} & \BEST{60.28\std{0.31}} & \BEST{0.3272\std{0.0001}} \\ \bottomrule
    \end{tabular}
\end{table}


\end{document}

%% file: math_commands.tex
\usepackage{amsmath,amsfonts,bm}

\def\1{\bm{1}}

\DeclareMathAlphabet{\mathsfit}{\encodingdefault}{\sfdefault}{m}{sl}
\SetMathAlphabet{\mathsfit}{bold}{\encodingdefault}{\sfdefault}{bx}{n}

\newcommand{\softmax}{\mathrm{softmax}}
\newcommand{\sigmoid}{\sigma}

